\documentclass[]{sensenova}

\usepackage{hyperref}
\usepackage{cleveref}
\usepackage{verbatim}

\usepackage{wrapfig}
\usepackage{graphicx}
\usepackage{subcaption}
\usepackage{listings}
\usepackage{algorithm}

\usepackage{mmstyles}
\usepackage[misc]{ifsym}

\usepackage[toc,page,header]{appendix}

\title{Phased DMD: Few-step Distribution Matching Distillation via \\ Score Matching within Subintervals}

\author{
\parbox{\textwidth}{\centering
Xiangyu Fan$^{1}$, Zesong Qiu$^{1}$, Zhuguanyu Wu$^{1}$, Fanzhou Wang$^{1}$, Zhiqian Lin$^{1}$, \protect\\
Tianxiang Ren$^{1}$, Dahua Lin$^{1}$, Ruihao Gong$^{1,2}$, Lei Yang$^{1,\textrm{\Letter}}$ \\
$^1$SenseTime Research, $^2$Beihang University \\
}
}

\newcommand{\ourmodel}{Phased DMD~}
\newcommand{\sgts}{SGTS~}

\abstract{
Distribution Matching Distillation (DMD) distills score-based generative models into efficient one-step generators,
without requiring a one-to-one correspondence with the sampling trajectories of their teachers.
Yet, the limited capacity of one-step distilled models compromises generative diversity and degrades performance in complex generative tasks, e.g., 
generating intricate object motions in text-to-video tasks.
Directly extending DMD to multi-step distillation increases memory usage and computational depth, leading to instability and reduced efficiency. 
While prior works propose stochastic gradient truncation as a potential solution,
we observe that it substantially reduces the generative diversity in text-to-image generation and slows motion dynamics in video generation, reducing performance to the level of one-step models.
To address these limitations, we propose \textbf{Phased DMD}, a multi-step distillation framework that bridges the idea of phase-wise distillation with Mixture-of-Experts (MoE), reducing learning difficulty while enhancing model capacity.
\ourmodel incorporates two key ideas: \textit{\textbf{progressive distribution matching}} and \textit{\textbf{score matching within subintervals}}.
First, our model divides the SNR range into subintervals, progressively refining the model to higher SNR levels, to better capture complex distributions.
Next, to ensure accurate training within each subinterval, we derive rigorous mathematical formulations for the objective.
We validate \ourmodel by distilling state-of-the-art image and video generation models, including Qwen-Image-20B and Wan2.2-28B.
Experiments demonstrate that \ourmodel 
enhances motion dynamics, improves visual fidelity in video generation, and increases output diversity in image generation.
Our code and models are available at \url{https://x-niper.github.io/projects/Phased-DMD/}.
}

\begin{document}
\maketitle
\section{Introduction}
\label{sec:intro}

Recently, state-of-the-art (SOTA) diffusion models have made significant progress in image and video generation.
In image generation, SOTA models~\cite{wu2025qwenimagetechnicalreport,gpt_image,HunyuanImage-2.1,nano_banana} demonstrate precise prompt control,
enabling complex text-to-image rendering and accurate layout specification.
In video generation, these models~\cite{wan2025,kong2024hunyuanvideo,veo3,sora} exhibit substantial improvements in dynamic scene generation,
such as fast-moving objects in sports and complex camera movements like ego-centric videos.
However, the increasing parameter sizes and computational demands of base models highlight
the importance of accelerating diffusion model sampling.

Several techniques have been proposed to accelerate diffusion models,
including classifier-free guidance (CFG) distillation~\cite{cfg_distill},
step distillation~\cite{song2023consistencymodels,wang2024phased,salimans2022progressive,yin2024DMD2,luo2023diff_instruct,luo2024diff_instruct_pp,zhou2024sid,huang2024fgm,lin2024sdxl_lighting,lin2025APT,frans2024_shotcut,geng2025_meanflow}, SVDQuant~\cite{li2024svdquant}, Mixture-of-Experts (MoE)~\cite{balaji2022ediff,feng2023ernie,wan2025}, and parallel computation~\cite{fang2024xdit}.
Among these, step distillation methods based on Variational Score Distillation (VSD)~\cite{wang2023prolificdreamer_vsd}, including diff-instruct~\cite{luo2023diff_instruct}, DMD~\cite{yin2024DMD2,yin2024_DMD1}, SID~\cite{zhou2024sid}, 
achieve high-quality generation by distilling models into single-step generators.
However, single-step distilled models suffer from limited network capacity~\cite{lin2024sdxl_lighting,lin2025APT,yin2024DMD2}, diminishing output diversity and undermining performance on complex tasks like intricate text rendering and dynamic scene generation.
% A key limitation of single-step distilled models lies in  their limited network capacity~\cite{lin2024sdxl_lighting,lin2025APT,yin2024DMD2}, which diminishes output diversity and undermines their performance on complex tasks such as intricate text rendering and dynamic scene generation.

\begin{figure*}[tb]
  \centering
    \includegraphics[width=0.9\linewidth]{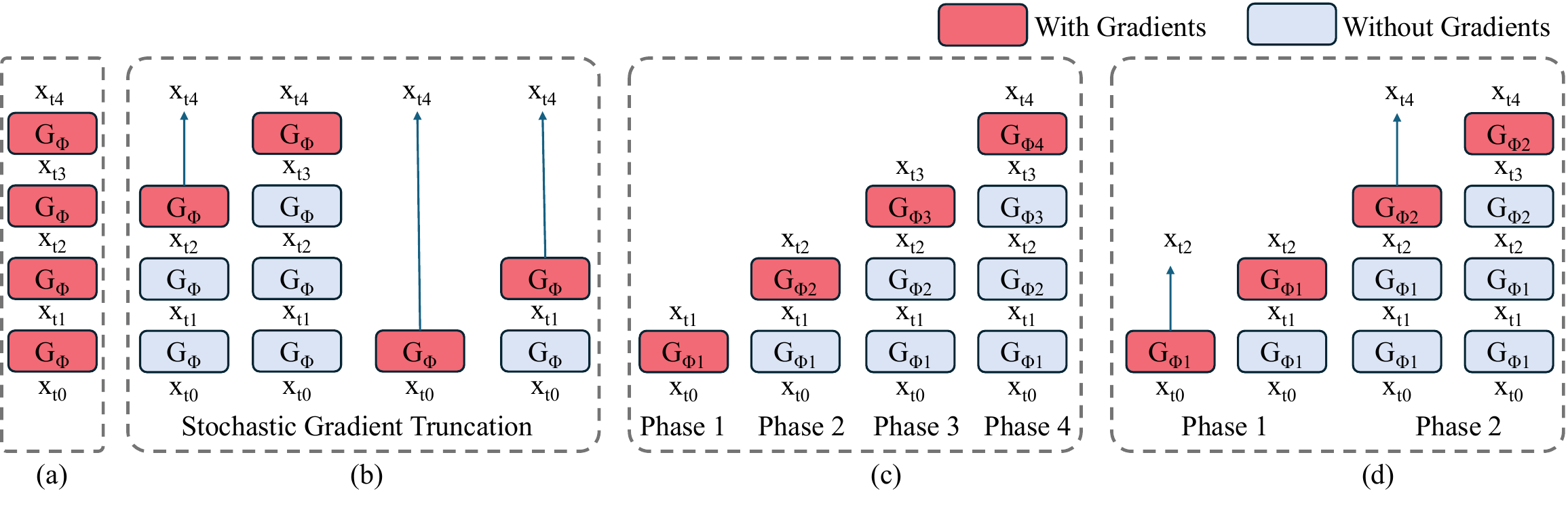}

  % \begin{subfigure}{0.24\linewidth}
  %   \includegraphics[width=\linewidth]{figures/images/phase_dmd_DMD.pdf}
  %   \caption{DMD}
  %   \label{subfig:vanilla_DMD}
  % \end{subfigure}
  % % \hfill
  % \begin{subfigure}{0.24\linewidth}
  %   \includegraphics[width=\linewidth]{figures/images/phase_dmd_DMD_wi_sgt.pdf}
  %   \caption{DMD with SGTS}
  %   \label{subfig:DMD_wi_SGTS}
  % \end{subfigure}
  % \begin{subfigure}{0.24\linewidth}
  %   \includegraphics[width=\linewidth]{figures/images/phase_dmd_phase_DMD_4phase.pdf}
  %   \caption{4-phase DMD}
  %   \label{subfig:4_phase_dmd}
  % \end{subfigure}
  % \begin{subfigure}{0.24\linewidth}
  %   \includegraphics[width=\linewidth]{figures/images/phase_dmd_phase_DMD_2phase.pdf}
  %   \caption{2-phase DMD}
  %   \label{subfig:2_phase_dmd}
  % \end{subfigure}
  \caption{
    Schematic diagram of the backward simulation process in
    (a) Vanilla few-step DMD,
    (b) DMD2~\citep{yin2024DMD2},
    (c) \ourmodel
    and (d) \ourmodel with SGTS.}
  \label{fig:Schematic_diagram}
\end{figure*}

Few-step distillation has emerged as a solution to this trade-off, effectively balancing computational cost with the preservation of generative quality and diversity~\cite{luo2025TDM,yin2024DMD2}.
However, as illustrated in Fig.~\ref{fig:Schematic_diagram}a, directly applying VSD to few-step distillation through 
% \textbf{backward simulation}~\cite{yin2024DMD2} 
multi-step generation 
presents challenges, including increased computational graph depth and higher memory overhead, which hinders its scalability to larger models and video generation tasks. 
Moreover, the absence of explicit constraints on intermediate generator steps compromises training stability and results in suboptimal performance for few-step models.
To address these issues, DMD2~\cite{yin2024DMD2} and Self-Forcing \cite{huang2025_self_forcing} introduce a stochastic gradient truncation strategy (SGTS),
where few-step backward simulation may terminate at a random step and gradient backpropagation is restricted to the final denoising step (see Fig.~\ref{fig:Schematic_diagram}b).
This approach improves training convergence and stability by supervising all intermediate steps while enhancing memory efficiency via gradient detachment for non-final steps.
However, \sgts can terminate backward simulation after only one step during training, essentially distilling a one-step generator for that iteration. 
As a result, the generative diversity and motion dynamics of videos produced by few-step generators trained with \sgts are reduced to levels comparable to those of their one-step generators.

Diffusion theory~\cite{song2020SDE} suggests the existence of infinitely many neural networks serving as score estimators across a range of signal-to-noise ratios (SNR), spanning from zero to infinity.
During the generation process, diffusion models exhibit distinct temporal dynamics~\cite{balaji2022ediff}.
In particular, the low-SNR stage emphasizes modeling visual structures and dynamics, while the high-SNR stage focuses on refining visual details.
In practice, a single neural network is typically utilized throughout the denoising process, requiring the model to learn and execute a diverse range of denoising tasks simultaneously.
% to simultaneously learn and perform a variety of denoising tasks.
Recent studies~\cite{balaji2022ediff,feng2023ernie,wan2025} have integrated a Mixture of Experts (MoE) architecture into diffusion models.
By assigning specialized experts to different SNR levels, MoE enhances model capacity and generative performance without increasing inference cost.
The performance improvement is particularly pronounced in video generation \citep{wan2025},
where the low-SNR expert excels at capturing dynamic content.

In this work, we propose \textbf{Phased DMD}, a novel distillation framework for few-step generation.
Our motivation is to achieve scalable distillation for large generative models and video generation tasks by leveraging gradient truncation to manage memory constraints, while simultaneously avoiding the associated one-step degeneration problem to preserve superior generative diversity and motion dynamics.
Our method is built upon two key components:
\begin{itemize}
\item \textbf{Progressive distribution matching:}
In each phase, a single expert is distilled for a specific SNR subinterval, progressively advancing toward higher SNR levels.
The backward simulation terminates at the phase's corresponding SNR level rather than the clean state, thereby mitigating the one-step degradation associated with SGTS. 
This framework is conceptually analogous to ProGAN~\cite{karras2017_proGAN}, which progressively trains a generator to handle higher resolutions. It is fundamentally distinct from progressive distillation~\cite{salimans2022progressive}, where the objective is to halve the number of sampling steps in each phase.
% Conceptually similar to ProGAN~\cite{karras2017_proGAN}, which progressively trains a generator to handle higher resolutions, \ourmodel divides SNR into subintervals and progressively distills models toward higher SNR levels. Notably, the concept of progressive training in \ourmodel is \textbf{NOT} similar to prior progressive distillation~\cite{salimans2022progressive} where the number of target distillation steps is reduced by half in each phase.

\item \textbf{Score matching within SNR subintervals:}
Since the clean sample is inaccessible in all but the final phase, the training objective must be reformulated to preserve the theoretical integrity of distribution matching. To this end, we derive a revised objective for unbiased score estimation when the clean sample is unavailable (see Sec.~\ref{subsubsec:score_matching_within_subinterval}).
% As each phase is trained within a subinterval, the training objective undergoes a transformation. To ensure theoretical rigor, we derive the training objective for the fake score estimator within each subinterval.
\end{itemize}
%
% Our approach is inspired by a broader vision: 
% \textit{By decomposing a complex task into learnable phases, each phase naturally forms an expert, collectively enhancing the model's capacity in a MoE manner}.

As illustrated in Fig.~\ref{fig:Schematic_diagram}c, \ourmodel offers several advantages:
\textbf{First}, by partitioning SNR into subintervals, the model learns complex data distributions incrementally, improving training stability and generative performance while avoiding degeneration into a one-step generator.
\textbf{Second}, each phase involves only a single gradient-recorded sampling step, avoiding additional computational and memory overhead.
\textbf{Third}, notably, \ourmodel inherently produces a few-step MoE generative model, irrespective of whether the teacher model employs an MoE architecture.
\textbf{Last}, as shown in Fig.~\ref{fig:Schematic_diagram}d, \ourmodel can be combined with \sgts, enabling the training of a 4-step generator within 2 phases, simplifying the overall framework complexity.
We validate \ourmodel by distilling SOTA image and video generation models,
including Qwen-Image \citep{wu2025qwenimagetechnicalreport} with 20B parameters and Wan2.1/Wan2.2 \citep{wan2025} with 14/28B parameters.
Results demonstrate that \ourmodel 
enhances motion dynamics, improves visual fidelity in video generation, and increases output diversity in image generation.
% better preserves output diversity in text-to-image tasks, as well as realistic dynamic motion and precise camera control in text-to-video and image-to-video tasks. 

% video motion dynamics compared to standard DMD while maintaining the base models’ key capabilities,
% such as faithful text rendering in Qwen-Image and realistic dynamic motion in Wan2.2.

Our contributions are summarized as follows:

\begin{itemize}

\item \textbf{Phased DMD:} A data-free distillation framework for few-step diffusion models.
This framework combines ideas from DMD and MoE, achieving higher performance ceilings while maintaining memory usage similar to single-step distillation.

\item \textbf{Theoretical Objective:} We derive the theoretical training objective for subinterval diffusion models without relying on clean samples.
We highlight the necessity of this correctness for DMD distillation.

\item \textbf{SOTA Performance:} Without requiring GAN loss or regression loss, \ourmodel achieves SOTA results on text-to-image and text-to-video tasks.
To the best of our knowledge, this is the largest reported distillation validation for diffusion models.
Experiments show that our method 
effectively reduces diversity loss while preserving the base models’ key capabilities,
including complex text rendering and highly dynamic video generation.
% We will release the code and models to the community.
\end{itemize}

\section{Method}
\label{sec:method}

To clarify the principle of Phased DMD, we begin by introducing the theoretical background and notations related to diffusion models \citep{kingma2023_vdm,zhang2024tackling}, score matching \citep{song2020SDE,karras2022_edm}, and distribution matching distillation \citep{yin2024_DMD1,yin2024DMD2}.
We explicitly highlight why the principle of DMD is applicable only to score-based generative models.
Building on this foundation, we present the motivation behind \ourmodel and explain how it inherently achieves improved generative diversity.
Following this, we detail the two key components of Phased DMD: progressive distribution matching and score matching within subintervals.

\subsection{Preliminary}

\subsubsection{Diffusion Models and Score Matching}

Consider a continuous-time Gaussian diffusion process defined over the interval ${\displaystyle 0 \leq t \leq 1}$.
The ground-truth distribution is denoted ${\displaystyle p(x_0)}$.
For any  ${\displaystyle 0 \leq t \leq 1}$, the forward diffusion process is described by the following conditional distribution:
\begin{equation}
p(x_t |x_0) = \mathcal{N} (\alpha_t x_0, \sigma_t^2  \boldsymbol{I})
\label{eq:p_xt_x0}
\end{equation}
where  ${\alpha_t}$ and ${\sigma_t^2}$ are positive, scalar functions of $t$.
The signal-to-noise ratio (SNR) is defined as ${\text{SNR}(t) = \frac{\alpha_t^2}{\sigma_t^2}}$.
It is assumed that 
 ${\text{SNR}(t)}$ is strictly monotonically decreasing over time. 
 % i.e., 
% for any ${\displaystyle 0 \leq s < t \leq 1}$, we have ${\text{SNR}(s) > \text{SNR}(t)}$.
No additional constraints are imposed on the relationship between ${\alpha_t}$ and ${\sigma_t}$,
ensuring the notations are compatible with different kinds of diffusion models~\cite{ho2020_ddpm,karras2022_edm,song2022_DDIM,podell2023_SDXL} and flow models~\cite{liu2022flow,esser2024_sd3}.
The diffusion process is Markovian~\cite{kingma2023_vdm}, meaning that  ${\displaystyle p(x_t |x_s,x_0) = p(x_t |x_s)}$.
For any ${\displaystyle 0 \leq s < t \leq 1}$, ${\displaystyle p(x_t |x_s)}$ is also Gaussian, and can be expressed as:
\begin{equation}
p(x_t |x_s) = \mathcal{N} (\alpha_{t|s} x_s, \sigma_{t|s}^2 \boldsymbol{I})
\label{eq:p_xt_xs}
\end{equation}
where ${\alpha_{t|s} = \frac{\alpha_{t}}{\alpha_{s}}}$ and ${\sigma_{t|s}^2 = \sigma_{t}^2 - \alpha_{t|s}^2 \sigma_{s}^2}$.
% 
% The marginal distribution of ${\displaystyle x_s}$ and ${\displaystyle x_t}$ are given by ${\displaystyle p(x_s) = \int p(x_s |x_0) p(x_0) dx_0}$
% and ${\displaystyle p(x_t) = \int p(x_t |x_0) p(x_0) dx_0}$.
% 
% If only  ${\displaystyle p(x_s)}$ is observed and not ${\displaystyle p(x_0)}$,
% % 
% the marginal distribution of ${\displaystyle x_t}$ can alternatively be expressed as: ${\displaystyle p(x_t) = \int p(x_t |x_s) p(x_s) dx_s}$.
% 
For the marginal distribution of $x_t$, we have the following equivalence:
\begin{equation}
\displaystyle p(x_t) = \int p(x_t |x_0) p(x_0) dx_0 = \int p(x_t |x_s) p(x_s) dx_s
\label{eq:p_xt_three_equal}
\end{equation}
In the training process,
${\alpha_t}$ and ${\sigma_t}$ are predefined functions of $t$,
while ${x_0}$ is sampled from the dataset distribution ${x_0} \sim p(x_0)$.
Timestep $t$ is sampled from a predefined distribution over the interval [0, 1],
such as a uniform or logit-normal distribution \citep{esser2024_sd3}, \ie,
$t \sim \mathcal{T}(0, 1)$.
The sample ${x_t}$ is then given by ${x_t = \alpha_t x_0 + \sigma_t \boldsymbol{\epsilon}}$,
where  $\boldsymbol{\epsilon} \sim \mathcal{N}(0, \boldsymbol{I})$.
We use $t \sim \mathcal{T}$ and $\boldsymbol{\epsilon} \sim \mathcal{N}$ for brevity in later paragraphs unless otherwise specified.
\citet{song2020SDE} unified diffusion models under the theoretical framework of score-based generative models and demonstrated that the continuous diffusion process is fundamentally governed by a Stochastic Differential Equation (SDE).
% 
% Prior works on diffusion models have explored various prediction targets for the parameterized neural network and they can be unified under the theoretical framework of score-based generative models
% Prior works in the field of diffusion models has adopted different kinds of prediction types as the target of the parametrized neural networks,
% such as noise prediction \citep{ho2020_ddpm}, sample prediction \citep{karras2022_edm}, velocity prediction \citep{salimans2022progressive} and flow velocity prediction \citep{liu2022flow}.
% \citet{song2020SDE} unifies diffusion models in the theory of score based generative models and shows that
% the continuous diffusion process is essentially governed by a stochastic differential equation (SDE).
% \citet{karras2022_edm} shows the connection between the coefficients in the SDE and ${\alpha_t,  \sigma_t}$.
% 
Here, we adopt flow velocity prediction as an example and demonstrate its connection to score matching.
Let $\boldsymbol{\psi}$ denote a diffusion model.
The relationship between flow matching loss and score matching is expressed below.
% 
% \begin{align}
% \begin{equation}
% \begin{aligned}
% J_{flow}(\boldsymbol{\theta}) = & E_{x_0 \sim p(x_0),
% \boldsymbol{\epsilon} \sim \mathcal{N}, t \sim \mathcal{T}, x_t = \alpha_t x_0 + \sigma_t \boldsymbol{\epsilon}} \\
% & [\|\boldsymbol{\psi}_{\boldsymbol{\theta}}(x_t, t) - (\boldsymbol{\epsilon} - x_0) \|^2] \label{eq:FlowMatchTarget} \\
% % 
% = &  E_{x_0 \sim p(x_0), t \sim \mathcal{T}, x_t \sim p(x_t | x_0)} \\
% &  [\|\boldsymbol{\psi}_{\boldsymbol{\theta}}(x_t) + x_t / \alpha_{t} + (\sigma_t +\sigma_t^2/\alpha_t)  \nabla{x_t}\log(p(x_t | x_0)) \|^2]  \\
% % 
% = & E_{t \sim \mathcal{T}, x_t \sim p(x_t)} \\
% & [\|\boldsymbol{\psi}_{\boldsymbol{\theta}}(x_t) + x_t / \alpha_{t} + (\sigma_t +\sigma_t^2/\alpha_t)  \nabla{x_t}\log(p(x_t)) \|^2] \label{eq:FlowMatchTarget_2}
% \end{aligned}
% \end{equation}
% % \end{align} 
\begin{gather}
E_{x_0, t,
\boldsymbol{\epsilon}, x_t = \alpha_t x_0 + \sigma_t \boldsymbol{\epsilon}}[\|\boldsymbol{\psi}(x_t, t) - (\boldsymbol{\epsilon} - x_0) \|^2] \label{eq:FlowMatchTarget_1} \\
= E_{t, x_t}
[\|\boldsymbol{\psi}(x_t, t) + \frac{x_t}{\alpha_{t}} + (\sigma_t + \frac{\sigma_t^2}{\alpha_t})  \nabla_{x_t}\log(p(x_t)) \|^2] \label{eq:FlowMatchTarget_2}
\end{gather}
% 
% The formula in the last line 
Eq.~\ref{eq:FlowMatchTarget_2}
is derived based on the equivalence between denoising score matching (DSM) and explicit score matching (ESM),
as originally proven in \cite{vincent2011_connection_DSM_ESM}.
In supplementary materials, we provide the detailed derivation of Eq.~\ref{eq:FlowMatchTarget_2}.
Additionally, we demonstrate the connection between sample prediction (a.k.a. x-prediction) and score matching in the supplementary materials.

\subsubsection{Distribution Matching Distillation}

The DMD framework comprises three components: a trainable generator ${\boldsymbol{G}_{\boldsymbol{\phi}}}$, a trainable fake score estimator ${\boldsymbol{F}_{\boldsymbol{\theta}}}$, and a frozen pretrained teacher score estimator ${\boldsymbol{T}_{\boldsymbol{\hat{\theta}}}}$, parameterized by  ${\boldsymbol{\phi}}$, ${\boldsymbol{\theta}}$ and ${\boldsymbol{\hat{\theta}}}$, respectively.
Usually, both ${\boldsymbol{\phi}}$ and  ${\boldsymbol{\theta}}$ are initialized from ${\boldsymbol{\hat{\theta}}}$. 
Formally, the objective of DMD is to minimize the reverse Kullback-Leibler (KL) divergence between the real data distribution ${\displaystyle p_{real}(x_0)}$ and the generated data distribution ${\displaystyle p_{fake}(x_0)}$, the latter produced by ${\boldsymbol{G}_{\boldsymbol{\phi}}}$.
\begin{equation}
% \begin{aligned}
\nabla_{\boldsymbol{\phi}} D_{KL}(p_{fake} \| p_{real}) = E_{z, x_0 = \boldsymbol{G}_{\boldsymbol{\phi}}(z)} [(\nabla_{x_0}\log p_{fake}(x_0) - \nabla_{x_0} \log p_{real}(x_0)) \frac{d \boldsymbol{G}}{d \boldsymbol{\phi}} ] 
\label{eq:reverse_KL}
% \end{aligned}
\end{equation} 
where $z \sim \mathcal{N}(0, \boldsymbol{I})$ is a random Gaussian noise input. We use $D_{KL}$ to abbreviate $D_{KL}(p_{fake} \| p_{real})$ in later paragraphs.
To leverage diffusion models as score estimators, the generated samples are diffused and the objective becomes:
% 
% \begin{equation}
% % D_{KL}(p_{fake}(x_t) \| p_{real}(x_t)) =  E_{\boldsymbol{\epsilon} \sim \mathcal{N}, x_0 = \boldsymbol{G}_{\boldsymbol{\phi}}(\boldsymbol{\epsilon}), t \sim \mathcal{T}, x_t \sim p(x_t | x_0)} [log p_{fake}(x_t) - log p_{real}(x_t)]
% E_{z, x_0 = \boldsymbol{G}_{\boldsymbol{\phi}}(z), t, \boldsymbol{\epsilon}, x_t = \alpha_t x_0 + \sigma_t \boldsymbol{\epsilon}} [\log p_{fake}(x_t) - \log p_{real}(x_t)]
% \label{eq:reverse_KL_noised}
% \end{equation}
% 
% By combining Eq.~\ref{eq:FlowMatchTarget_2} and Eq.~\ref{eq:reverse_KL_noised},
% we can approximate the objective as:
% 
% \begin{equation}
% E_{z, x_0 = \boldsymbol{G}_{\boldsymbol{\phi}}(z), t, \boldsymbol{\epsilon}, x_t = \alpha_t x_0 + \sigma_t \boldsymbol{\epsilon}} [\lambda_t (\boldsymbol{T}_{\boldsymbol{\hat{\theta}}}(x_t, t) - \boldsymbol{F}_{\boldsymbol{\theta}}(x_t, t))]
% \label{eq:reverse_KL_noised_v_theta}
% \end{equation}
% 
% where $\lambda_t = 1 / (\sigma_t +\sigma_t^2/\alpha_t)$.
% % 
% $\boldsymbol{\theta}$ is initialized from $\boldsymbol{\hat{\theta}}$ and ${\boldsymbol{F}_{\boldsymbol{\theta}}}$ is updated on ${\displaystyle p_{fake}(x_0)}$ according to Eq.~\ref{eq:FlowMatchTarget}.
% 
% Taking the gradient of Eq.~\ref{eq:reverse_KL_noised_v_theta} with respect to the generator parameters, we have:
% 
\begin{equation}
\nabla_{\boldsymbol{\phi}}D_{KL} \approx E_{z, x_0 = \boldsymbol{G}_{\boldsymbol{\phi}}(z), t, \boldsymbol{\epsilon}, x_t = \alpha_t x_0 + \sigma_t \boldsymbol{\epsilon}} [w_t (\boldsymbol{T}_{\boldsymbol{\hat{\theta}}}(x_t, t) - \boldsymbol{F}_{\boldsymbol{\theta}}(x_t, t))\frac{d\boldsymbol{G}}{d \boldsymbol{\phi}}] 
\label{eq:reverse_KL_noised_v_theta_gradient}
\end{equation}
where $w_t = \frac{\alpha_t^2}{\alpha_t \sigma_t  +\sigma_t^2}$.

Intuitively, in DMD, the generator $\boldsymbol{G}_{\boldsymbol{\phi}}$ is optimized to approximate the real data distribution.  The fake score estimator $\boldsymbol{F}_{\boldsymbol{\theta}}$ is trained to estimate the score
of the generator’s output distribution. The update direction for the generator is then governed by
the discrepancy between the teacher score estimator and the fake score estimator.
The theoretical foundation of DMD rests on two key assumptions concerning the fake score estimator ${\boldsymbol{F}_{\boldsymbol{\theta}}}$.

\textbf{A1: Converged
${\boldsymbol{F}_{\boldsymbol{\theta}}}$.
} 
Analogous to GANs~\cite{goodfellow2014generativeadversarialnetworks}, 
DMD itself is an adversarial training paradigm that proceeds in two stages per iteration. 
In the fake diffusion optimization stage, ${\boldsymbol{F}_{\boldsymbol{\theta}}}$ is trained on the generated distribution according to the following loss:
\begin{equation}
E_{z, x_0 = \boldsymbol{G}_{\boldsymbol{\phi}}(z), t, \boldsymbol{\epsilon}, x_t = \alpha_t x_0 + \sigma_t \boldsymbol{\epsilon}} [\|\boldsymbol{F}_{\boldsymbol{\theta}}(x_t, t) - (\boldsymbol{\epsilon} - x_0) \|^2] \label{eq:FlowMatchTarget_in_DMD}
\end{equation}
This enables ${\boldsymbol{F}_{\boldsymbol{\theta}}}$ to function as an effective score estimator for $p_{fake}(x_t)$.
In the generator optimization stage, ${\boldsymbol{G}_{\boldsymbol{\phi}}}$ is updated according to Eq.~\ref{eq:reverse_KL_noised_v_theta_gradient},
thereby encouraging the generated distribution to more closely approximate the real data distribution.
Theoretically, the convergence of DMD hinges on the convergence of ${\boldsymbol{F}_{\boldsymbol{\theta}}}$ for each update of ${\boldsymbol{G}_{\boldsymbol{\phi}}}$. In practice, ${\boldsymbol{F}_{\boldsymbol{\theta}}}$ is updated more frequently than ${\boldsymbol{G}_{\boldsymbol{\phi}}}$, allowing it to accurately estimate the score of the evolving generated distribution~\cite{yin2024DMD2}.

\textbf{A2: Unbiased
${\boldsymbol{F}_{\boldsymbol{\theta}}}$.
}
Suppose ${\boldsymbol{T}_{\boldsymbol{\hat{\theta}}}}(x_t, t) \approx a_t  \nabla_{x_t}\log(p_{real}(x_t)) + b_t x_t $ and ${\boldsymbol{F}_{\boldsymbol{\theta}}}(x_t, t) \approx c_t  \nabla_{x_t}\log(p_{fake}(x_t)) + d_t x_t $, where $a_t, b_t, c_t, d_t$ are scalar functions of $t$, then the derivation from Eq.~\ref{eq:reverse_KL} to Eq.~\ref{eq:reverse_KL_noised_v_theta_gradient} holds provided $a_t = c_t$ and $b_t = d_t$. In DMD, this condition is inherently satisfied, owing to the identical training targets in Eq.~\ref{eq:FlowMatchTarget_1} and Eq.~\ref{eq:FlowMatchTarget_in_DMD}.
% 
% \subsection{From One-step Distillation to Few-step Distillation}
\subsubsection{Few-step Generator}
% Directly extending one-step distillation to N-step distillation requires only a little modification on Eq.~\ref{eq:reverse_KL_noised_v_theta_gradient}.
In $N$-step distillation, a diffusion scheduler $\mathcal{S}$ is employed with $N + 1$ timesteps, denoted by the sequence $ \boldsymbol{\vec{t}} = \{t_0, t_1, t_2, ..., t_N\}$, where $0 = t_N < t_{i} < t_{i-1} < t_0 = 1$ for all $i \in \{2, ..., N-1\}$.
The backward simulation~\cite{yin2024DMD2} begins with $x_{t_0} = z \sim \mathcal{N}(0, \boldsymbol{I})$. 
The sample $x_0$ is then generated iteratively:
for $i = 0, 1, ..., N-1$, $x_{t_{i+1}} = \mathcal{S}(\boldsymbol{G}_{\boldsymbol{\phi}}(x_{t_i}, t_i), x_{t_i}, t_i, t_{i+1})$.
Although DMD2~\cite{yin2024DMD2} utilizes the LCM~\cite{luo2023_LCM} scheduler, $\mathcal{S}$ can be any stochastic or deterministic solver.
Let $ \operatorname{pipeline}(\boldsymbol{G}_{\boldsymbol{\phi}}, \boldsymbol{\vec{t}}, z, \mathcal{S} )$ denote this iterative sampling procedure. Eq.~\ref{eq:reverse_KL_noised_v_theta_gradient} is thus adapted as follows: 
\begin{equation}
\nabla_{\boldsymbol{\phi}}D_{KL} \approx 
E_{z, x_0 = \operatorname{pipeline}(\boldsymbol{G}_{\boldsymbol{\phi}}, \boldsymbol{\vec{t}}, z, \mathcal{S} ), t, \boldsymbol{\epsilon}, x_t = \alpha_t x_0 + \sigma_t \boldsymbol{\epsilon}}
 [w_t (\boldsymbol{T}_{\boldsymbol{\hat{\theta}}}(x_t, t) - \boldsymbol{F}_{\boldsymbol{\theta}}(x_t, t))\frac{d\boldsymbol{G}}{d \boldsymbol{\phi}}]
\label{eq:reverse_KL_noised_v_theta_gradient_fewstep}
\end{equation}
As illustrated in Fig.~\ref{fig:Schematic_diagram}a, during generator optimization stage, the depth of the computational graph  increases linearly with 
$N$ owing to the few-step inference process, thereby compromising training stability and elevating memory overhead.
To mitigate this challenge, DMD2~\citep{yin2024DMD2} implemented a stochastic gradient truncation strategy (SGTS), as illustrated in Fig.~\ref{fig:Schematic_diagram}b. Although this strategy was not explicitly named in DMD2, it was later referred to as such in a following work~\citep{huang2025_self_forcing}. 
Under this approach, an index $j$ is sampled uniformly at random from $\{1, 2, ..., N\}$ and the corresponding timestep $t_{j}$ is set to $0$.
The sampling pipeline is then executed solely for the steps $i = 0, 1, ..., j - 1$.
Crucially, when $j=1$, the training iteration collapses to a one-step distillation.
Consequently, although SGTS enhances memory efficiency and training stability, it compromises generative diversity, impairs prompt-following capability and yields slower video motion.
\subsection{Phased DMD}
In contrast to DMD2, which may degenerate into one-step distillation during certain iterations and thereby diminish effective model capacity, \ourmodel circumvents this limitation by partitioning the distillation process into distinct phases and imposing supervision at intermediate timesteps. 
In every phase except the final one, the generator is optimized to minimize the reverse KL divergence at an intermediate timestep, while the fake diffusion model is updated through score matching over a subinterval of the diffusion process. 
\subsubsection{Distribution Matching at Intermediate Timesteps}
\label{subsubsec:distribution_matching_at_Intermediate_timesteps}
The rationale underlying \ourmodel can be elucidated by revisiting Eq.~\ref{eq:reverse_KL_noised_v_theta_gradient_fewstep}.
To sample $x_t$, prior methods~\cite{yin2024_DMD1,yin2024DMD2,huang2025_self_forcing} first generate $x_0$ and then diffuse it to $x_t$ according to Eq.~\ref{eq:p_xt_x0}.
In \ourmodel, the backward simulation is adapted to produce intermediate samples  $x_{t_k}$, \ie, $x_{t_k} = \operatorname{pipeline}(\boldsymbol{G}_{\boldsymbol{\phi_1}}, ..., \boldsymbol{G}_{\boldsymbol{\phi_k}}, \{t_1, t_2, ..., t_k\}, z, \mathcal{S} )$, where $0 < k \leq N$, rather than $x_0$. For brevity, the backward simulation is henceforth denoted as $x_{t_k} = \operatorname{pipeline_k(z)}$ in the subsequent discussion.
The sample $x_{t_k}$ is then diffused according to Eq.~\ref{eq:p_xt_xs},
with $s = t_k$, and $t$ is sampled from the subinterval $(t_k, 1)$, \ie, $t \sim \mathcal{T}(t_k, 1)$.
The generator optimization objective for the $k$-th phase is formulated as:
\begin{equation}
\nabla_{\boldsymbol{\phi_k}}D_{KL} \approx  E_{z, x_{t_k} = \operatorname{pipeline_k}(z), t, \boldsymbol{\epsilon}, x_t = \alpha_{t|t_k} x_{t_k} + \sigma_{t|t_k} \boldsymbol{\epsilon}} 
[w_{t|t_k} (\boldsymbol{T}_{\boldsymbol{\hat{\theta}}}(x_t, t) - \boldsymbol{F}_{\boldsymbol{\theta_i}}(x_t, t)) \frac{d\boldsymbol{G}}{d \boldsymbol{\phi_k}}]
\label{eq:reverse_KL_noised_v_theta_gradient_fewstep_phase_i}
\end{equation}
where $w_{t|t_k} = \frac{\alpha_t \alpha_{t|t_k}}{\alpha_t \sigma_t  +\sigma_t^2}$.
Empirically, we observe that sampling $t \sim \mathcal{T}(t_k, 1)$ rather than $t \sim \mathcal{T}(t_k, t_{k-1})$, aligns more effectively with the progressive architecture of \ourmodel and delivers superior performance. Please refer to the supplementary materials for an ablation study comparing the two sampling ranges.

As illustrated in Fig.~\ref{fig:Schematic_diagram}c, \ourmodel progressively distills the generator toward higher SNR levels.
In each phase $k$, only a single expert  ${\boldsymbol{G}_{\boldsymbol{\phi}_k}}$ is trained.
This expert maps the distribution $p(x_{t_{k-1}})$ to $p(x_{t_{k}})$.
At the beginning of each phase, the fake diffusion model $\boldsymbol{F}_{\boldsymbol{\theta_k}}$ is initialized from the pretrained teacher model $\boldsymbol{T}_{\boldsymbol{\hat{\theta}}}$ rather than from the model of the preceding phase, \ie, $\boldsymbol{F}_{\boldsymbol{\theta_{k-1}}}$.
%
% Although the resulting MoE generator requires more GPU memory than a single-network generator, the overhead is manageable for three reasons.
% % 
% First, an optimizer is required only for the $k$-th trainable expert.
% % 
% Second, this overhead can be substantially reduced using Low-Rank Adaptation (LoRA) \citep{hu2021_lora}.
% Specifically, all experts can share a common backbone network, with individual experts activated by switching their respective LoRA weights.
% % 
% Finally, \ourmodel can be combined with \sgts (as shown in Fig.~\ref{fig:Schematic_diagram}d), and the number of distillation phases can be less than the number of sampling steps.
% 
\subsubsection{Score Matching Within Subintervals}
\label{subsubsec:score_matching_within_subinterval}
A central challenge in \ourmodel is that clean samples $x_0$ are inaccessible in all phases except the final one. 
Consequently, the training objective for the fake diffusion model $\boldsymbol{F}_{\boldsymbol{\theta_k}}$, as presented in Eq.~\ref{eq:FlowMatchTarget_in_DMD}, becomes inapplicable.
To address this, we derive a training objective based on score matching over a subinterval, which yields unbiased score estimates.
Specifically, a diffusion model $\boldsymbol{\psi}$ trained over the full interval  $t \sim \mathcal{T}(0, 1)$ via Eq.~\ref{eq:FlowMatchTarget_1} should instead be optimized over the subinterval $t \sim \mathcal{T}(s, 1)$ using the following loss:
% derived from Eq.~\ref{eq:FlowMatchTarget_2}:
% 
\begin{equation}
E_{x_s, t, \boldsymbol{\epsilon}, x_t = \alpha_{t|s} x_s + \sigma_{t|s} \boldsymbol{\epsilon}}
[\|\boldsymbol{\psi}(x_t, t) - (\frac{\alpha_s^2 \sigma_t  + \alpha_t \sigma_s^2}{\alpha_s^2 \sigma_{t|s}}  \boldsymbol{\epsilon} - \frac{1}{\alpha_s} x_s ) \|^2] \label{eq:FlowMatchTarget_conditional}
\end{equation}
% 
% The first derivation is valid based on Eq.~\ref{eq:p_xt_three_equal} and the equivalence between DSM and ESM \citep{vincent2011_connection_DSM_ESM}.
% 
%
For x-prediction diffusion models, as employed in~\cite{karras2022_edm,zhou2024sid,luo2025TDM,huang2025_self_forcing}, the diffusion model $\boldsymbol{\mu}$ should be optimized over the subinterval $t \sim \mathcal{T}(s, 1)$ using the following loss:
\begin{equation}
E_{x_s, t, \boldsymbol{\epsilon}, x_t = \alpha_{t|s} x_s + \sigma_{t|s} \boldsymbol{\epsilon}}
[\|\boldsymbol{\mu}(x_t, t) - (\frac{1}{\alpha_s}  x_s - \frac{\alpha_t \sigma_s^2}{\alpha_s^2 \sigma_{t|s}} \boldsymbol{\epsilon} ) \|^2] \label{eq:xpredictTarget_conditional}
\end{equation}
Please refer to the supplementary materials for the detailed derivation.
As $\sigma_{t|s} \to 0$ when $t \to s$, the formulation in Eq.~\ref{eq:FlowMatchTarget_conditional} encounters singularity and numerical instability. To mitigate this, we apply a clamping function, resulting in the final loss term:
\begin{equation}
\operatorname{clamp}(\frac{1}{\sigma_{t|s}^2}) \|\sigma_{t|s} \boldsymbol{\psi}(x_t, t) - (\frac{\alpha_s^2 \sigma_t + \alpha_t \sigma_s^2}{\alpha_s^2} \boldsymbol{\epsilon} - \frac{\sigma_{t|s}}{ \alpha_s} x_s  ) \|^2
\label{eq:FlowMatchTarget_conditional_singularity}
\end{equation}
Here, $\operatorname{clamp}(\frac{1}{\sigma_{t|s}^2})$ restricts the value within a predefined range $[0, 10]$ to prevent overflow.
% 
% As proved in~\missref, optimizing flow models according to Eq.~\ref{eq:FlowMatchTarget_conditional_singularity} within the subinterval $t \sim \mathcal{T}(s, 1)$ is an \textbf{unbiased} estimation of scores, \ie, Eq.~\ref{eq:FlowMatchTarget_2} holds within $(s, 1)$.

\begin{figure}[tb]
  \centering

  \begin{subfigure}{0.30\linewidth}
    \includegraphics[width=\linewidth]{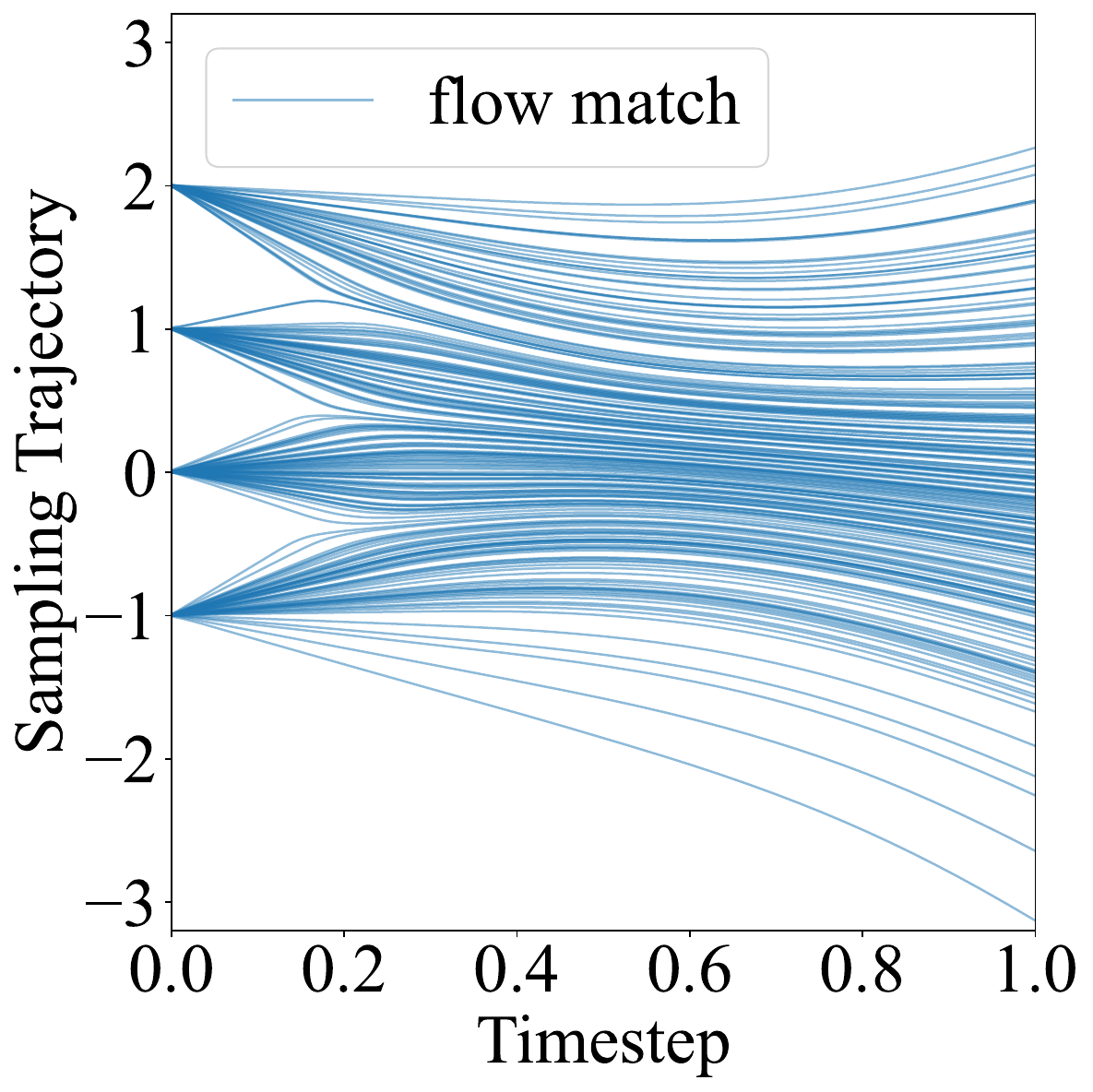}
    \caption{}
    \label{subfig:toy_experiment_flow_match}
  \end{subfigure}
  \begin{subfigure}{0.30\linewidth}
    \includegraphics[width=\linewidth]{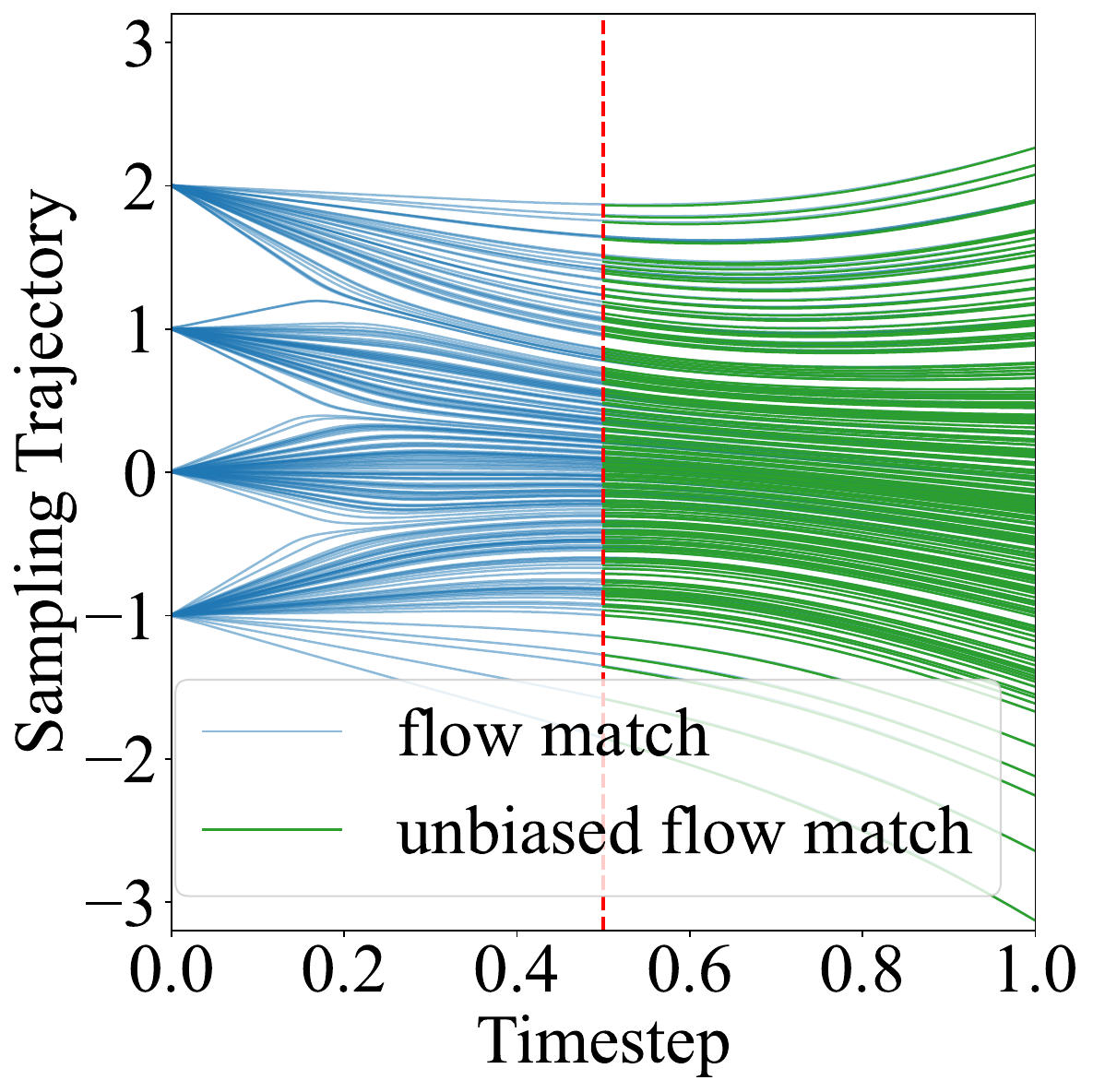}
    \caption{}
    \label{subfig:toy_experiment_unbiased_flow_match}
  \end{subfigure}
  \begin{subfigure}{0.30\linewidth}
    \includegraphics[width=\linewidth]{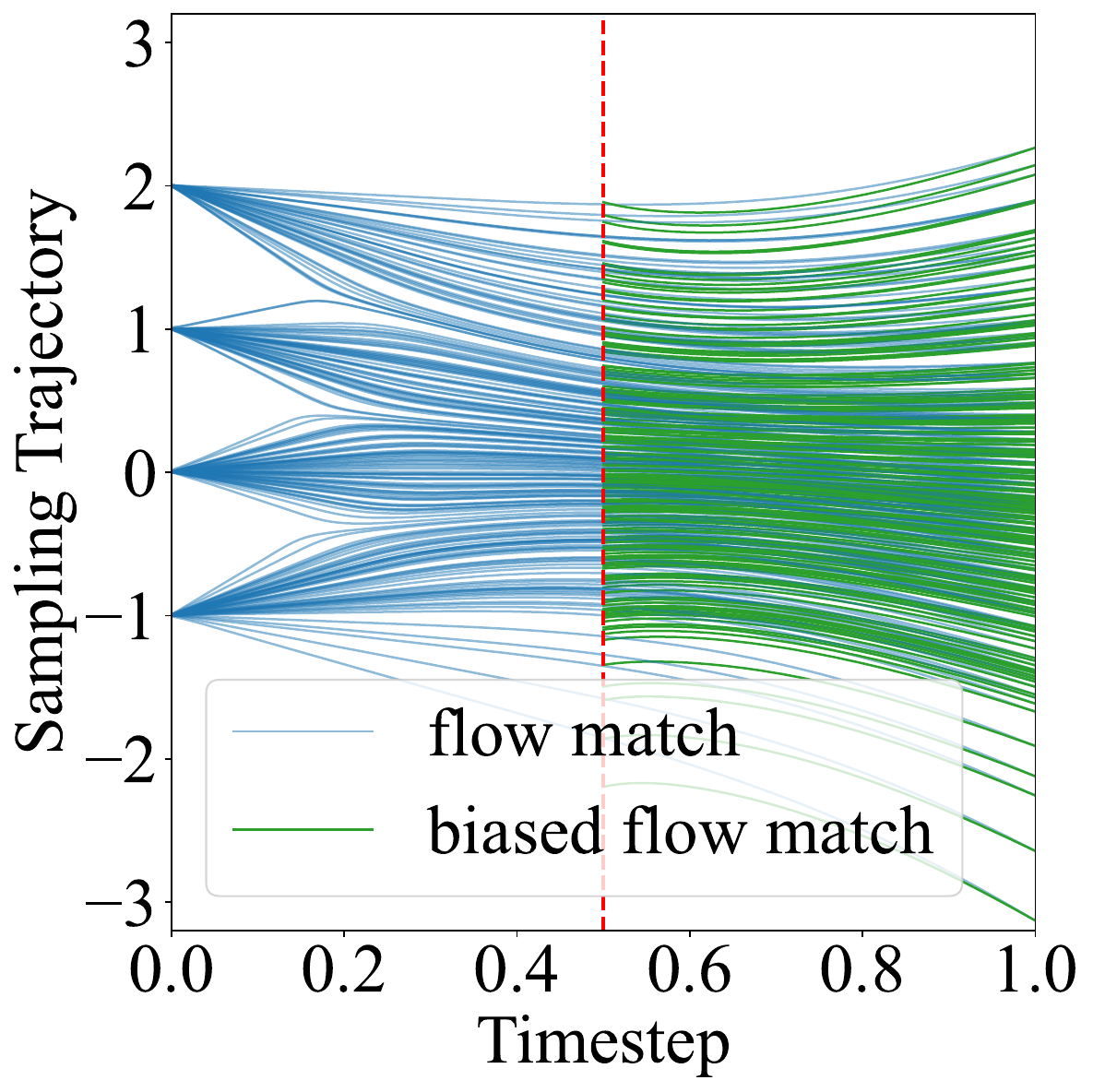}
    \caption{}
    \label{subfig:toy_experiment_biased_flow_match}
  \end{subfigure}
  \caption{
    Sampling trajectories of 200 samples in a 1D toy experiment.
    (a) Training with the full-interval objective (Eq.~\ref{eq:FlowMatchTarget_1}).
    (b) Training on  $0.5 < t < 1$ with the correct subinterval objective 
    (Eq.~\ref{eq:FlowMatchTarget_conditional_singularity}).
    (c) Training on  $0.5 < t < 1$ with an incorrect target: $\|(\boldsymbol{\psi}(x_t, t) - (\boldsymbol{\epsilon} - x_s ) \|^2$.
  }
  \label{fig:toy_experiment}
\end{figure}

We design a one-dimensional  toy experiment to validate the efficacy of this training objective, wherein $x_0$ takes only four discrete values: \{-1, 0, 1, 2\}, as shown in Fig.~\ref{fig:toy_experiment}. 
The diffusion model is parameterized by a four-layer MLP with hidden dimension 512. Three models are trained from scratch: (a) one using loss Eq.~\ref{eq:FlowMatchTarget_1} over the full interval $(0, 1]$; (b) another using loss Eq.~\ref{eq:FlowMatchTarget_conditional_singularity} over the subinterval $(0.5, 1]$; and (c) a third using loss term $\|\boldsymbol{\psi}(x_t, t) - (\boldsymbol{\epsilon} - x_s ) \|^2$ over the same subinterval.
Following training, trajectories are sampled using an Euler solver with 100 steps.
The close overlap of the sampling trajectories in Fig.~\ref{subfig:toy_experiment_unbiased_flow_match} confirms that, 
within the specified subinterval, the flow model trained with Eq.~\ref{eq:FlowMatchTarget_conditional_singularity} is equivalent to one trained with the standard objective in Eq.~\ref{eq:FlowMatchTarget_1}.
In contrast, Fig.~\ref{subfig:toy_experiment_biased_flow_match} illustrates how an incorrect objective formulation induces biased score estimation, thereby violating the \textbf{A2} assumption.
%
% Refer detailed settings of toy example to Appendix~\ref{sec:appendix_toy_example}.
% 

In the $k$-th phase of Phased DMD, the fake diffusion model $\boldsymbol{F}_{\boldsymbol{\theta_k}}$ is optimized over the subinterval $(t_k, 1]$ using Eq.~\ref{eq:FlowMatchTarget_conditional_singularity}. 
As established in the foregoing analysis,
% According to the analysis above, 
$\boldsymbol{F}_{\boldsymbol{\theta_k}}$ satisfies the \textbf{A2} assumption. Following DMD2~\cite{yin2024DMD2}, we perform 5 updates to $\boldsymbol{F}_{\boldsymbol{\theta_k}}$ for each update to ${\boldsymbol{G}_{\boldsymbol{\phi}_k}}$.

\section{Experiments and Results}
\label{sec:exps}
To evaluate the efficacy of \ourmodel, 
we conduct experiments across text-to-image (T2I), text-to-video (T2V) and image-to-video (I2V) generation tasks.
SOTA base models are employed as teachers, including Wan2.1-T2V-14B, Wan2.2-T2V-A14B, Wan2.2-I2V-A14B and Qwen-Image-20B. An overview of the experimental configurations is presented in Tab.~\ref{tab:experiment_summary}. 
Due to its substantial computational demands, the vanilla few-step DMD is applied only to the smallest configuration, namely, T2I task using the Wan2.1-T2V-14B base model.

We conduct experiments on 64 GPUs, employing PyTorch FSDP and gradient checkpointing to reduce GPU memory consumption. Context parallelism is applied for T2V and I2V distillation.
The following settings are used consistently across all experiments: a batch size of 64; a fake diffusion model learning rate of $4 \times 10^{-7}$ with full-parameter training; a generator learning rate of $5 \times 10^{-5}$ using LoRA~\cite{hu2021_lora} with $rank=64$ and $\alpha = 8$. Following prior works~\cite{huang2025_self_forcing}, AdamW optimizer is employed for both the fake diffusion model and the generator, with hyperparameter $\beta_1 = 0, \beta_2 = 0.999$. The fake diffusion model is updated 5 times for each generator update.  Euler solver is used in backward simulation, due to its simplicity.
% We apply \ourmodel to state-of-the-art (SOTA) image and video generative models.
% 
To align with the two-expert architecture of Wan2.2, we adopt a 4-step, 2-phase configuration for \ourmodel, as illustrated in Fig.~\ref{fig:Schematic_diagram}d.
Consequently, each base model is distilled into two expert networks.

To demonstrate that the performance gains arise primarily from our novel distillation paradigm rather than from an increase in trainable parameters, Wan2.2-T2V-A14B is distilled for both T2I and T2V tasks.  Wan2.2-T2V-A14B already incorporates an MoE architecture, and both standard DMD and our Phased DMD distill it into two experts, thereby enabling a direct comparison under equivalent parameter budgets.
% 
% Cosine Similarity is computed by DinoV3 \citep{simeoni2025dinov3}
\begin{table}[tbh]
\caption{Overview of the experimental setup. ``Vanilla DMD'' refers to vanilla few-step DMD shown in Fig.~\ref{fig:Schematic_diagram}a. ``Ours'' refers to \ourmodel. Some experiments employ mixed data resolutions. The reported values for Frame, Height and Width represent one resolution.}
\label{tab:experiment_summary}
\centering
\begin{tabular}{c c c c c c c }
\toprule
% \multirow{3}{*}{\bf{Model}} & \multirow{3}{*}{\bf{Task}} & \multirow{3}{*}{\bf{DMD}} & \bf{DMD} & \bf{Phased}\\
% & & & \bf{with} &  \bf{DMD}\\
% & & &  \bf{SGTS} &  \bf{(Ours)}\\
\bf{Base Model} & \bf{Task} & \bf{Vanilla DMD} & \bf{DMD2} & \bf{Ours} & \bf{Timesteps} & \bf{Frame, Height, Width} \\
\midrule
Wan2.1-T2V-14B & T2I  & $\checkmark$  & $\checkmark$ & $\checkmark$ & 1000, 938, 833, 625 & 1, 720, 1280 \\
Wan2.2-T2V-A14B & T2I  &  $\times$ & $\checkmark$ & $\checkmark$ & 1000, 938, 833, 625 & 1, 720, 1280  \\
Wan2.2-T2V-A14B & T2V  &  $\times$ & $\checkmark$ & $\checkmark$ & 1000, 938, 833, 625 & 81, 720, 1280  \\
Wan2.2-I2V-A14B & I2V  &  $\times$ & $\checkmark$ & $\checkmark$ & 1000, 938, 833, 625 & 81, 720, 1280  \\
Qwen-Image-20B & T2I  &  $\times$ & $\checkmark$ & $\checkmark$ & 1000, 900, 750, 500 & 1, 1382, 1382  \\
\bottomrule
\end{tabular}
\end{table}

% Cosine Similarity is computed by DinoV3 \citep{simeoni2025dinov3}
% \begin{table}[b]
% \caption{
% Comparison of motion dynamics preservation across distillation methods, measured by mean absolute optical flow and VBench dynamic degree. Phased DMD demonstrates superior performance in maintaining the base model's motion quality.}
% \begin{center}
% \begin{tabular}{ccc}
% \toprule
% \bf{Method}  & \bf{Optical Flow} $ \uparrow $ & \bf{Dynamic Degree} $ \uparrow $\\
% \midrule
% Base model & $10.66$  & 79.35 \% \\
% \midrule
% DMD with SGTS & $5.27 $  &  72.90 \% \\
% Phased DMD(Ours) & $\textbf{7.76}$ & \textbf{78.71 \%} \\
% \bottomrule
% \label{tab:motion_speed_quant}
% \end{tabular}
% \end{center}
% \end{table}

\begin{table}[h]
% \footnotesize
\caption{
Quantitative comparison of video generation performance. 
``OF'' refers to optical flow. ``DD'' refers to dynamic degree.
\ourmodel outperforms DMD2 in 
motion dynamics (OF, DD) and video quality (FID, FVD) on both T2V and I2V tasks.
% optical flow (OF), danamic degree (DG), FID and FVD. Phased DMD demonstrates superior performance over DMD with SGTS in motion dynamics and video quality on both T2V and I2V tasks.
}
\centering
\begin{tabular}{c c c c c c c c c}
\toprule
\multirow{2}{*}{\bf{Method}} & \multicolumn{4}{c}{\bf{T2V}} & \multicolumn{4}{c}{\bf{I2V}} \\
\cmidrule(lr){2-5} \cmidrule(lr){6-9}
& \bf{OF} $ \uparrow $ & \bf{DD} $ \uparrow $ & \bf{FID} $ \downarrow $ & \bf{FVD} $ \downarrow $  & \bf{OF} $ \uparrow $ & \bf{DD} $ \uparrow $ & \bf{FID} $ \downarrow $ & \bf{FVD} $ \downarrow $ \\
\midrule
Base model & \bf{10.26} & \underline{79.55 \%} & \bf{0.0} & \bf{0.0} & \underline{9.32} & \underline{82.27 \%}  & \bf{0.0} & \bf{0.0} \\
\midrule
DMD2 & $3.23 $ & 65.45 \%  & 55.70 & 763.1 & 7.87 & 80.00 \%  & 18.45 & 370.0 \\
Phased DMD (Ours) & $\underline{9.30}$ & \bf{82.27 \%}  & \underline{47.24} & \underline{700.9} & \textbf{9.84} & \bf{83.64 \%}  & \underline{17.47} & \underline{334.7} \\
\bottomrule
\label{tab:motion_speed_quant}
\end{tabular}
\end{table}

\subsection{Phased DMD in Video Generation}
\begin{figure}[htbp]
    \centering
    \begin{minipage}{0.48\textwidth}
        \centering
    \begin{subfigure}[b]{0.32\textwidth}
      \centering
      \includegraphics[width=\textwidth]{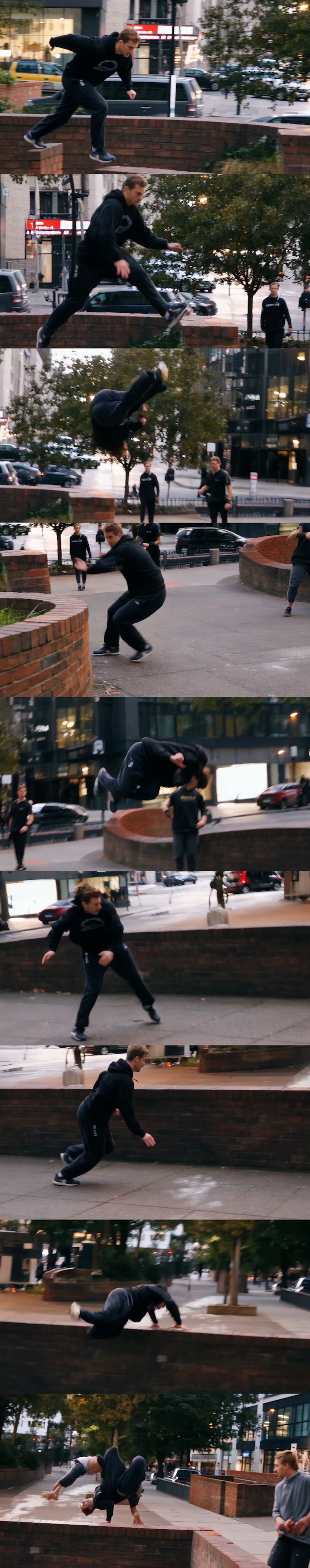}
      \caption{base}
      \label{subfig:video_speed_base}
    \end{subfigure}
    \hfill
    \begin{subfigure}[b]{0.32\textwidth}
      \centering
      \includegraphics[width=\textwidth]{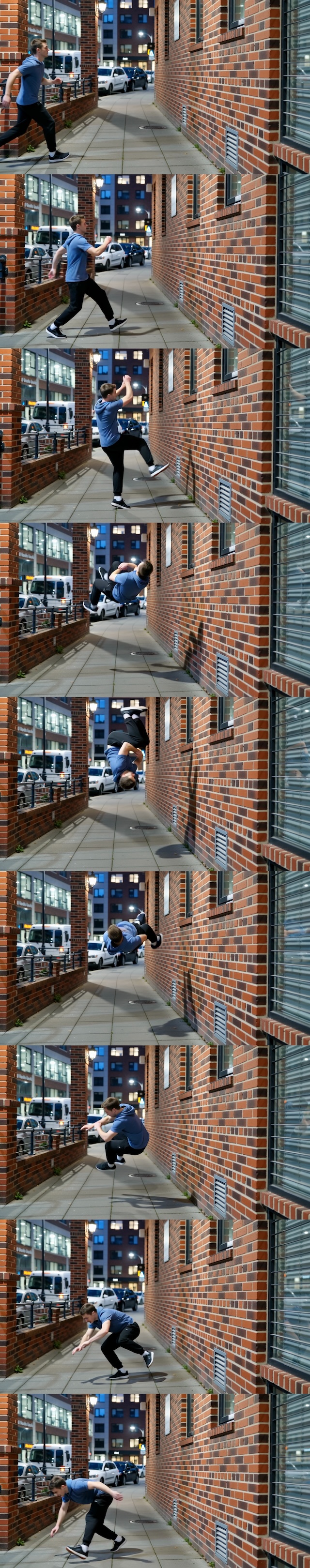}
      \caption{DMD2}
      \label{subfig:video_speed_DMD_wi_SGTS}
    \end{subfigure}
    \hfill
    \begin{subfigure}[b]{0.32\textwidth}
      \centering
      \includegraphics[width=\textwidth]{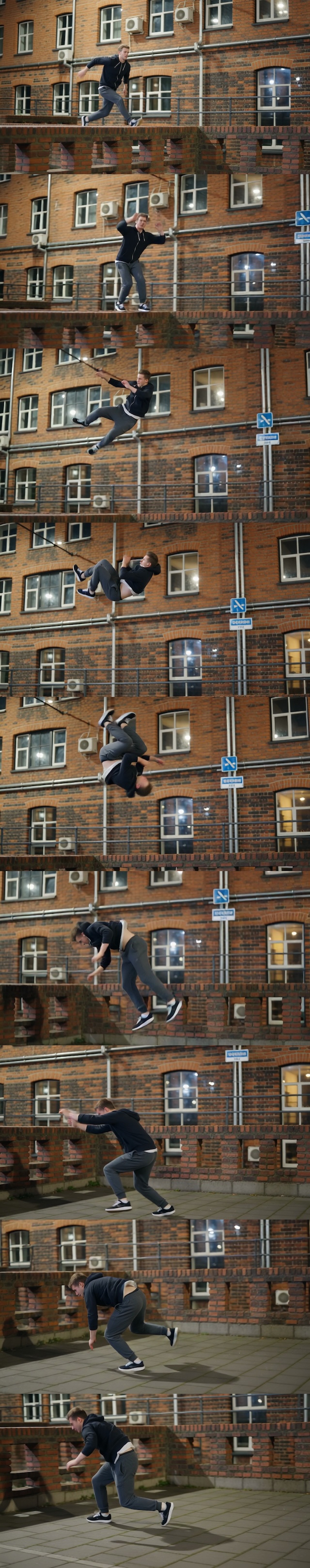}
      \caption{\ourmodel}
      \label{subfig:video_speed_phased_dmd}
    \end{subfigure}
    \captionof{figure}{
        Motion dynamics comparison of video frames generated by (a) Wan2.2-T2V-A14B base model and its distilled versions using (b) DMD2 and (c) Phased DMD. 
        Each video consists of 81 frames and frames with indices $\{0, 10, ..., 80\}$ are combined as a preview.
        A portion of the full prompt: ``A parkour athlete swiftly runs horizontally along a brick wall in an urban setting. Pushing off powerfully with one foot, they launch themselves explosively into a twisting front flip. ''
    }\label{fig:video_speed}
    \end{minipage}
    \hfill
    \begin{minipage}{0.48\textwidth}
        \centering
    \begin{subfigure}[b]{0.32\textwidth}
      \centering
      \includegraphics[width=\textwidth]{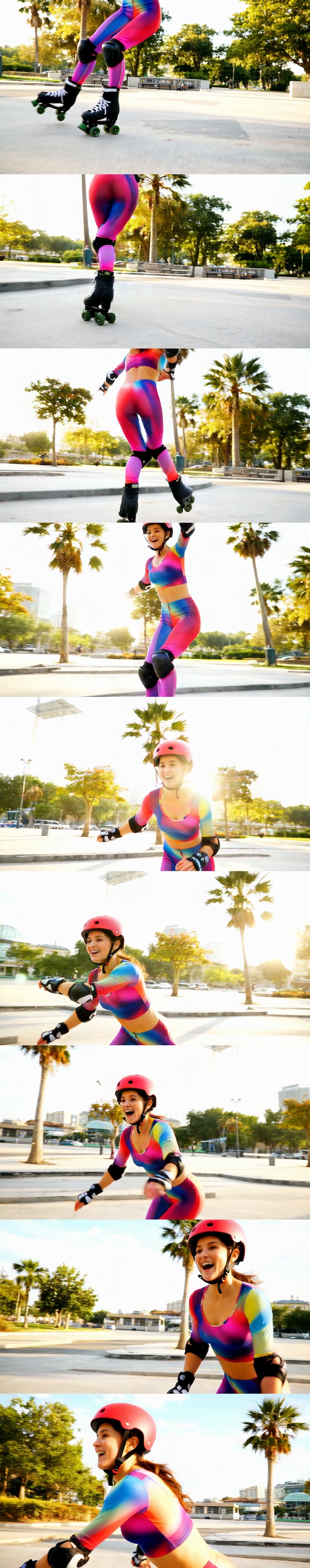}
      \caption{base}
      \label{subfig:video_camera_base}
    \end{subfigure}
    \hfill
    \begin{subfigure}[b]{0.32\textwidth}
      \centering
      \includegraphics[width=\textwidth]{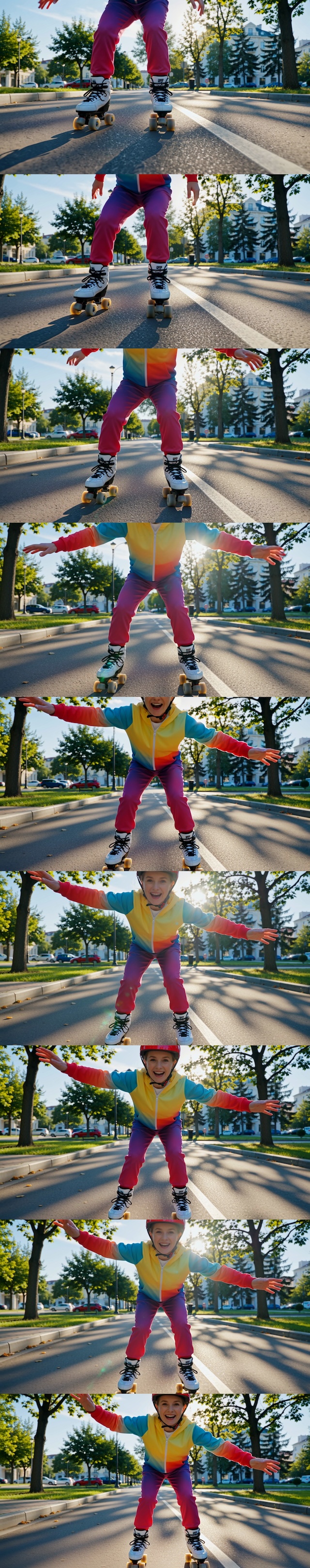}
      \caption{DMD2}
      \label{subfig:video_camera_DMD_wi_SGTS}
    \end{subfigure}
    \hfill
    \begin{subfigure}[b]{0.32\textwidth}
      \centering
      \includegraphics[width=\textwidth]{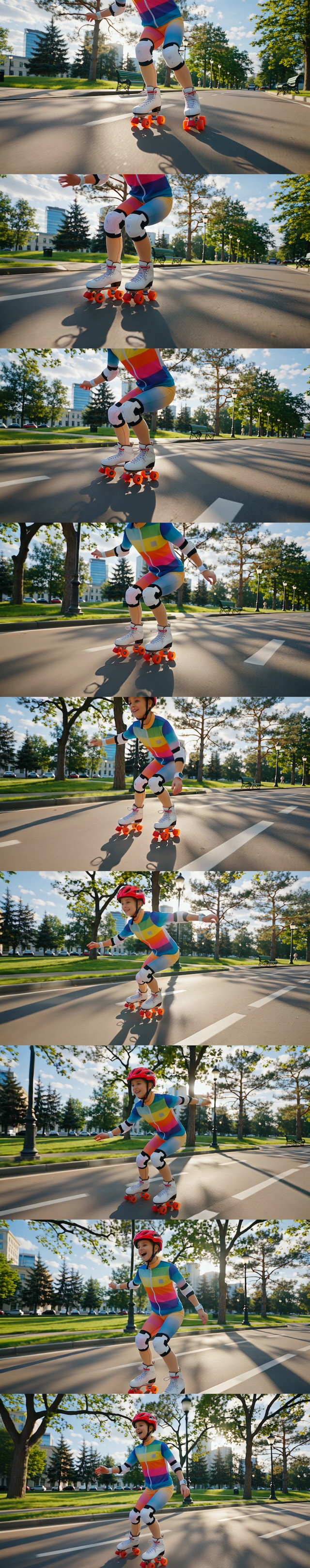}
      \caption{\ourmodel}
      \label{subfig:video_camera_phased_dmd}
    \end{subfigure}
        \captionof{figure}{
            Camera following capability comparison of video frames generated by (a) Wan2.2-T2V-A14B base model and its distilled versions using (b) DMD2 and (c) Phased DMD. 
            Each video consists of 81 frames and frames with indices $\{0, 10, ..., 80\}$ are combined as a preview.
            A portion of the full prompt: ``The camera starts focused on the skates carving sharp turns on the pavement and tilts up to reveal their entire body leaning into the motion.''
        }\label{fig:camera_control}
    \end{minipage}
\end{figure}

Wan2.2 video generation models exhibit remarkable proficiency in motion dynamics and camera control.
% 
% Nevertheless, we find that DMD2 impairs these qualities, due to the potential for one-step degeneration.
However, we observe that DMD2 tends to degrade these qualities, due to the one-step degeneration inherent in SGTS.
\ourmodel intrinsically addresses this limitation by explicitly eliminating dependence on $x_0$ in intermediate phases.
In the first phase, only the low-SNR expert is activated. 
As the pretrained low-SNR expert in Wan2.2 is already trained on the low-SNR subinterval, this correspondence better preserves its inherent capabilities.
Motion quality is evaluated using 220 text prompts for T2V and 220 image-prompt pairs for I2V, with one video generated per prompt using a fixed seed of 42. The base model is sampled with 40 steps and a guidance scale of 4, while the distilled models utilize 4 steps and a guidance scale of 1.
% 
% We evaluate motion quality using a set of 220 text prompts for T2V and 220 image-prompt pairs for I2V, generating one video per prompt with a fixed seed $42$.
% 
% The base model was sampled with 40 steps and guidance of 4, and the distilled models used 4 steps and guidance of 1.
% 
As illustrated in Fig.~\ref{fig:video_speed}, DMD2 produces slower motion dynamics compared to both the base model and \ourmodel. Similarly,  Fig.~\ref{fig:camera_control} reveals that DMD2 tends to generate close-up views, whereas \ourmodel and the base model more faithfully adhere to the prompt-specified camera instructions.

To quantitatively evaluate our method, motion intensity is quantified using mean absolute optical flow computed with UniMatch~\cite{xu2023unifying} and dynamic degree metric from VBench~\cite{huang2024vbench}.
Video visual quality is assessed via FID and FVD~\cite{skorokhodov2022stylegan_v} computed between the distilled models and the base models.
% 
% To quantitatively compare \ourmodel with DMD with SGTS
% Motion intensity is quantified using the \textit{mean absolute optical flow} computed with Unimatch \citep{xu2023unifying} and the \textit{dynamic degree} metric from VBench \citep{huang2024vbench}. 
% Video quality is quantified using \textit{FID} and \textit{FVD}~\cite{skorokhodov2022stylegan_v} between the distilled models and base models.
% 
As Tab.~\ref{tab:motion_speed_quant} shows, \ourmodel yields significantly stronger motion dynamics than DMD2, confirming its superior capacity to preserve the motion dynamics of the base model. Furthermore, the lower FID and FVD values indicate that \ourmodel better maintains the generative quality.
We also report results from additional VBench metrics. 
However, these metrics are less convincing, as they consistently rank the base model lowest, which contradicts human assessment.
Further comparative videos and quantitative analyses are provided in the supplementary materials.

\subsection{\ourmodel in Image Generation}
\begin{figure}[htbp]
    \centering
  seed 0
  \begin{subfigure}{0.11\linewidth}
    \includegraphics[width=\linewidth]{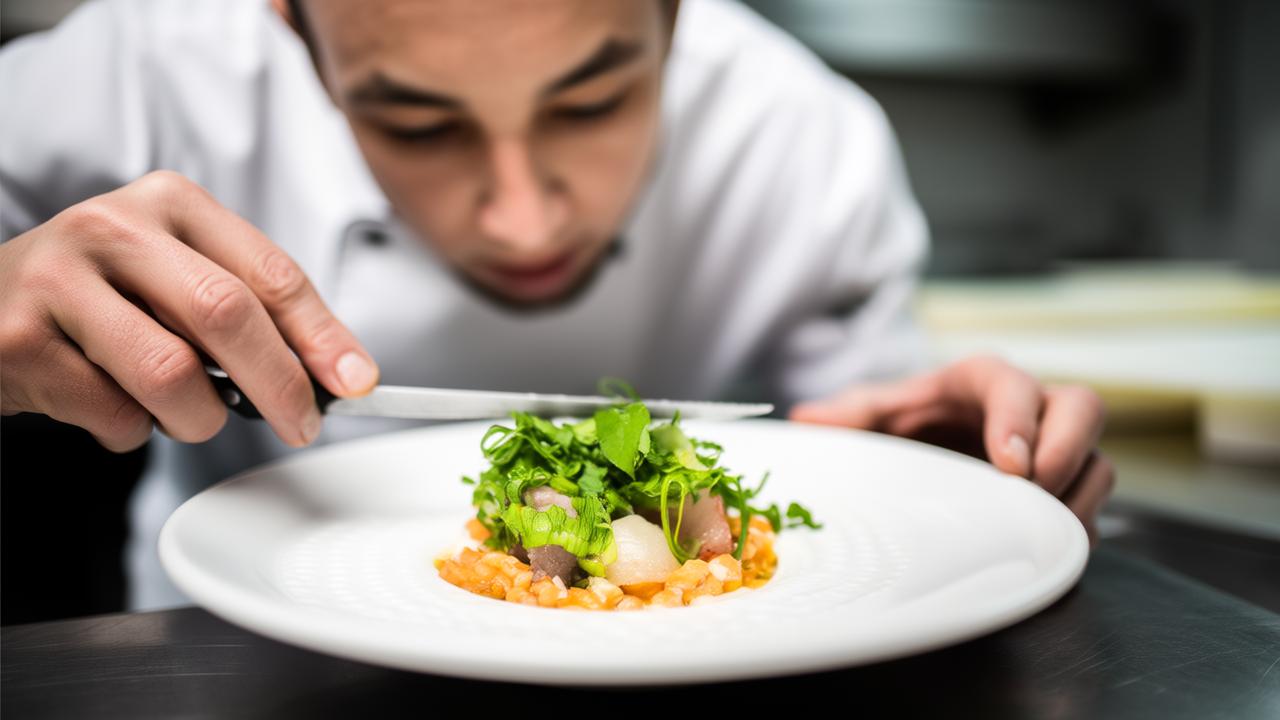}
  \end{subfigure}
  \begin{subfigure}{0.11\linewidth}
    \includegraphics[width=\linewidth]{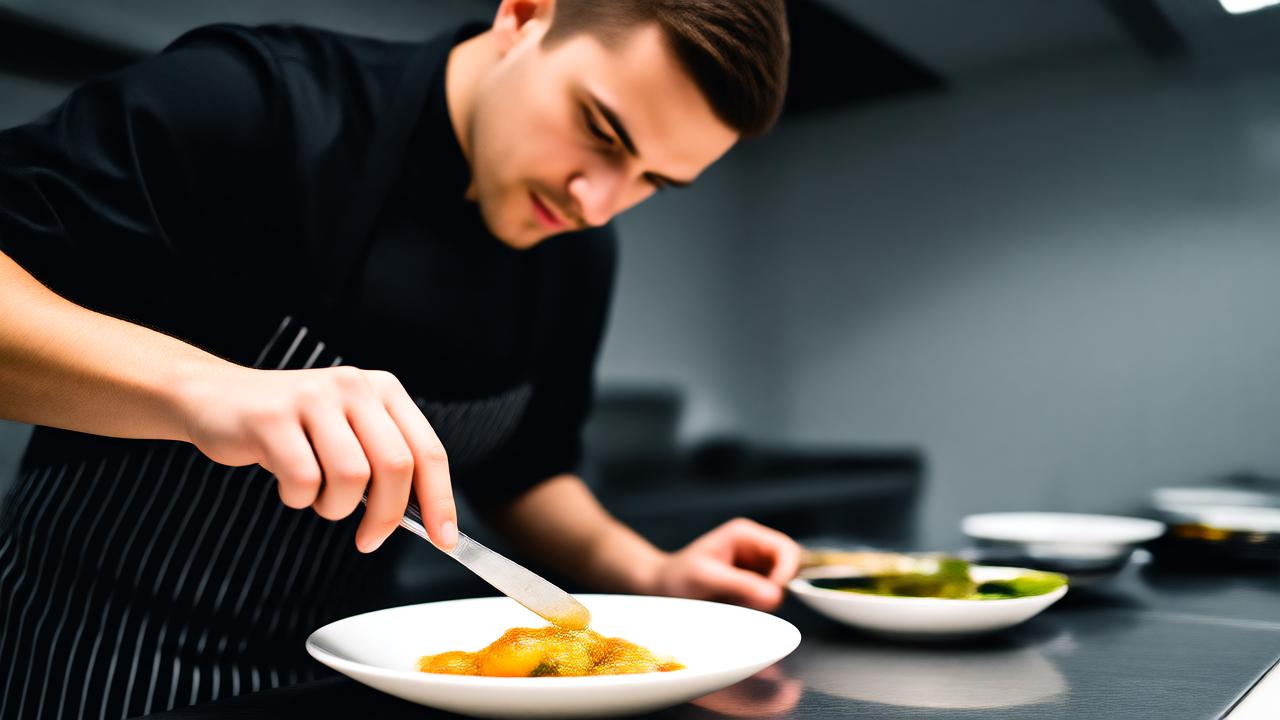}
  \end{subfigure}
  \begin{subfigure}{0.11\linewidth}
    \includegraphics[width=\linewidth]{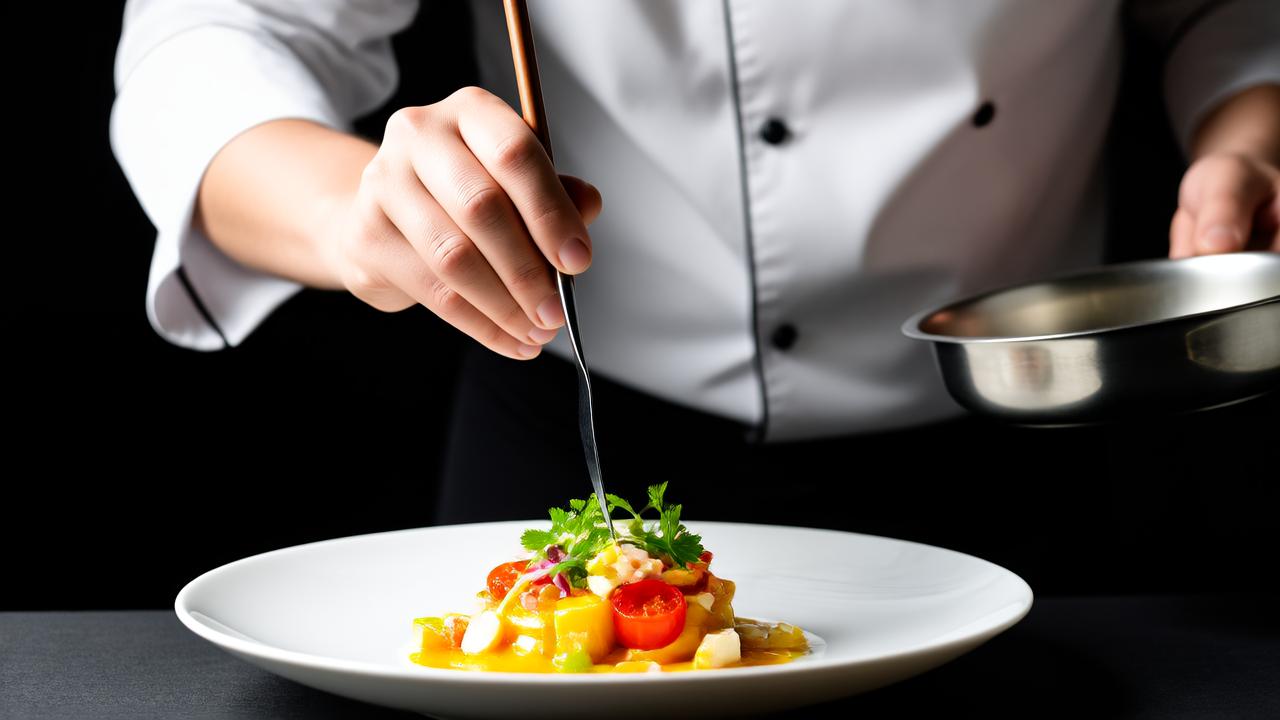}
  \end{subfigure}
  \begin{subfigure}{0.11\linewidth}
    \includegraphics[width=\linewidth]{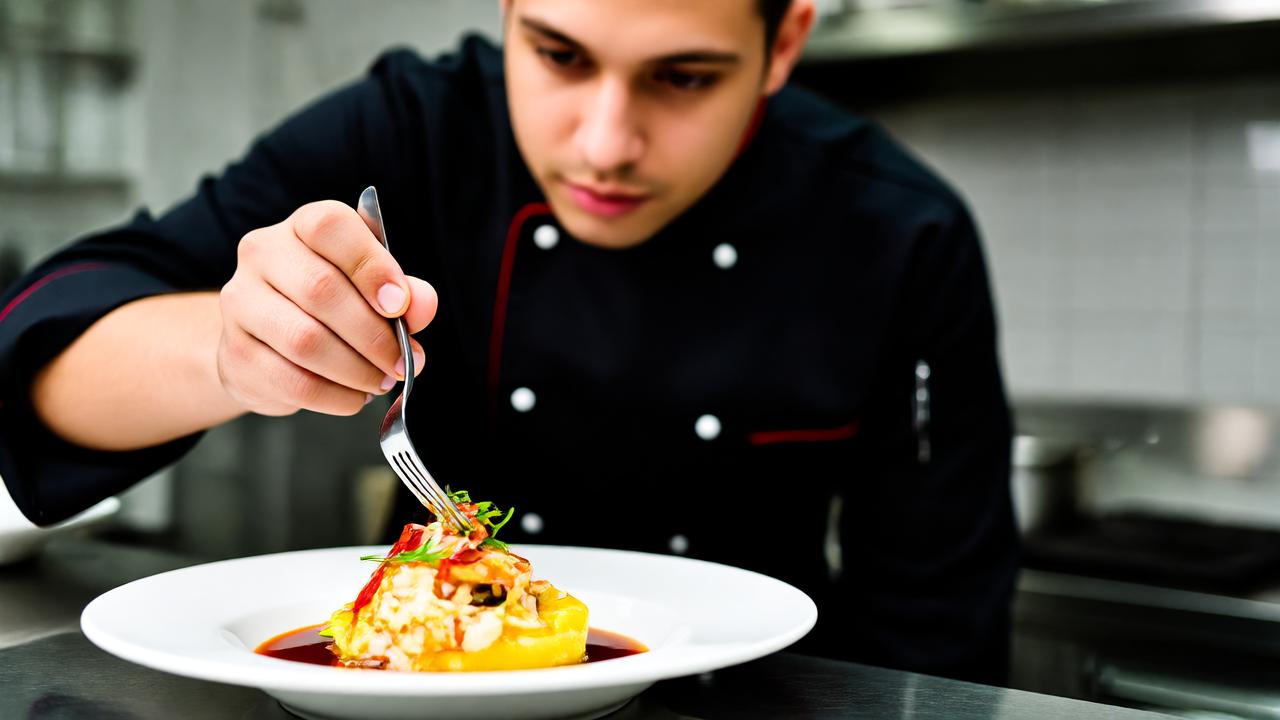}
  \end{subfigure}
  \begin{subfigure}{0.11\linewidth}
    \includegraphics[width=\linewidth]{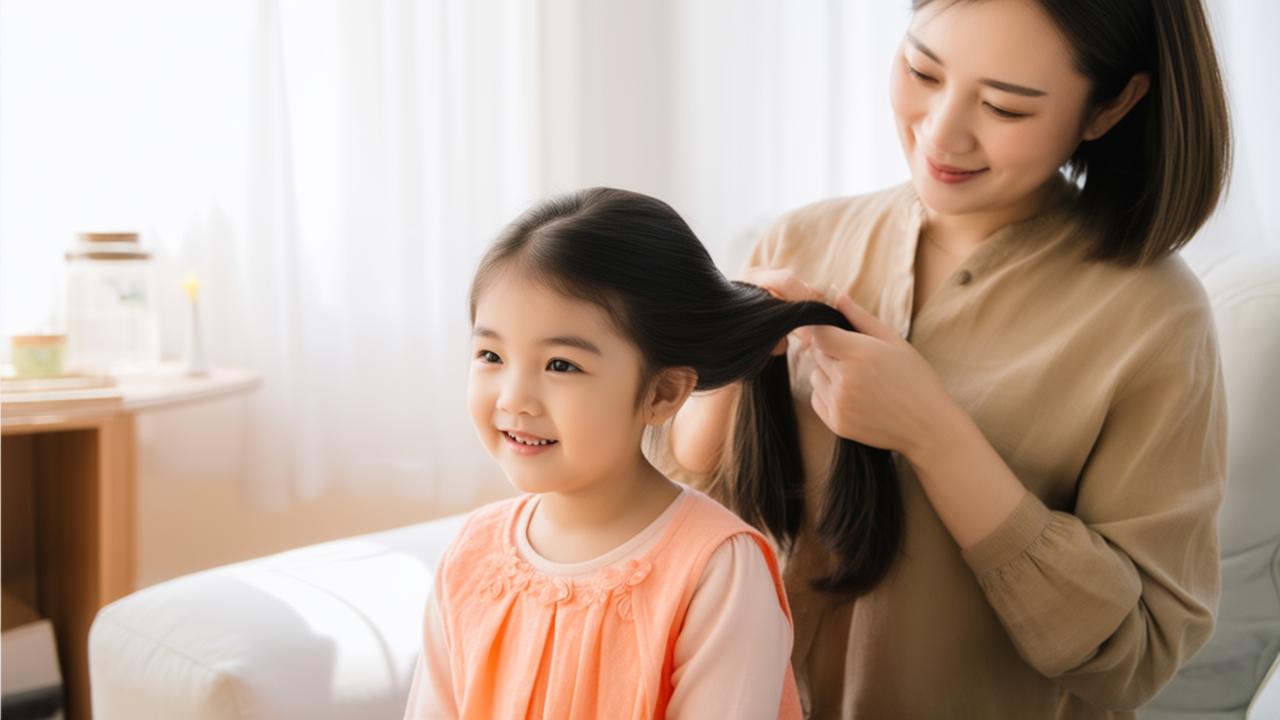}
  \end{subfigure}
  \begin{subfigure}{0.11\linewidth}
    \includegraphics[width=\linewidth]{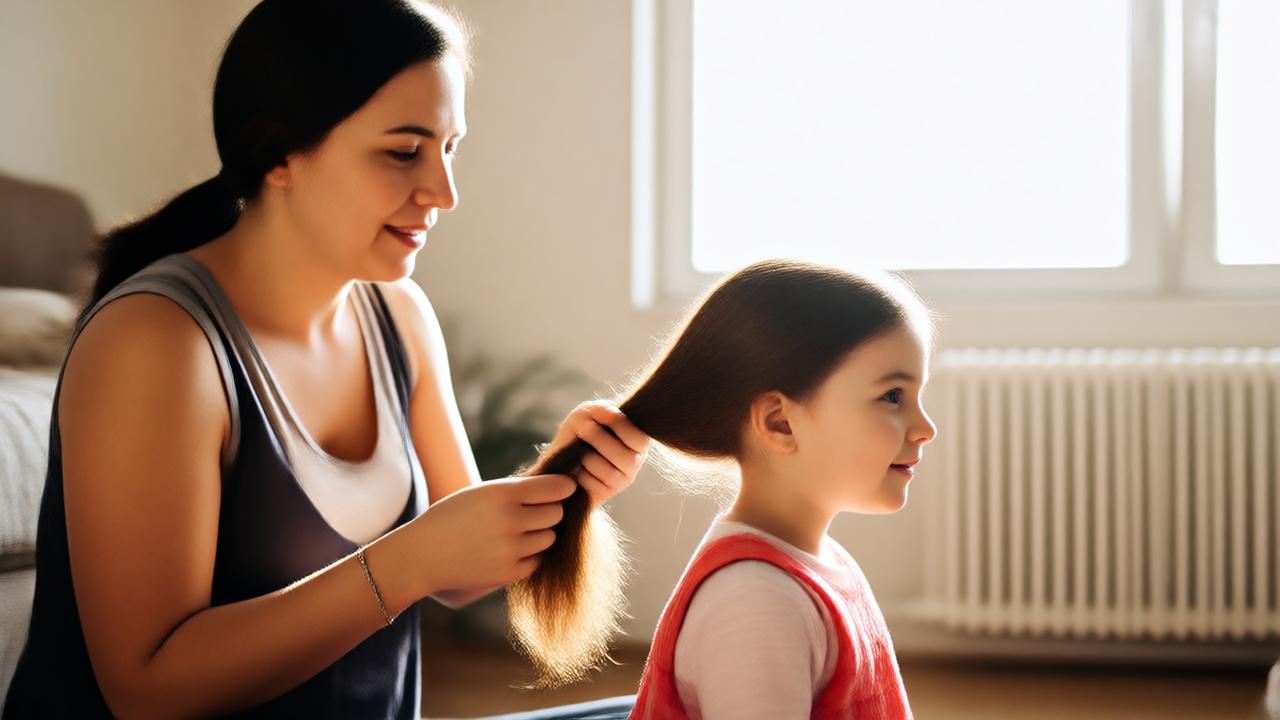}
  \end{subfigure}
  \begin{subfigure}{0.11\linewidth}
    \includegraphics[width=\linewidth]{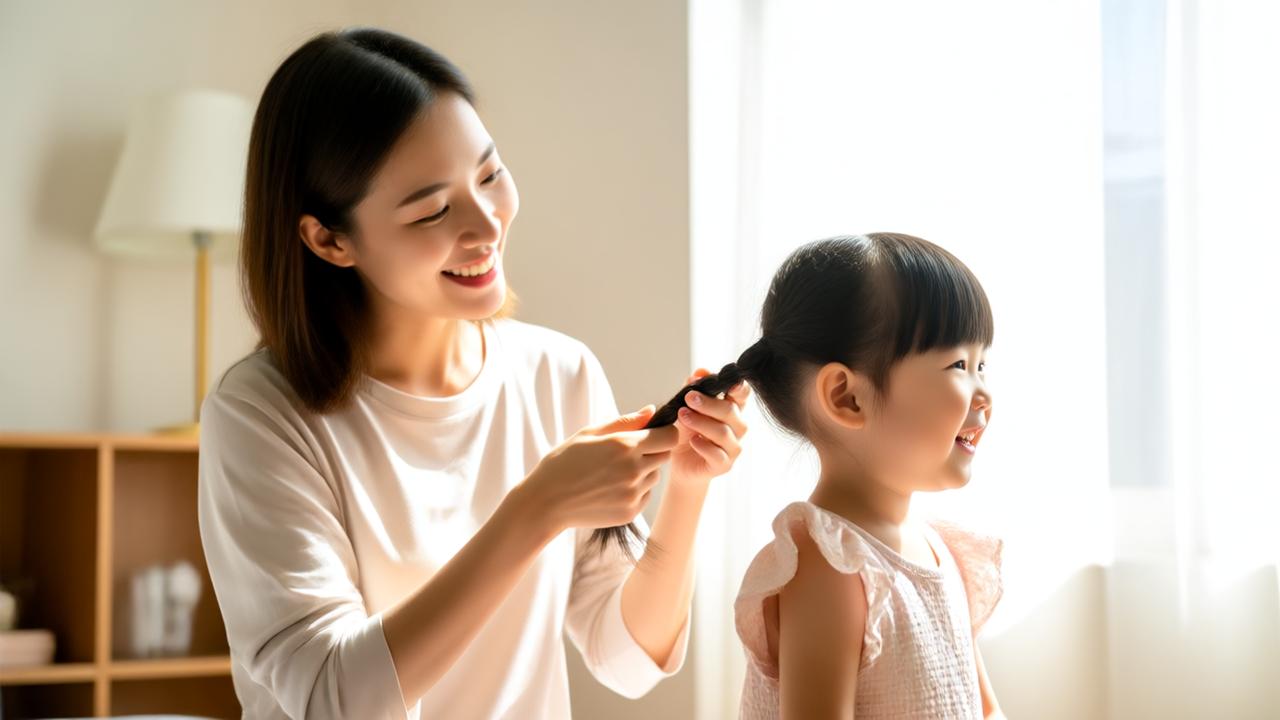}
  \end{subfigure}
  \begin{subfigure}{0.11\linewidth}
    \includegraphics[width=\linewidth]{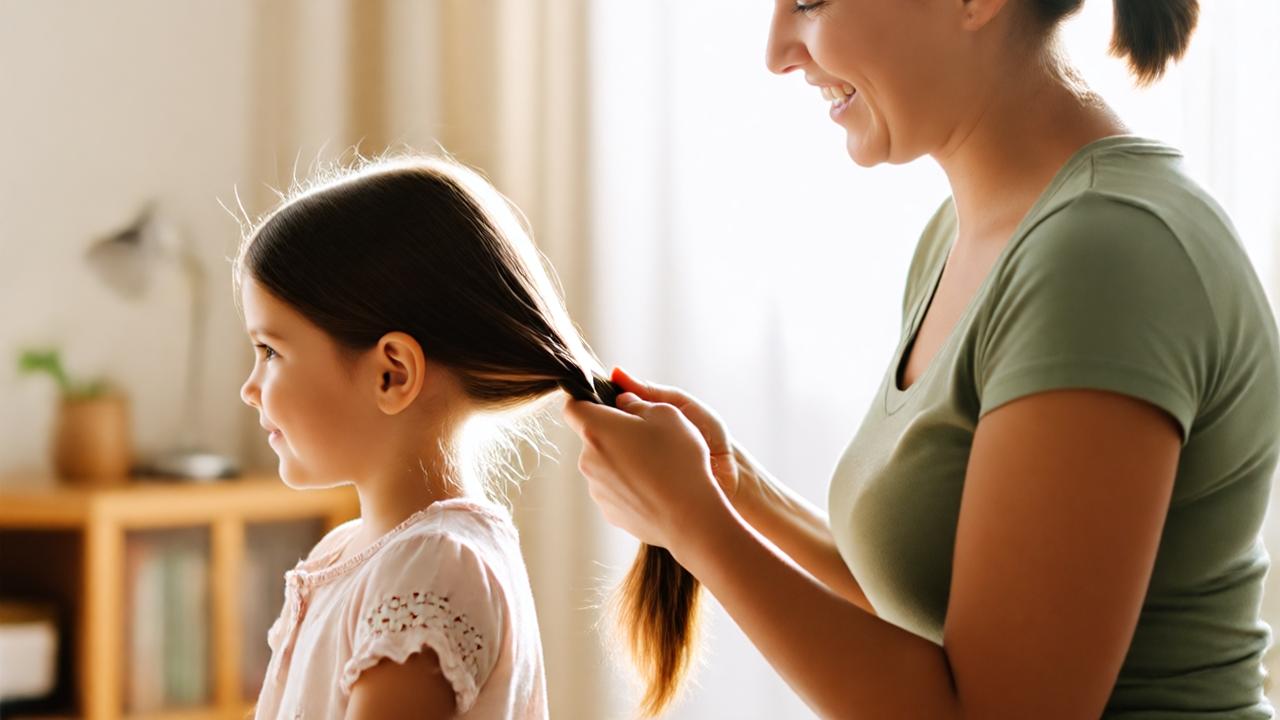}
  \end{subfigure}

  seed 1
  \begin{subfigure}{0.11\linewidth}
    \includegraphics[width=\linewidth]{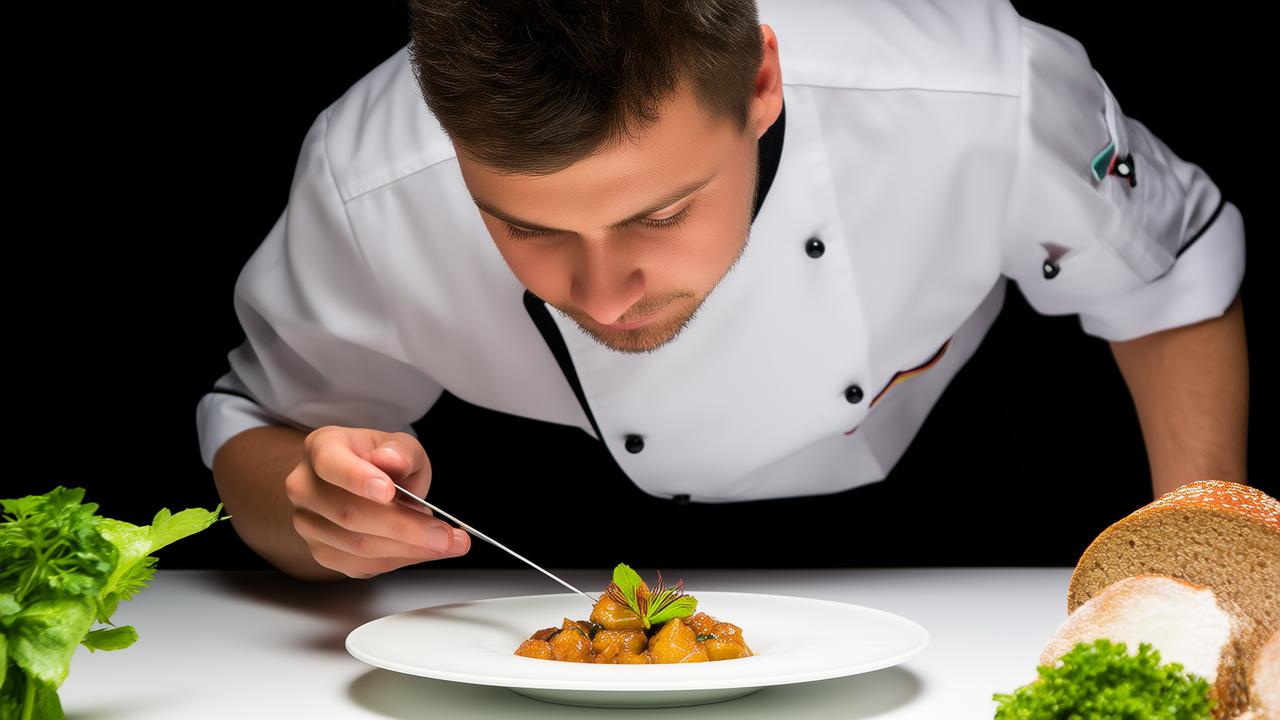}
  \end{subfigure}
  \begin{subfigure}{0.11\linewidth}
    \includegraphics[width=\linewidth]{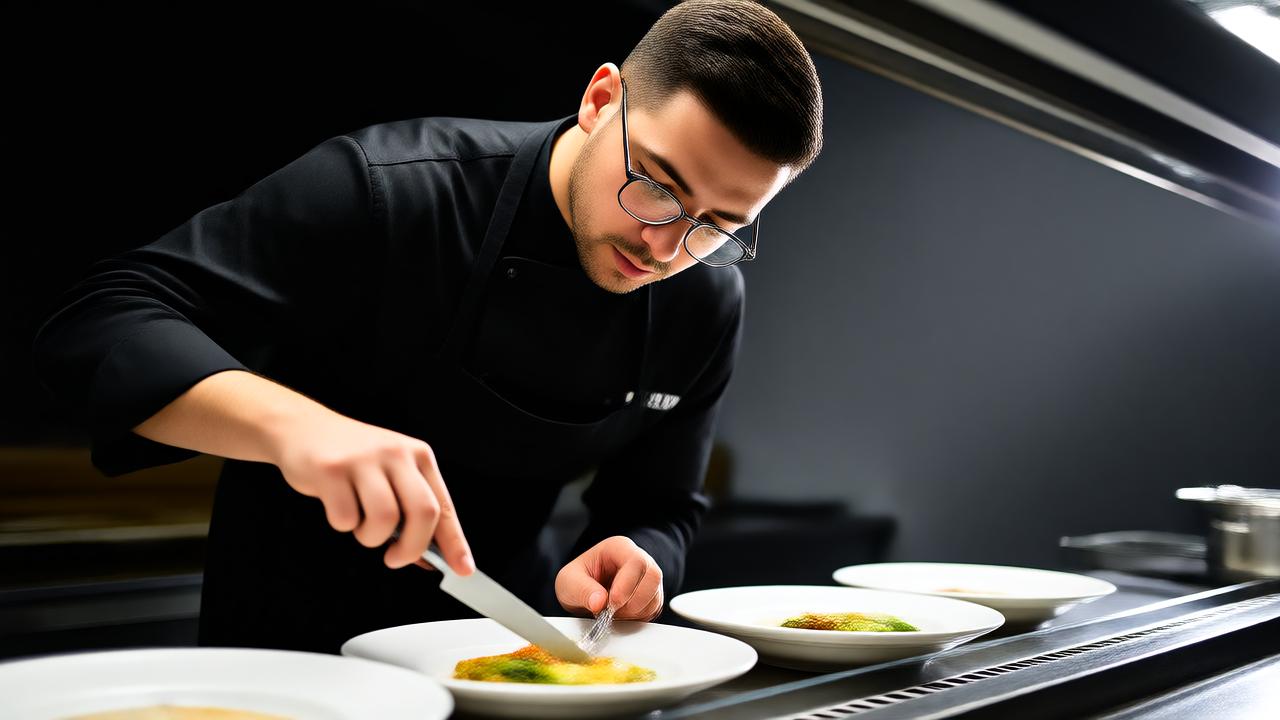}
  \end{subfigure}
  \begin{subfigure}{0.11\linewidth}
    \includegraphics[width=\linewidth]{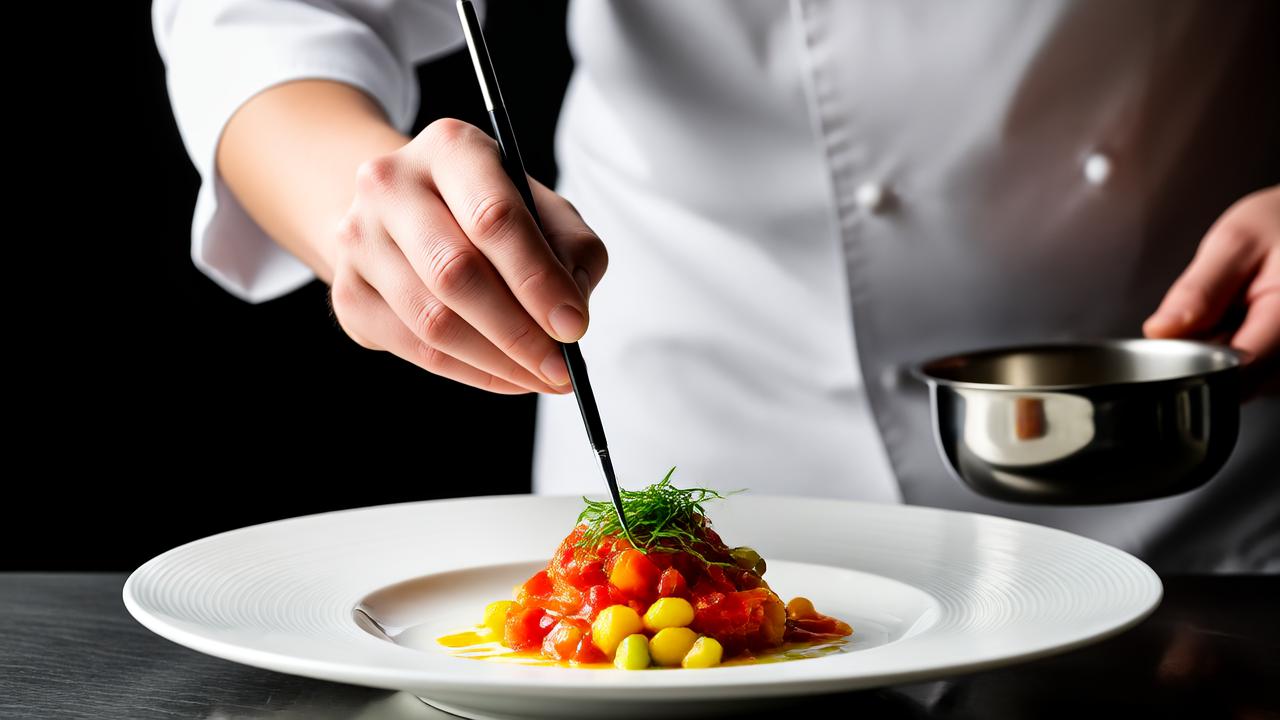}
  \end{subfigure}
  \begin{subfigure}{0.11\linewidth}
    \includegraphics[width=\linewidth]{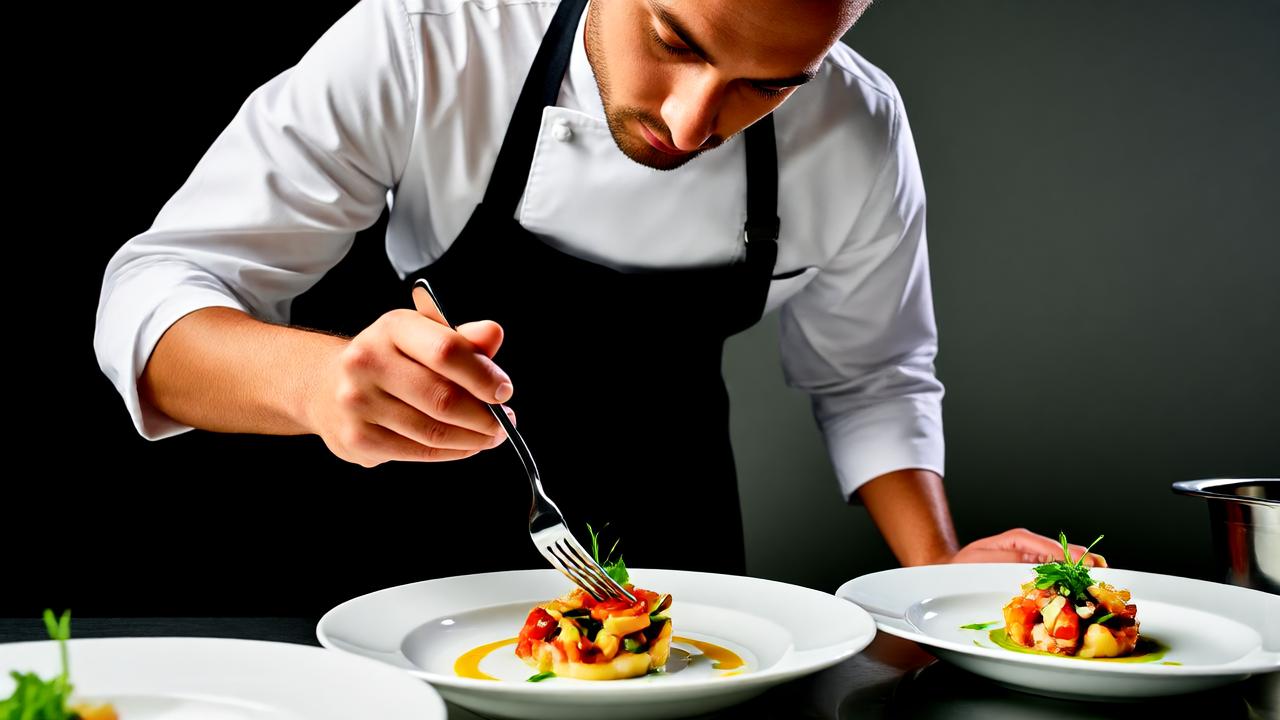}
  \end{subfigure}
  \begin{subfigure}{0.11\linewidth}
    \includegraphics[width=\linewidth]{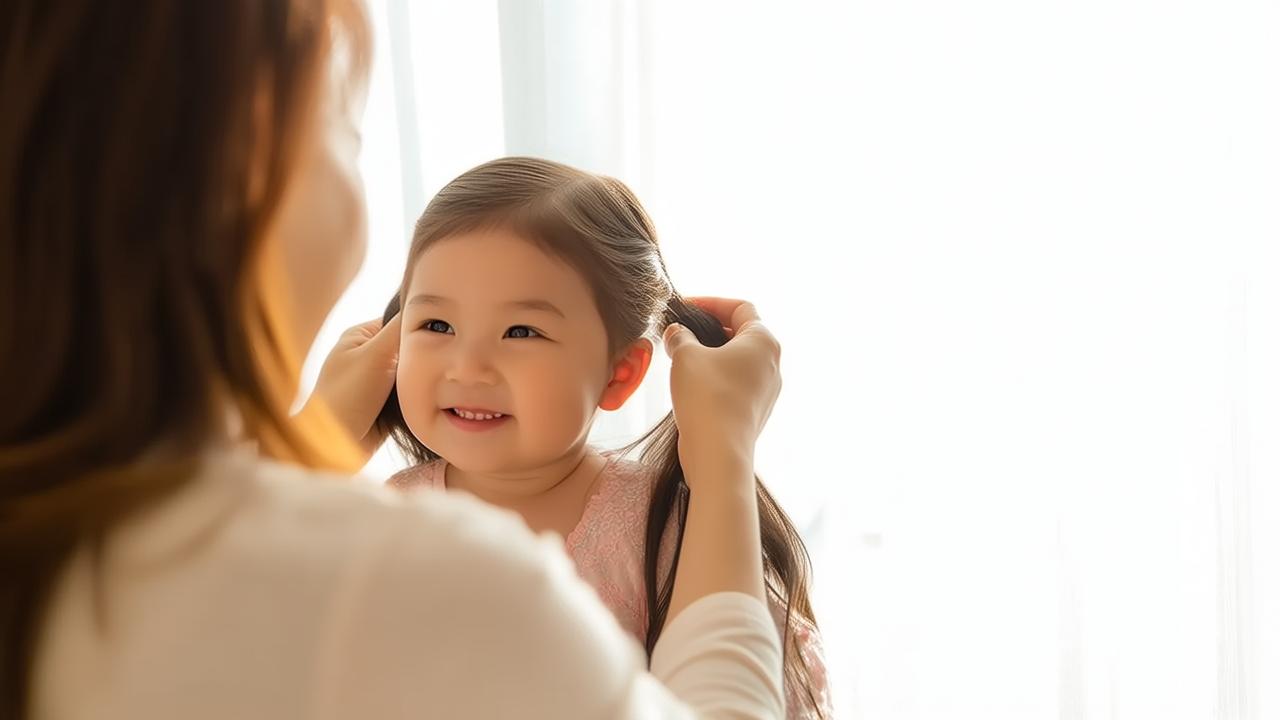}
  \end{subfigure}
  \begin{subfigure}{0.11\linewidth}
    \includegraphics[width=\linewidth]{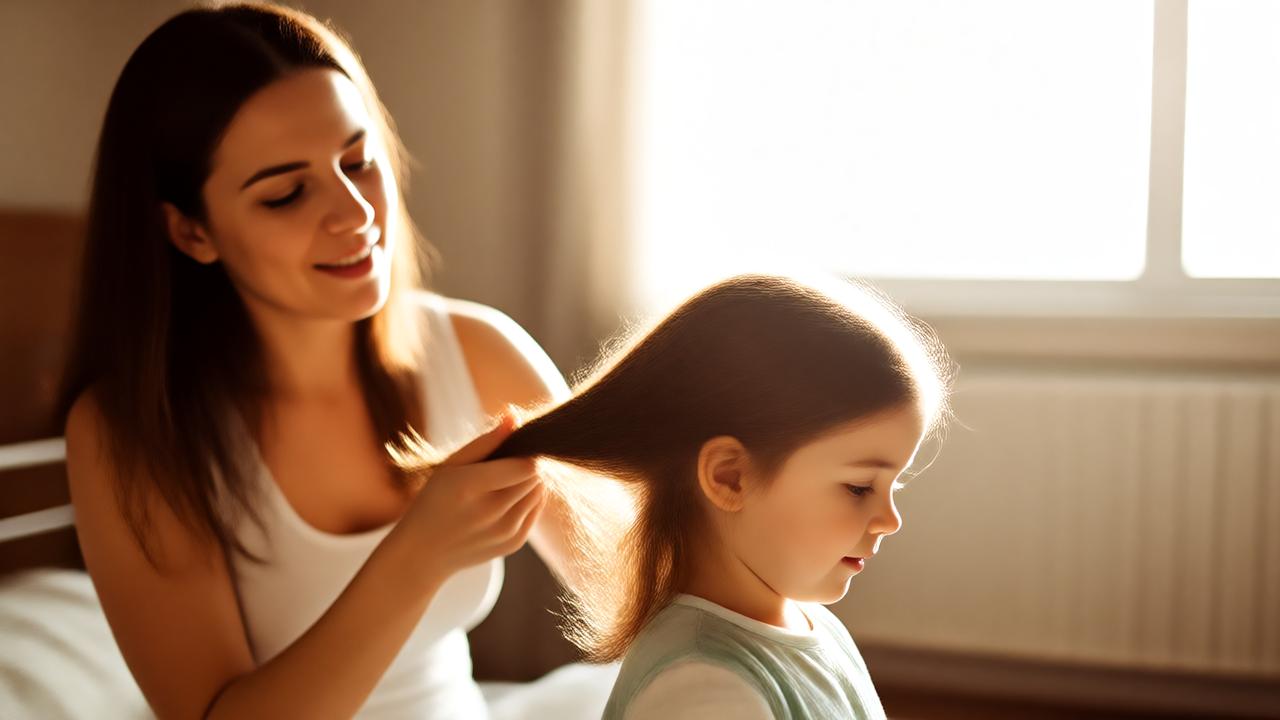}
  \end{subfigure}
  \begin{subfigure}{0.11\linewidth}
    \includegraphics[width=\linewidth]{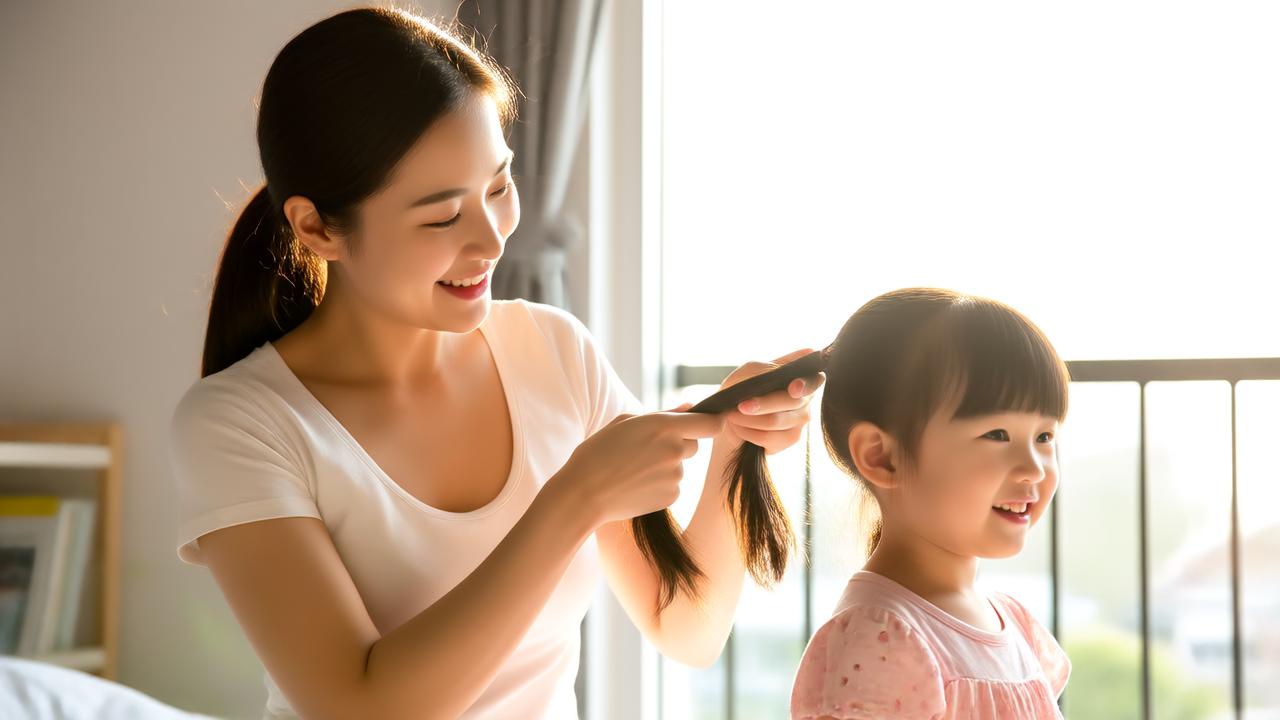}
  \end{subfigure}
  \begin{subfigure}{0.11\linewidth}
    \includegraphics[width=\linewidth]{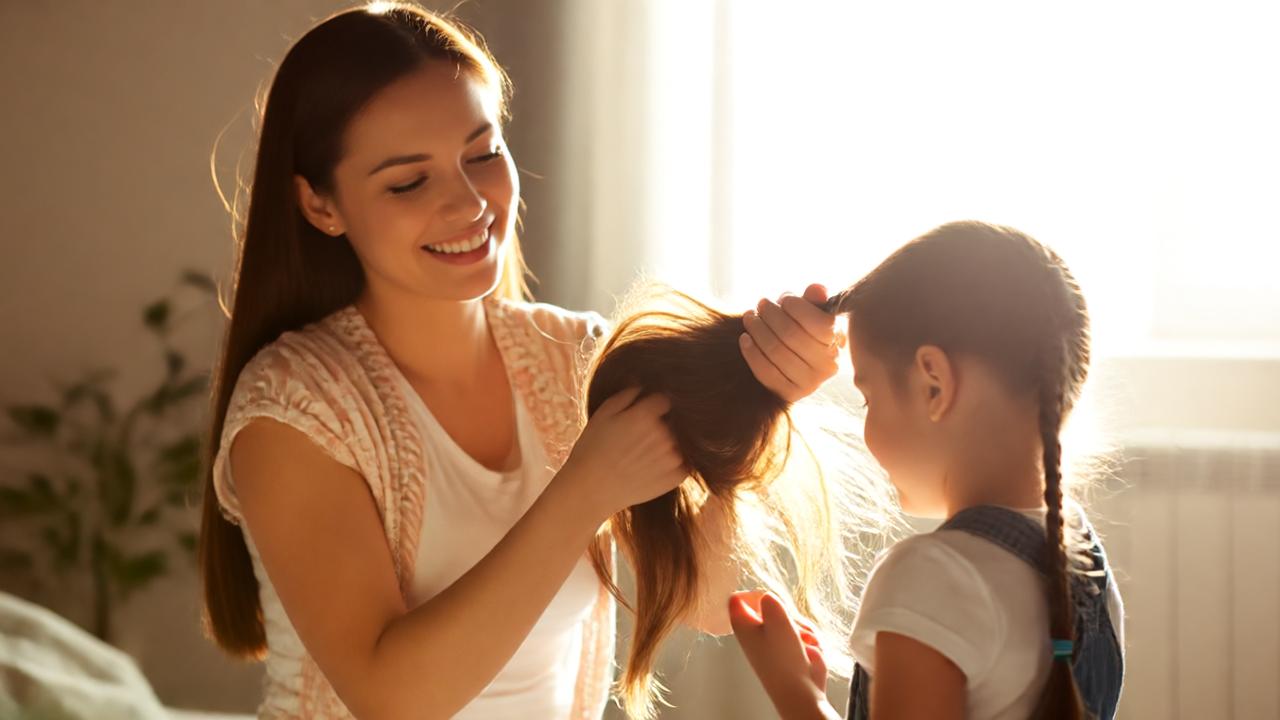}
  \end{subfigure}

  seed 2
  \begin{subfigure}{0.11\linewidth}
    \includegraphics[width=\linewidth]{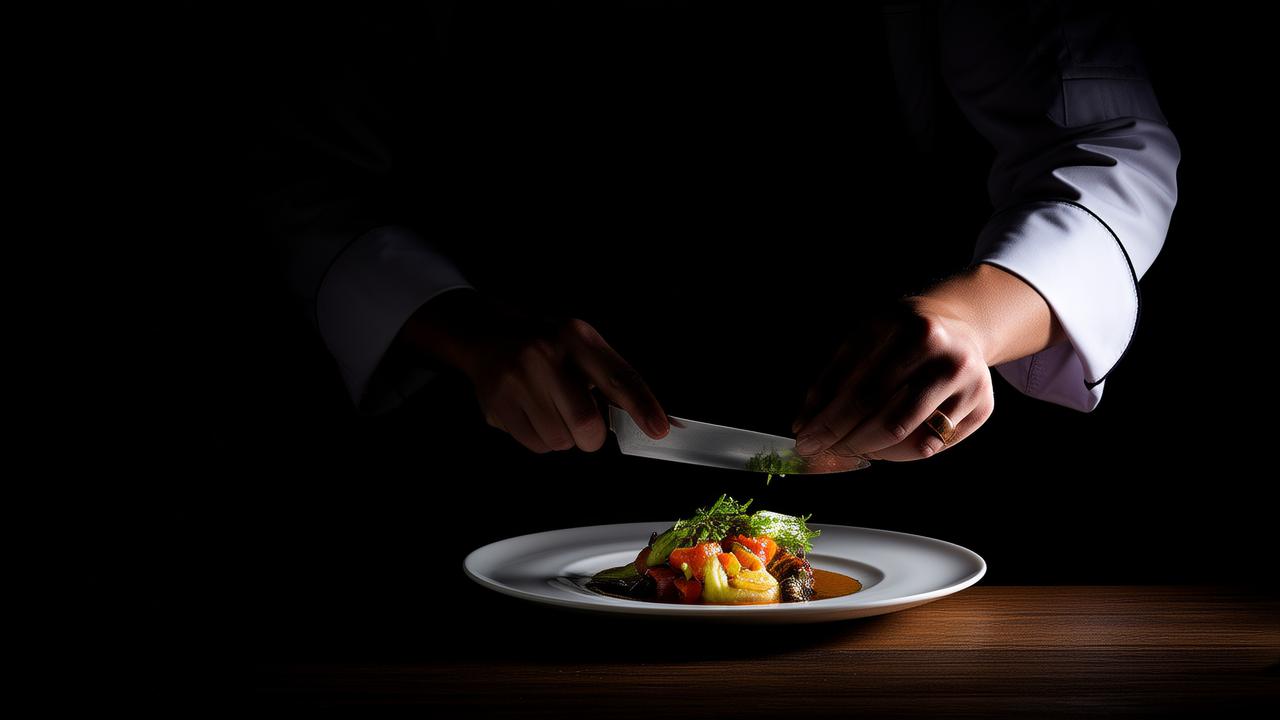}
  \end{subfigure}
  \begin{subfigure}{0.11\linewidth}
    \includegraphics[width=\linewidth]{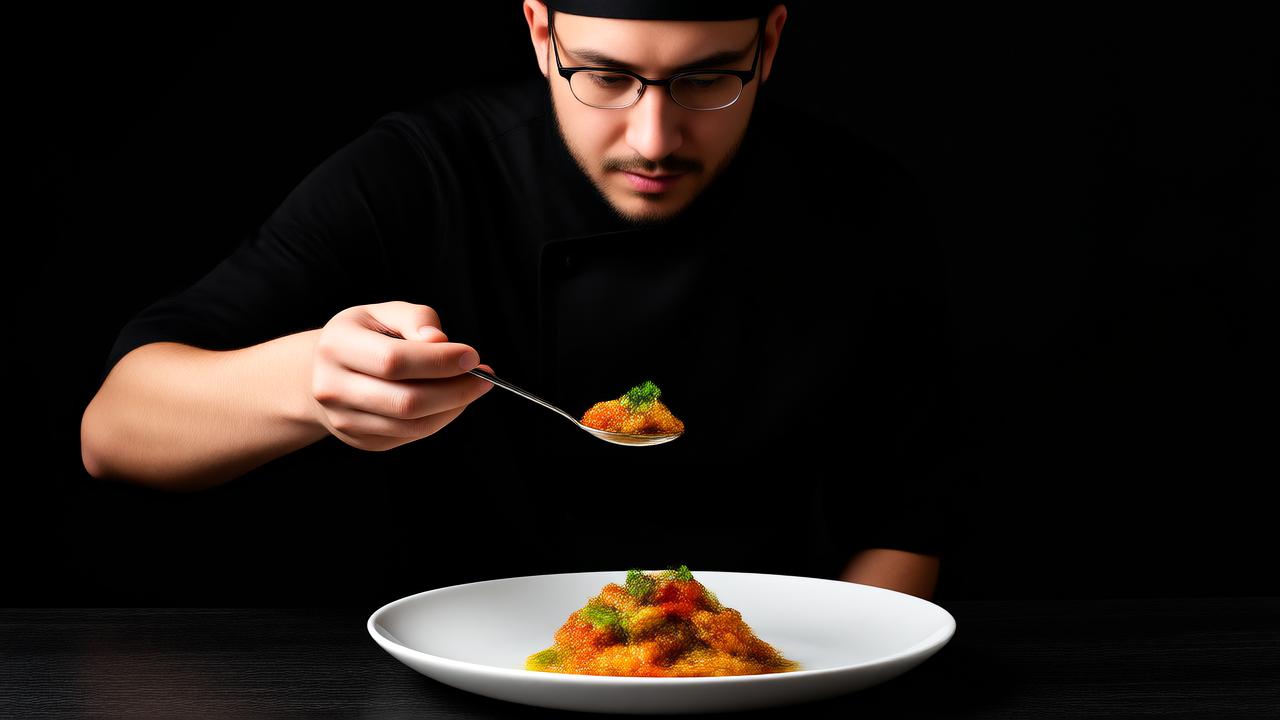}
  \end{subfigure}
  \begin{subfigure}{0.11\linewidth}
    \includegraphics[width=\linewidth]{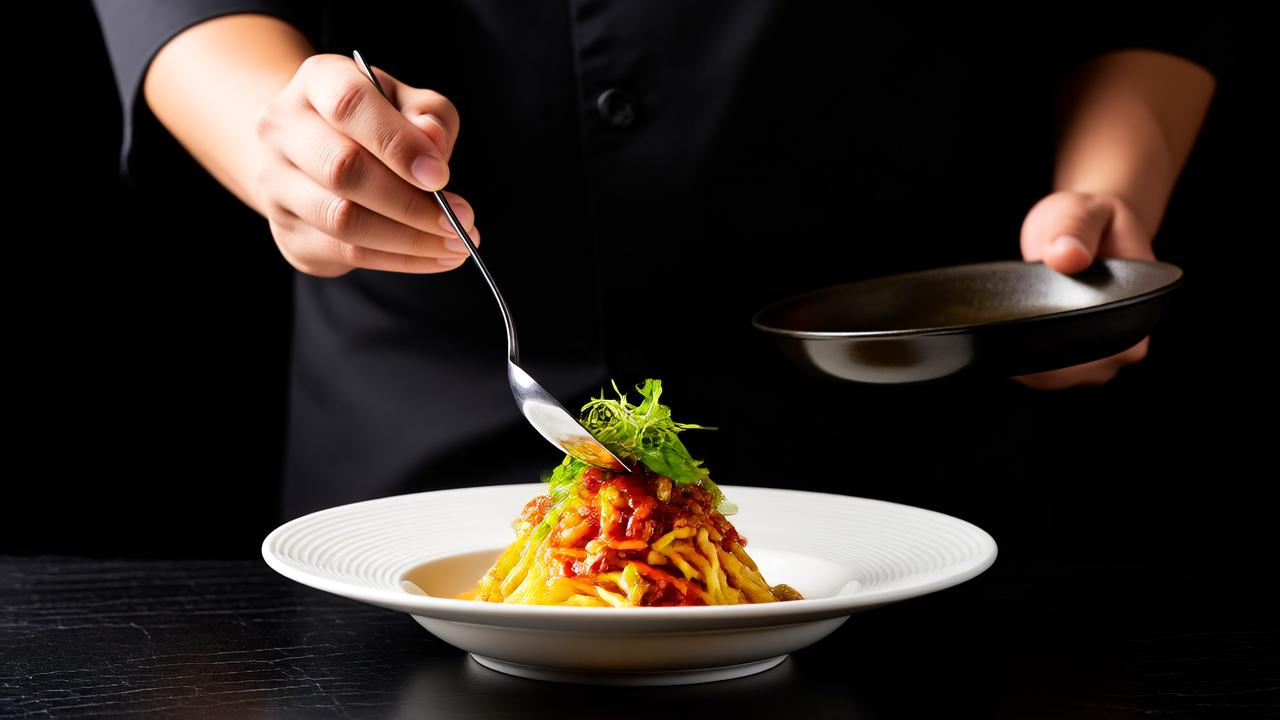}
  \end{subfigure}
  \begin{subfigure}{0.11\linewidth}
    \includegraphics[width=\linewidth]{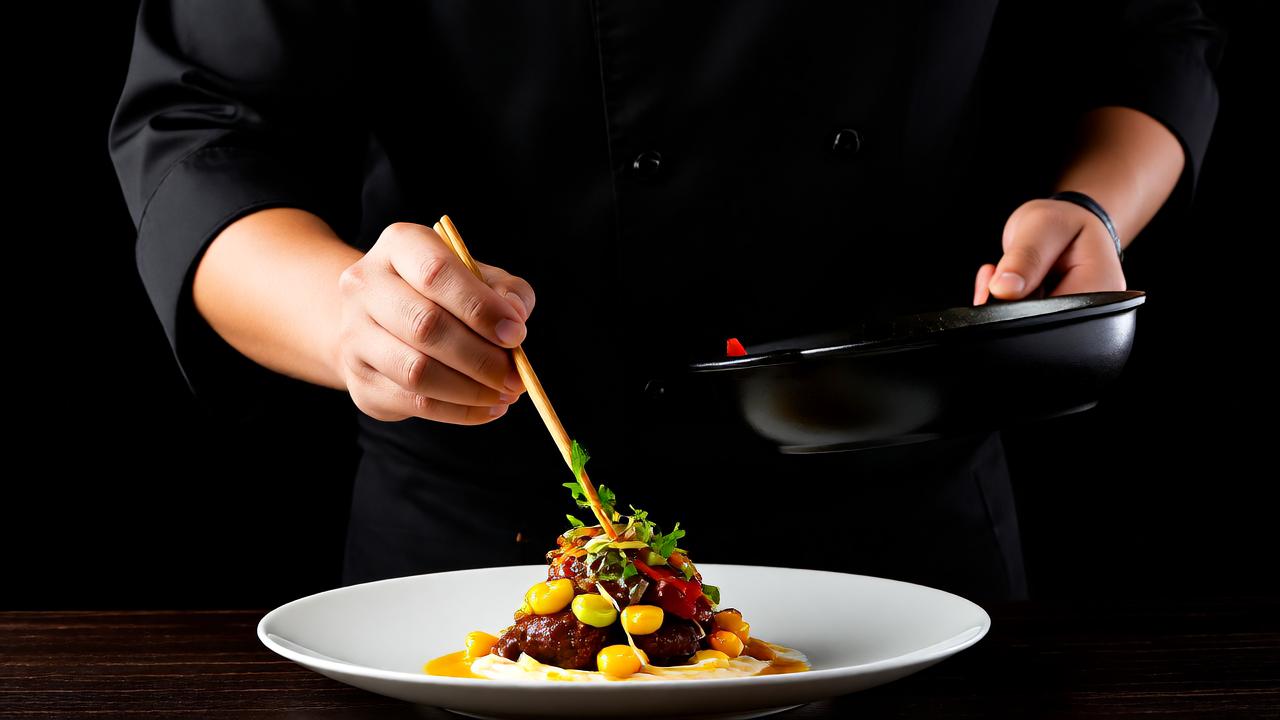}
  \end{subfigure}
  \begin{subfigure}{0.11\linewidth}
    \includegraphics[width=\linewidth]{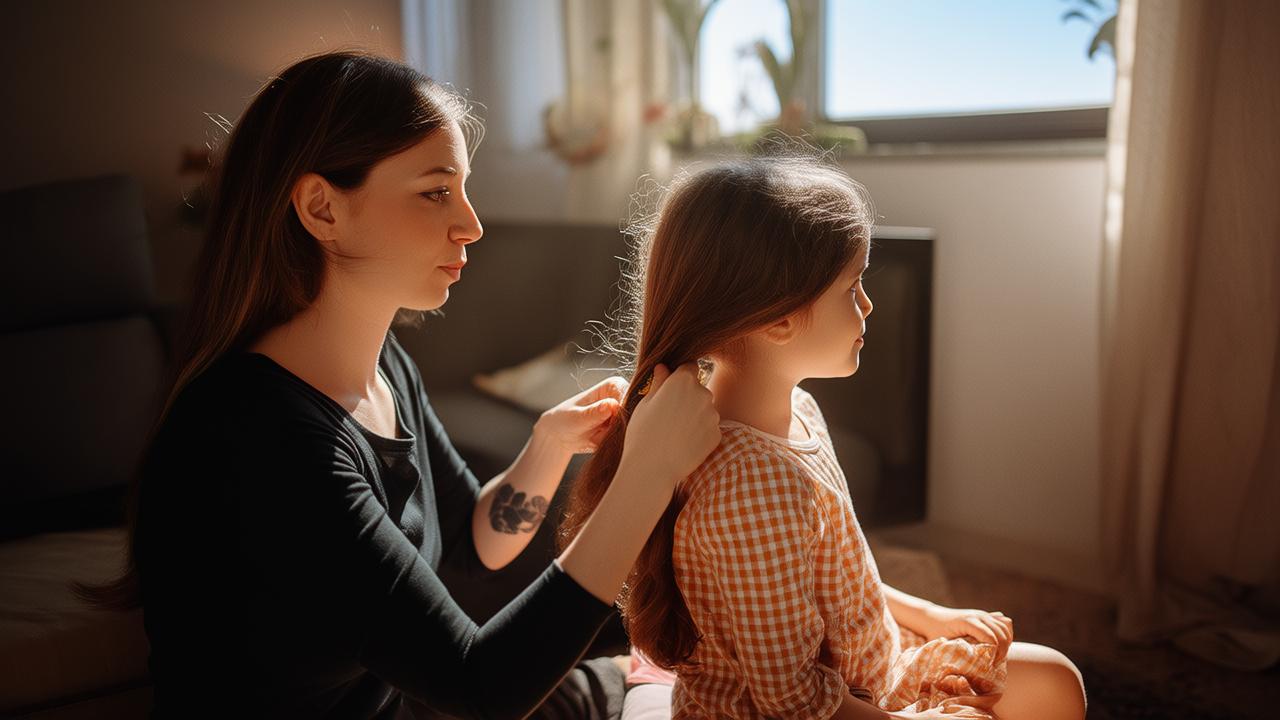}
  \end{subfigure}
  \begin{subfigure}{0.11\linewidth}
    \includegraphics[width=\linewidth]{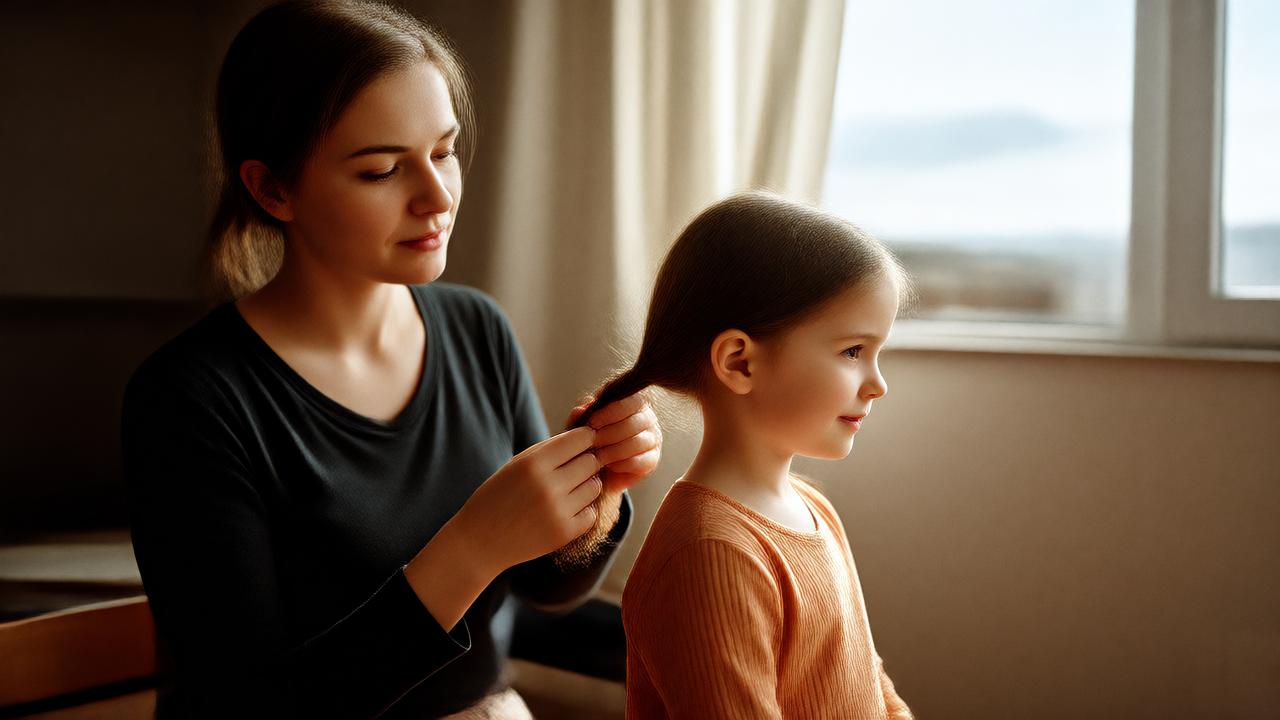}
  \end{subfigure}
  \begin{subfigure}{0.11\linewidth}
    \includegraphics[width=\linewidth]{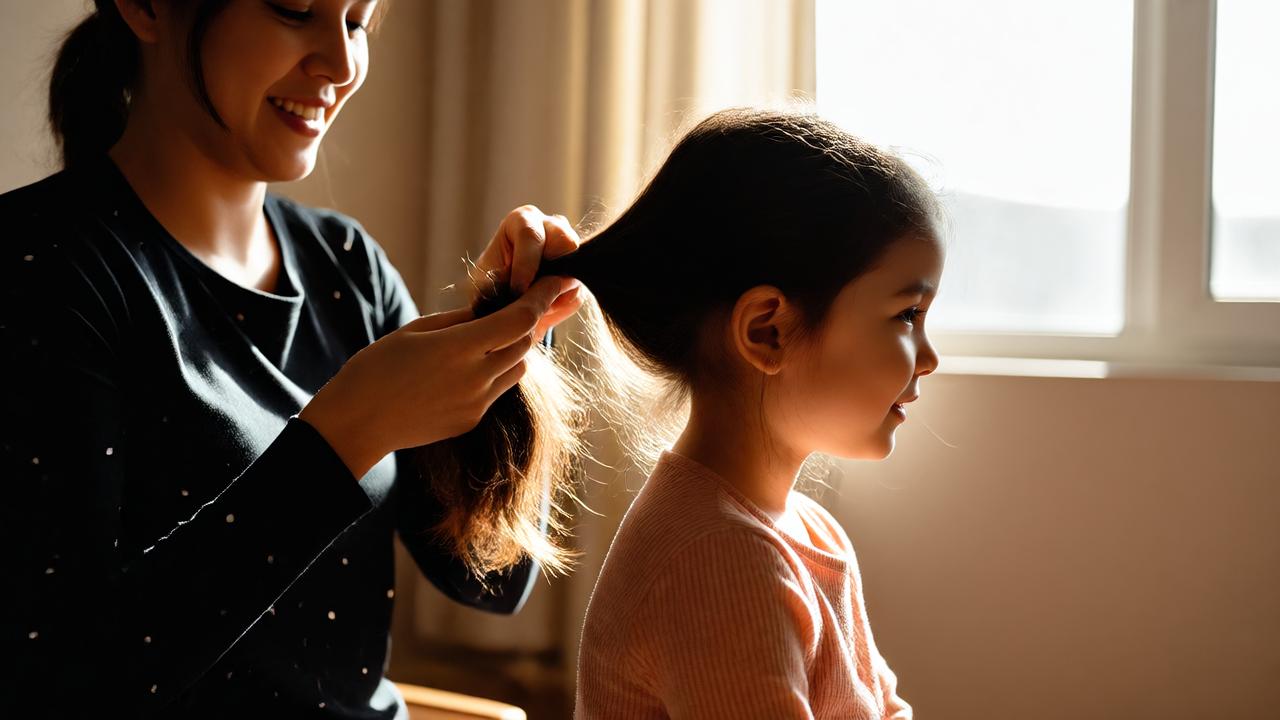}
  \end{subfigure}
  \begin{subfigure}{0.11\linewidth}
    \includegraphics[width=\linewidth]{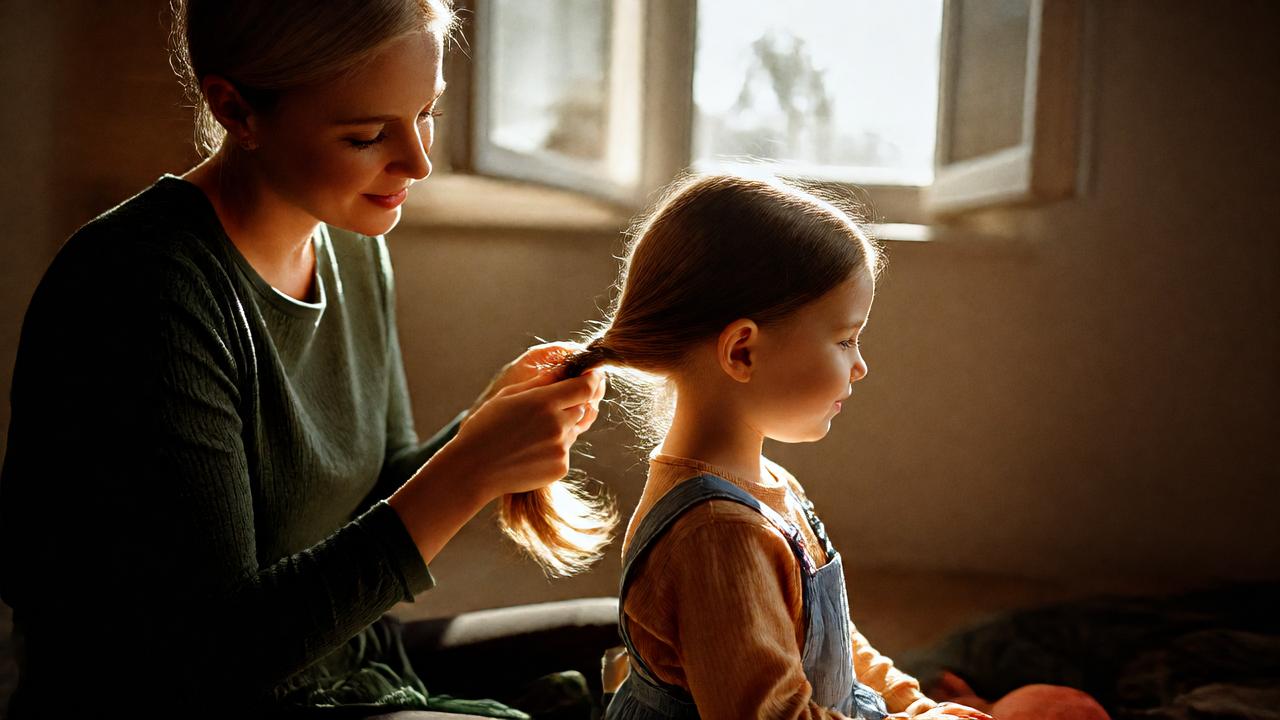}
  \end{subfigure}

  seed 3
  \begin{subfigure}{0.11\linewidth}
    \includegraphics[width=\linewidth]{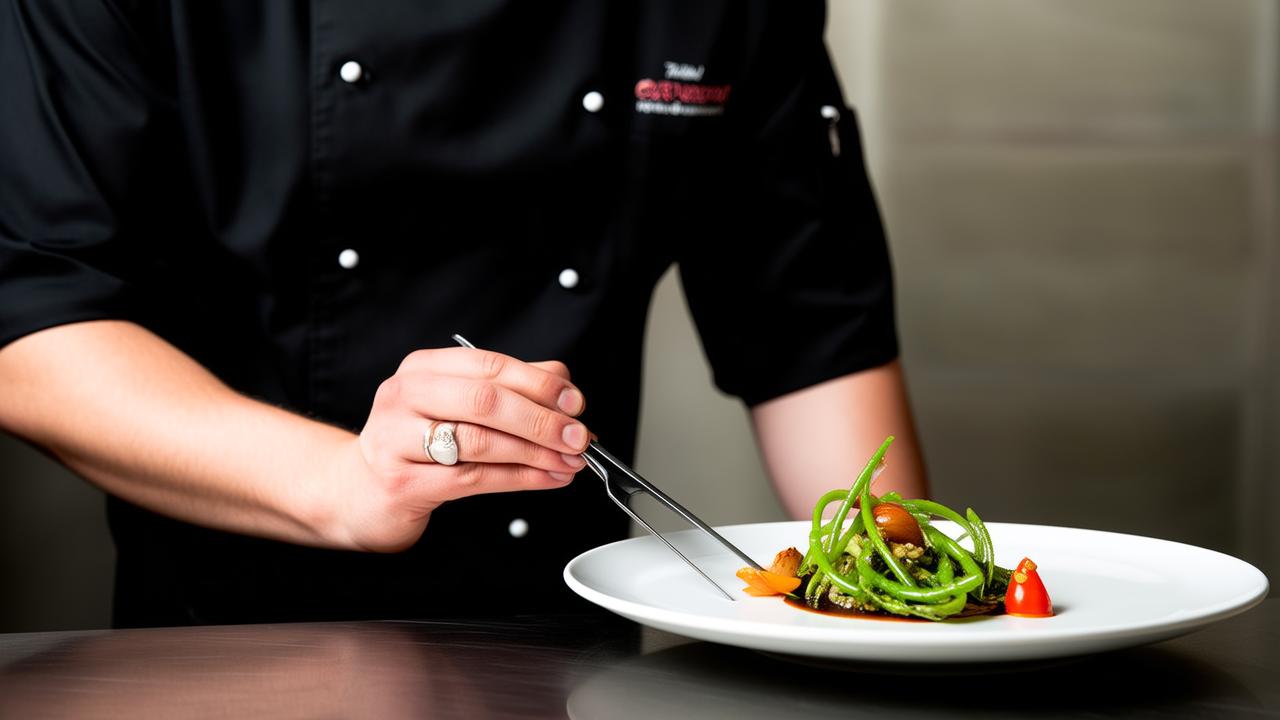}
  \end{subfigure}
  \begin{subfigure}{0.11\linewidth}
    \includegraphics[width=\linewidth]{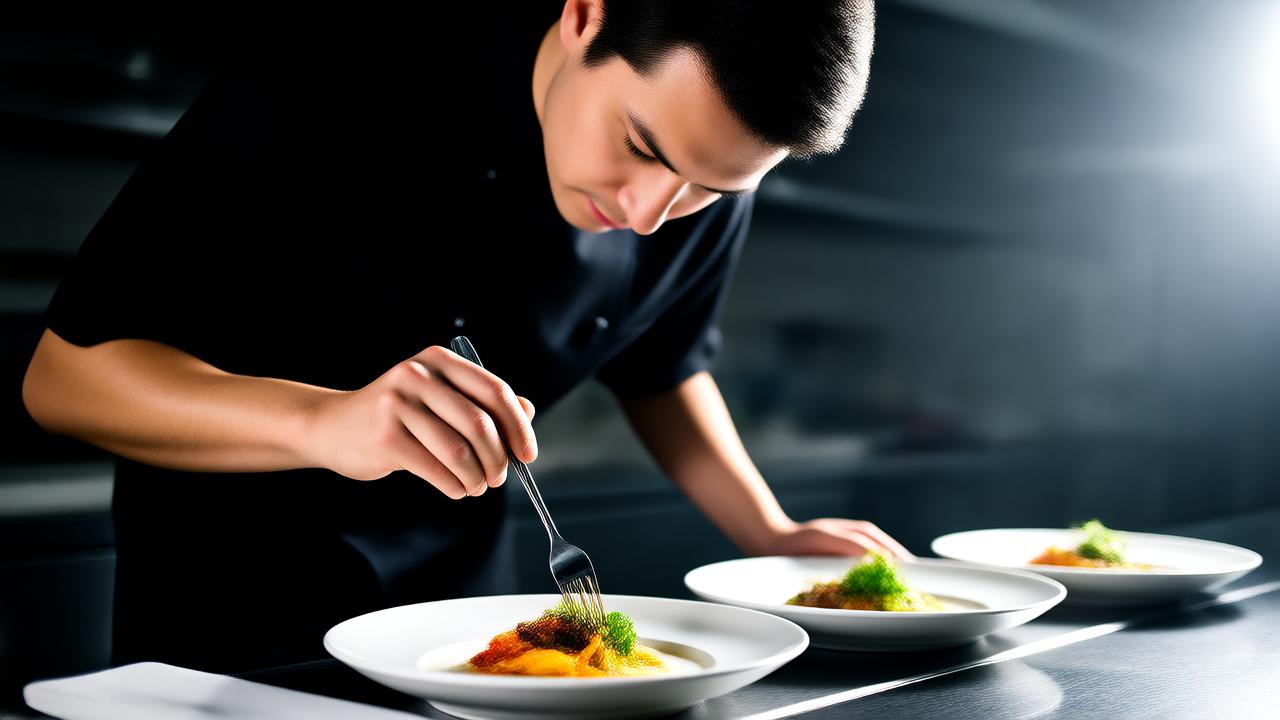}
  \end{subfigure}
  \begin{subfigure}{0.11\linewidth}
    \includegraphics[width=\linewidth]{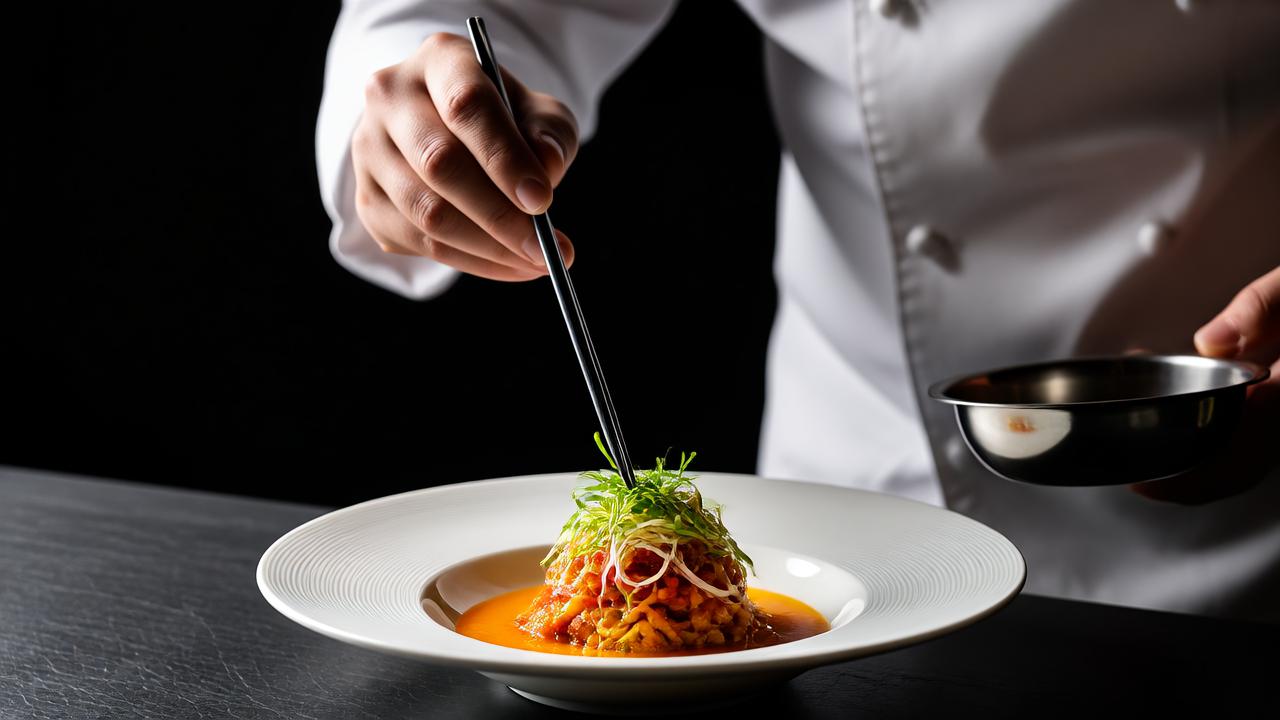}
  \end{subfigure}
  \begin{subfigure}{0.11\linewidth}
    \includegraphics[width=\linewidth]{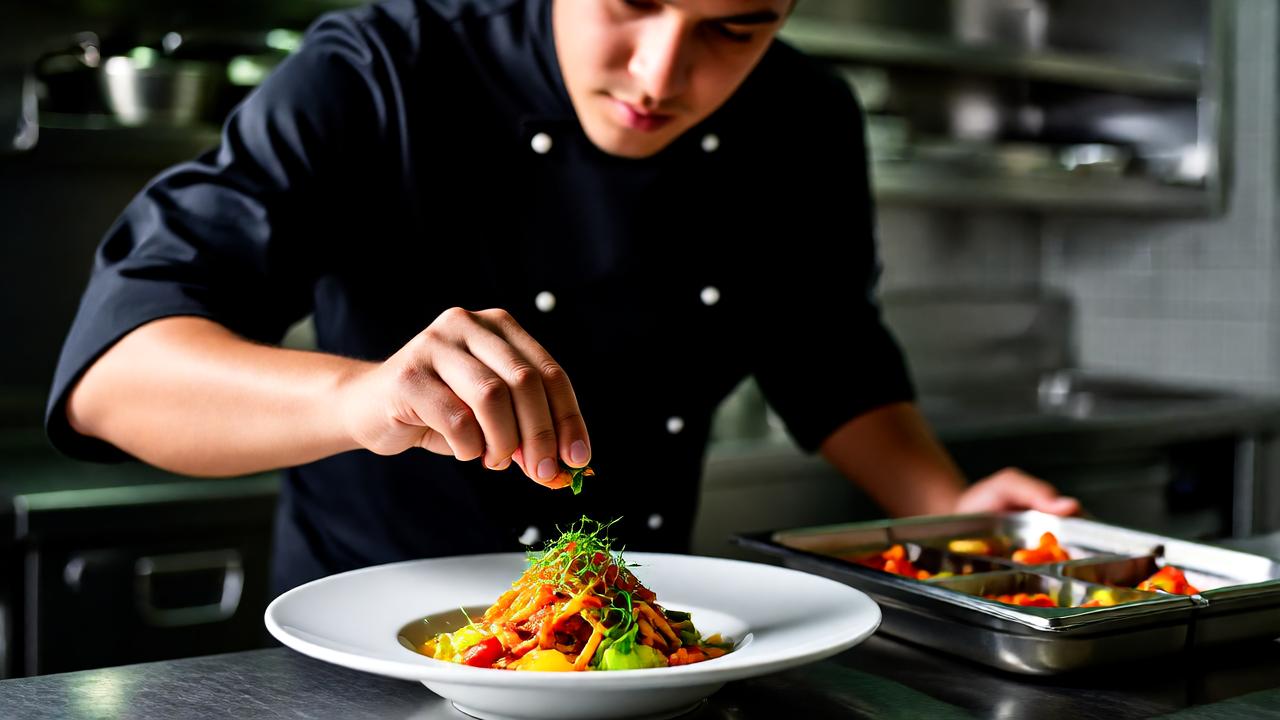}
  \end{subfigure}
  \begin{subfigure}{0.11\linewidth}
    \includegraphics[width=\linewidth]{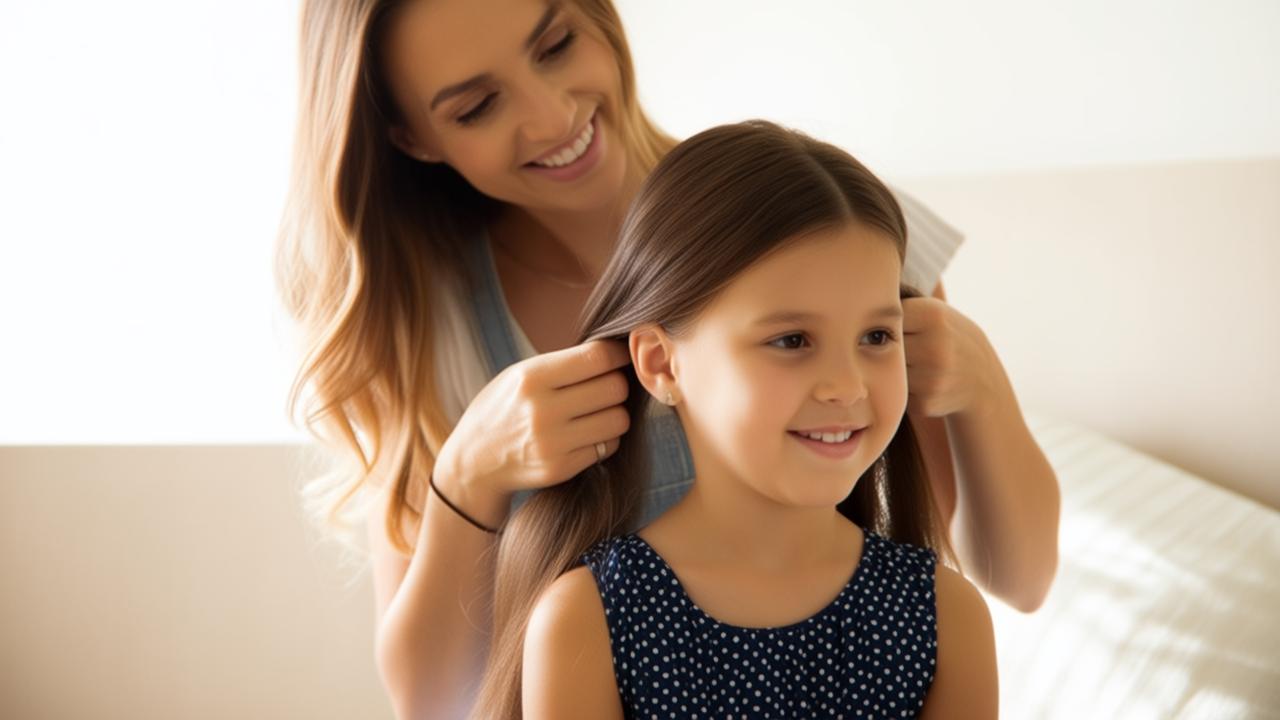}
  \end{subfigure}
  \begin{subfigure}{0.11\linewidth}
    \includegraphics[width=\linewidth]{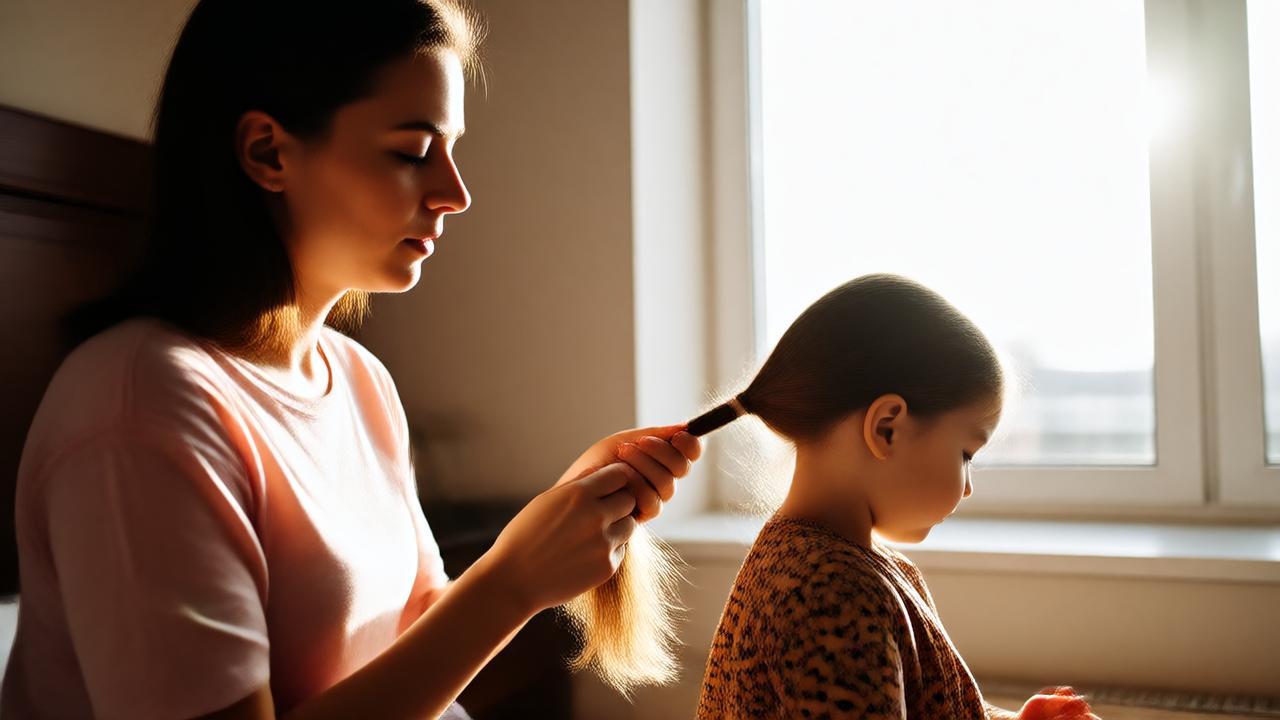}
  \end{subfigure}
  \begin{subfigure}{0.11\linewidth}
    \includegraphics[width=\linewidth]{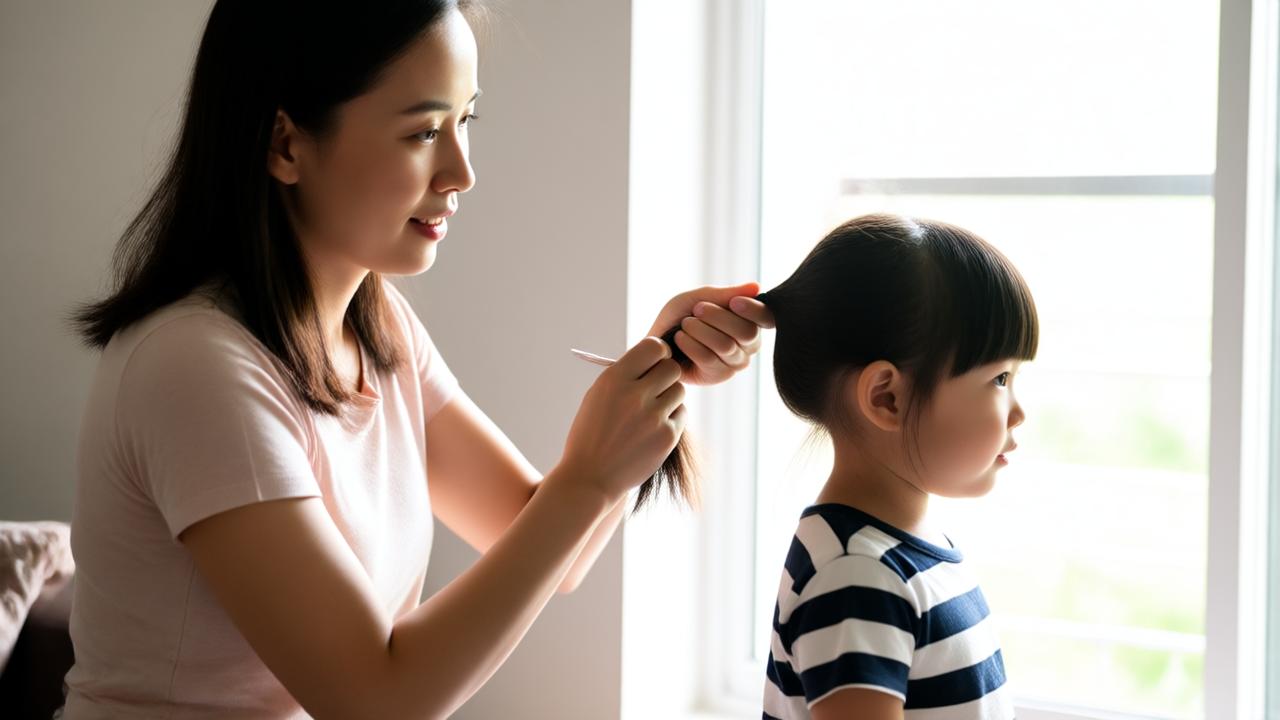}
  \end{subfigure}
  \begin{subfigure}{0.11\linewidth}
    \includegraphics[width=\linewidth]{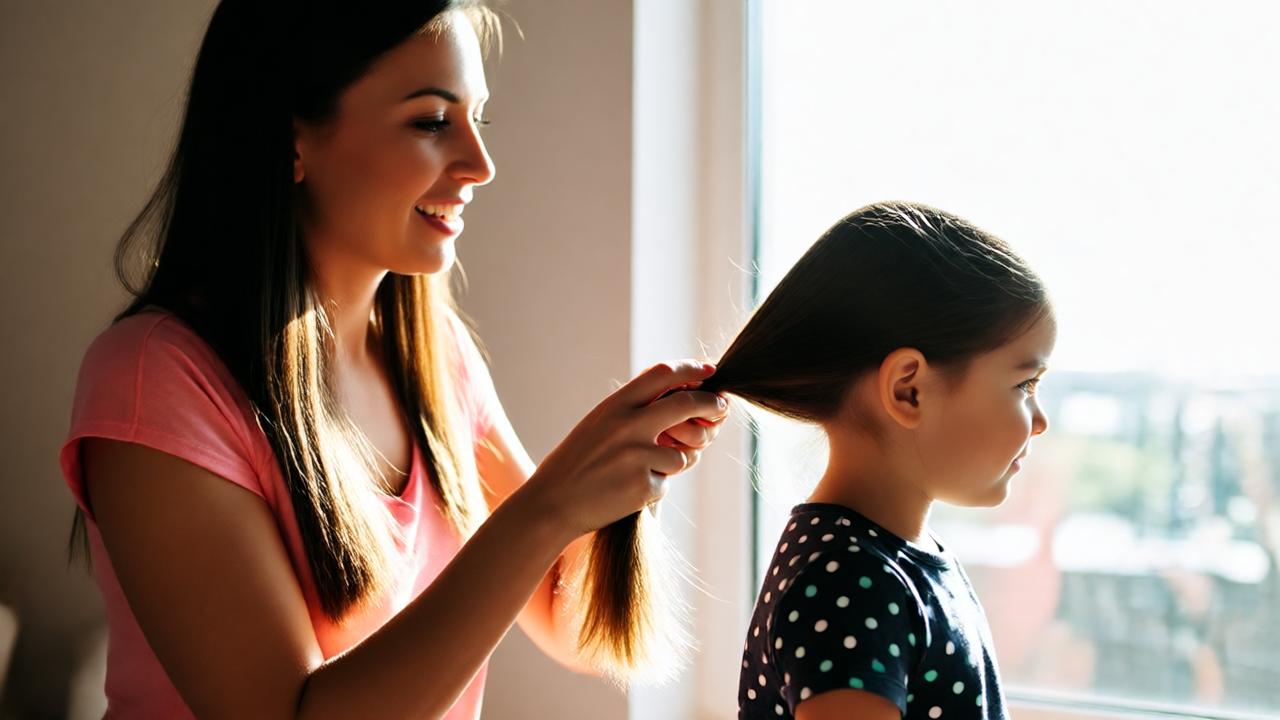}
  \end{subfigure}

  \phantom{seed3}
  \begin{subfigure}{0.11\linewidth}
  \subcaption{Base}
  \label{subfig:diversity_base}
  \end{subfigure}
  \begin{subfigure}{0.11\linewidth}
  \subcaption{Vanilla}
  \label{subfig:diversity_DMD}
  \end{subfigure}
  \begin{subfigure}{0.11\linewidth}
  \subcaption{DMD2}
  \label{subfig:diversity_DMD_wi_SGTS}
  \end{subfigure}
  \begin{subfigure}{0.11\linewidth}
  \subcaption{Ours}
  \label{subfig:diversity_phased_DMD}
  \end{subfigure}
  \begin{subfigure}{0.11\linewidth}
  \subcaption{Base}
  \label{subfig:diversity_base2}
  \end{subfigure}
  \begin{subfigure}{0.11\linewidth}
  \subcaption{Vanilla}
  \label{subfig:diversity_DMD2}
  \end{subfigure}
  \begin{subfigure}{0.11\linewidth}
  \subcaption{DMD2}
  \label{subfig:diversity_DMD_wi_SGTS2}
  \end{subfigure}
  \begin{subfigure}{0.11\linewidth}
  \subcaption{Ours}
  \label{subfig:diversity_phased_DMD2}
  \end{subfigure}

  \caption{
  Samples (seeds 0-3) from the Wan2.1-T2V-14B base model (40 steps, CFG=4) and its distilled variants (4 steps, CFG=1): (a, e) Base, (b, f) Vanilla few-step DMD, (c, g) DMD2, (d, h) Phased DMD.
  Left: ``A chef meticulously plating a dish.''. Right: ``A mother braiding her daughter’s hair, sunlight warming the room.''
  }
\end{figure}

We apply \ourmodel to distill three base models for text-to-image generation. 
% Phased DMD is applied to three base models for text-to-image distillation. 
To evaluate generative diversity, we constructed a test set of 21 prompts, each providing a short image description without detailed specifications. For each prompt, we generated 8 images using random seeds from 0 to 7.
The base model was sampled with 40 steps and a guidance scale of 4, while all distilled models used 4 steps and a guidance scale of 1.
% 
% For the base model, images are sampled using 40 steps with a guidance scale of 4. All distilled models are sampled using 4 steps and a guidance scale of 1.
As shown in Fig.~\ref{subfig:diversity_DMD}, images generated by the vanilla 4-step DMD model exhibit a loss of fine details. Although the 4-step DMD2 model improves image quality, it does so at the cost of reduced diversity. Fig.~\ref{subfig:diversity_DMD_wi_SGTS} reveals that the generated images often converge to a similar close-up view with limited compositional variation across seeds. In contrast, \ourmodel better preserves diversity, producing images with a wider range of natural compositions and lighting conditions, as illustrated in Fig.~\ref{subfig:diversity_phased_DMD}.

Generative diversity is evaluated using two complementary metrics: (1) the mean pairwise cosine similarity of DINOv3 features \citep{simeoni2025dinov3}, where lower values indicate higher diversity, and (2) the mean pairwise LPIPS distance \citep{zhang2018LPIPS}, where higher values denote greater diversity. Both metrics are computed across images generated from the same prompt using different seeds.
The quantitative results are presented in Tab.~\ref{tab:diversity_quant}. As expected, the base models achieve the highest diversity. 
% Notably, DMD with SGTS yields slightly lower diversity than vanilla DMD.
\ourmodel outperforms both vanilla DMD and DMD2, demonstrating its superior capability for preserving the generative diversity of the original model.
The diversity improvement on Qwen-Image is marginal. We argue this stems from the base model's own limited output diversity.

\begin{figure}[htb]
  \centering

  \begin{subfigure}{0.24\linewidth}
    \includegraphics[width=\linewidth]{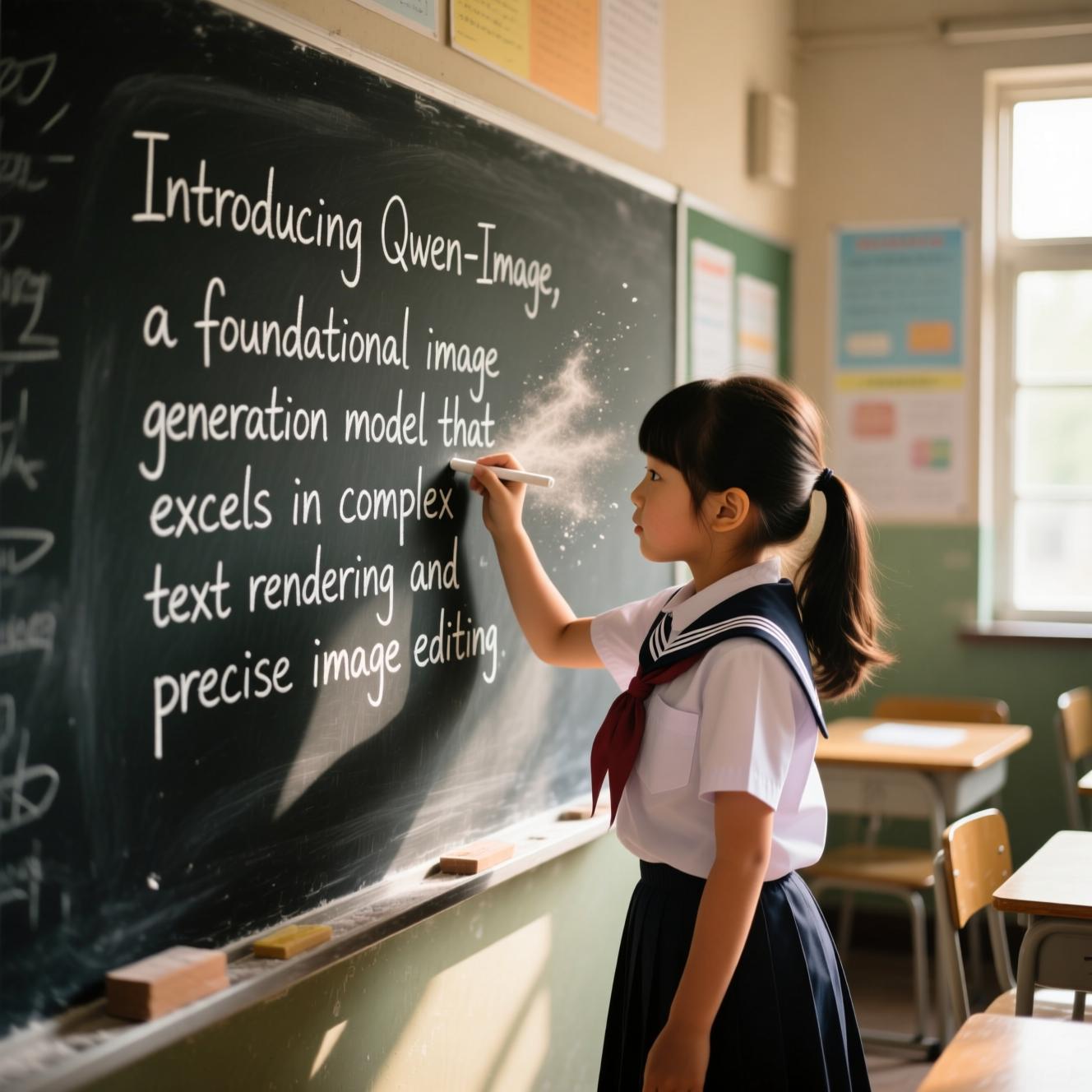}
  \end{subfigure}
  \begin{subfigure}{0.24\linewidth}
    \includegraphics[width=\linewidth]{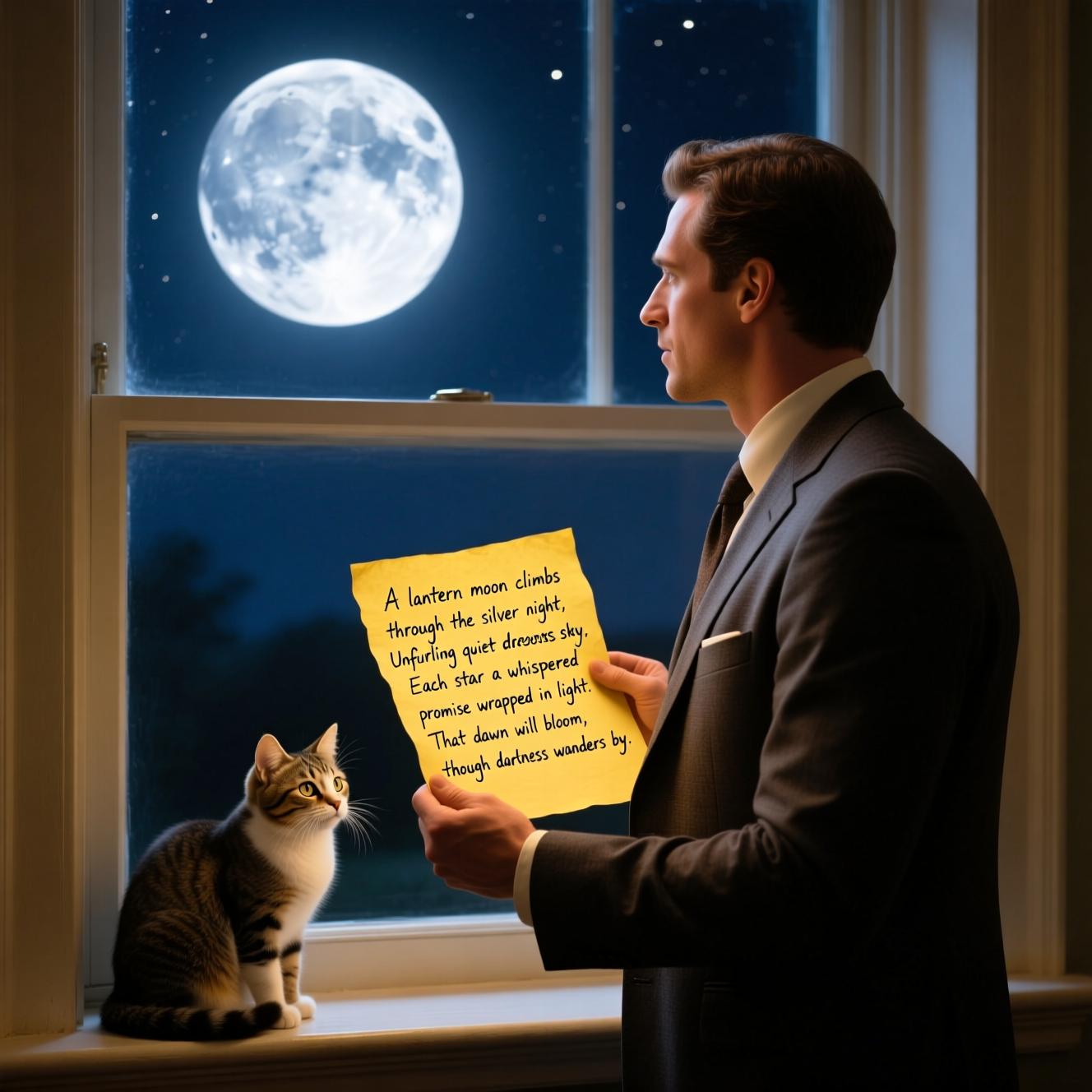}
  \end{subfigure}
  \begin{subfigure}{0.24\linewidth}
    \includegraphics[width=\linewidth]    {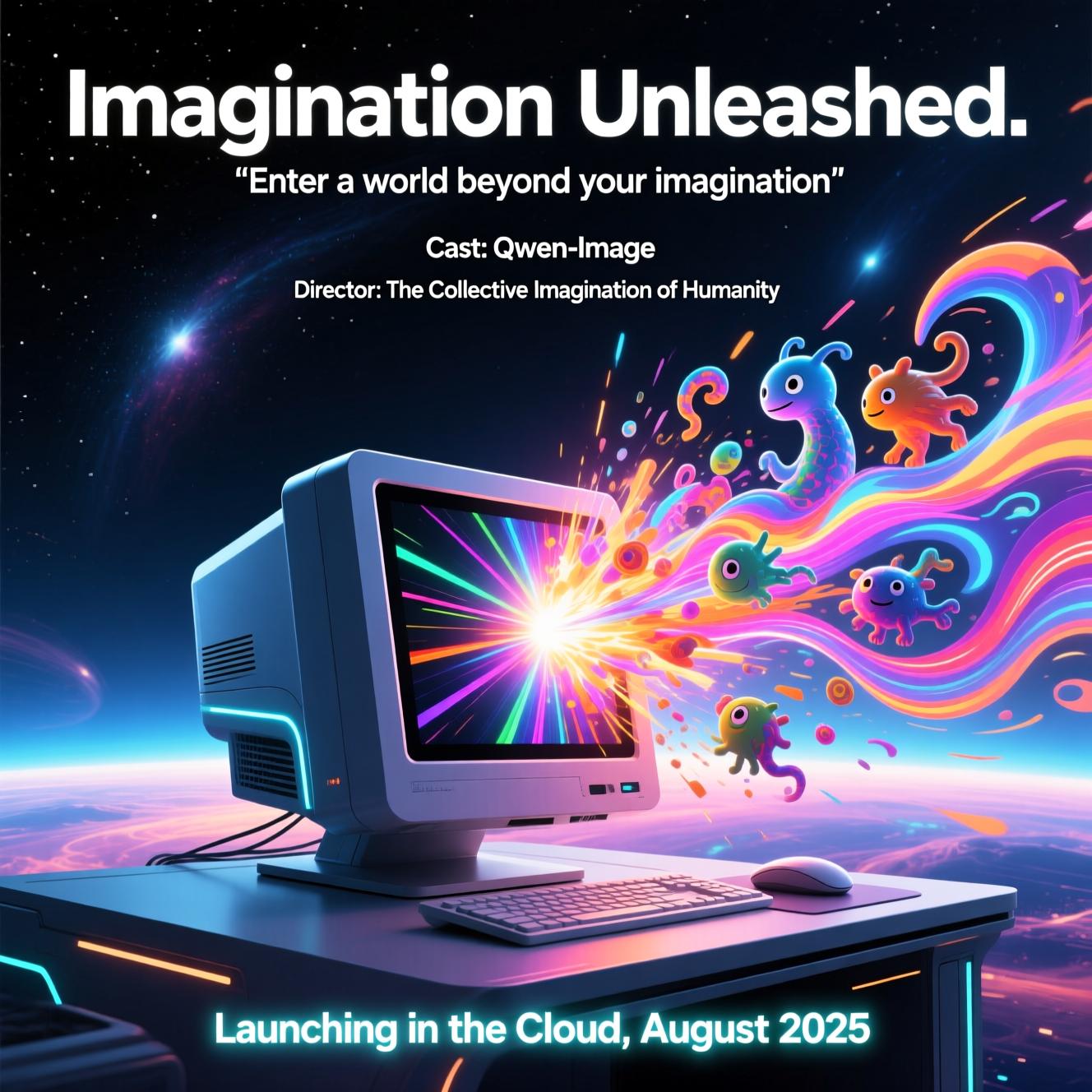}
  \end{subfigure}
  \begin{subfigure}{0.24\linewidth}
    \includegraphics[width=\linewidth]{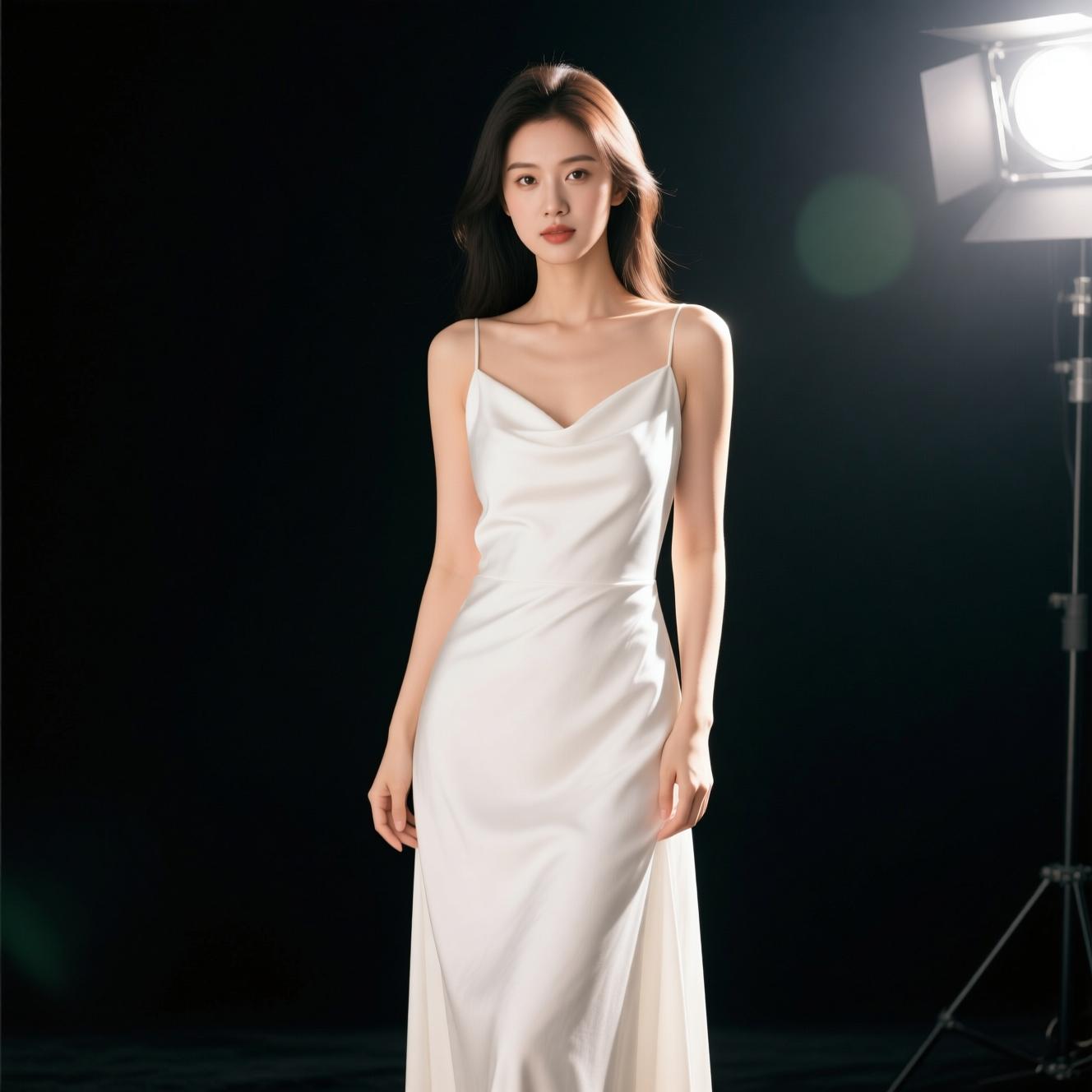}
  \end{subfigure}
  \caption{
    Examples generated by the Qwen-Image distilled with Phased DMD.
    % The base model's text rendering capability remains well-preserved.
  }
  \label{fig:qwen_examples}
\end{figure}

Qwen-Image is recognized for its faithful adherence to prompts and high-quality text rendering.
To evaluate the preservation of these capabilities after distillation, we applied \ourmodel to Qwen-Image and generated images using prompts from its official website \citep{qwen_image_blog}.
As shown in Fig.~\ref{fig:qwen_examples}, the model distilled with \ourmodel exhibits well-preserved capabilities, producing high-quality images with accurate text rendering.

% Cosine Similarity is computed by DinoV3 \citep{simeoni2025dinov3}
\begin{table*}[th]
\caption{
Two metrics for quantitative diversity evaluation: average pairwise DINOv3 cosine similarity (lower is better) and LPIPS distance (higher is better).
Phased DMD outperforms the vanilla DMD and DMD2 in preserving generative diversity of the base models.
% Quantitative diversity evaluation in text-to-image tasks. 
}
\label{tab:diversity_quant}
% \caption{
% Quantitative diversity evaluation using mean pairwise DINOv3 cosine similarity (lower is better).
% Phased DMD outperforms the vanilla DMD and DMD with SGTS in preserving generative diversity of the base models.
% }
\centering
\begin{tabular}{c c c c c c c}
\toprule
% & \multicolumn{2}{c}{\bf{Wan2.1-T2V-14B}} & \multicolumn{2}{c}{\bf{Wan2.2-T2V-A14B}} \\
% \midrule
% \multirow{2}{*}{\bf Method}  & \bf{DINOv3}    & {\bf LPIPS   }  & \bf{DINOv3}    & {\bf LPIPS   } \\
% & \bf{Similarity} $ \downarrow $ & \bf{Distance}   $\uparrow $ & \bf{Similarity} $ \downarrow $ & \bf{Distance}  $\uparrow $ \\
% \midrule
% Base model & 0.708 &  0.710  & 0.732  & 0.678   \\
% \midrule
% DMD & 0.825  & 0.661  & - & - \\
% DMD with SGTS & 0.826  & 0.677 & 0.828 & 0.578  \\
% Phased DMD (Ours) & \textbf{0.782}  & \textbf{0.695}  & \textbf{0.768} & \textbf{0.651}  \\
% & \bf{Wan2.1-T2V-14B} & \multicolumn{2}{c}{\bf{Wan2.2-T2V-A14B}} \\
% \midrule
% \multirow{2}{*}{\bf Method}  & \bf{DINOv3}    & {\bf LPIPS   }  & \bf{DINOv3}    & {\bf LPIPS   } \\
% \multirow{2}{*}{\bf Method}  & \bf{DINOv3}    & {\bf LPIPS   }  & \bf{DINOv3}    & {\bf LPIPS   } \\
\multirow{2}{*}{\bf Method} &   \multicolumn{2}{c}{\bf{Wan2.1-T2V-14B}} & \multicolumn{2}{c}{\bf{Wan2.2-T2V-A14B}} & \multicolumn{2}{c}{\bf{Qwen-Image}} \\
 & \bf DINOv3 $ \downarrow $ & \bf LPIPS $\uparrow $ & \bf DINOv3 $ \downarrow $ & \bf LPIPS $\uparrow $ & \bf DINOv3 $ \downarrow $ & \bf LPIPS $\uparrow $ \\
\midrule
Base model &  \bf{0.708} & \bf{0.607} & \bf{0.732} & \bf{0.531} &  \bf{0.907} & \bf{0.483} \\
\midrule
Vanilla DMD & 0.825 & 0.522 & - & - & - & - \\
DMD2 & 0.826 & 0.521 & 0.828 & 0.447 & \underline{0.941} & 0.309 \\
Phased DMD (Ours) & \underline{0.782} & \underline{0.544} & \underline{0.768} & \underline{0.481} & 0.958 &\underline{0.322} \\
\bottomrule
\end{tabular}
\end{table*}

\subsection{Merit of MoE}
\label{subsection:merit_of_moe}
\begin{figure}[hht]
  \centering

  \begin{subfigure}{0.24\linewidth}
    \includegraphics[width=\linewidth]{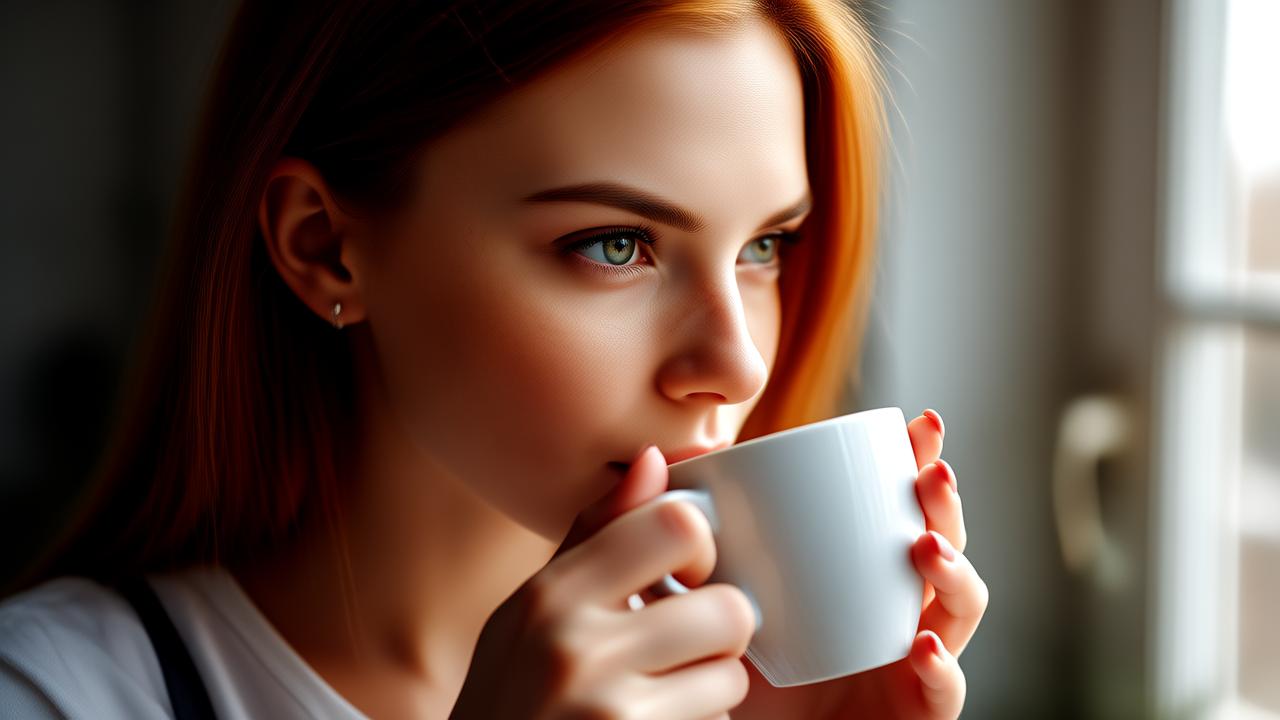}
  \end{subfigure}
  \begin{subfigure}{0.24\linewidth}
    \includegraphics[width=\linewidth]{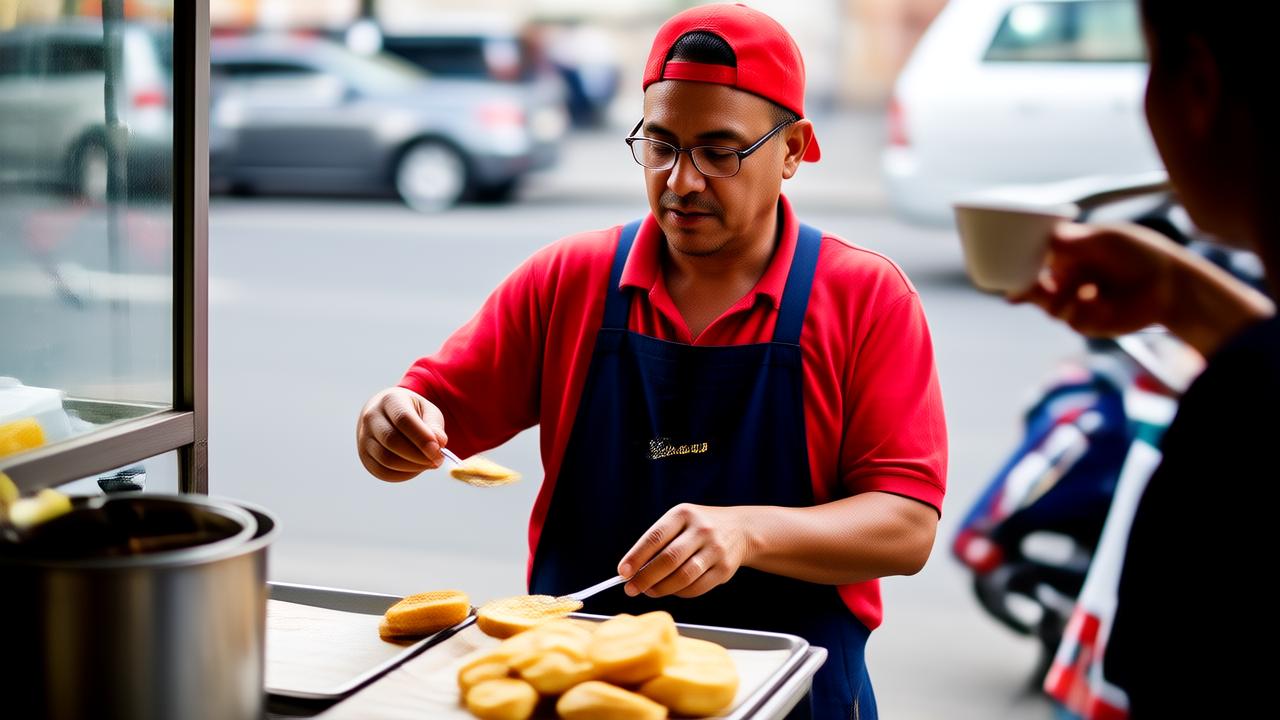}
  \end{subfigure}
  \begin{subfigure}{0.24\linewidth}
    \includegraphics[width=\linewidth]{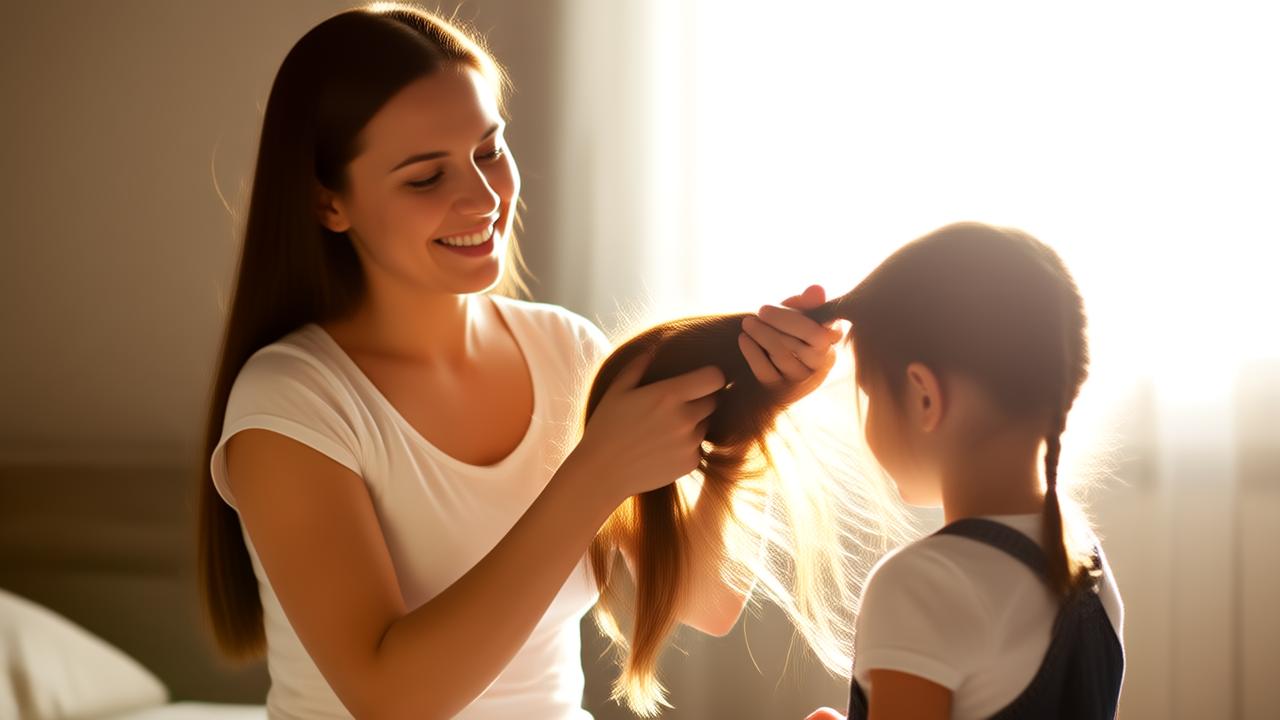}
  \end{subfigure}
  \begin{subfigure}{0.24\linewidth}
    \includegraphics[width=\linewidth]{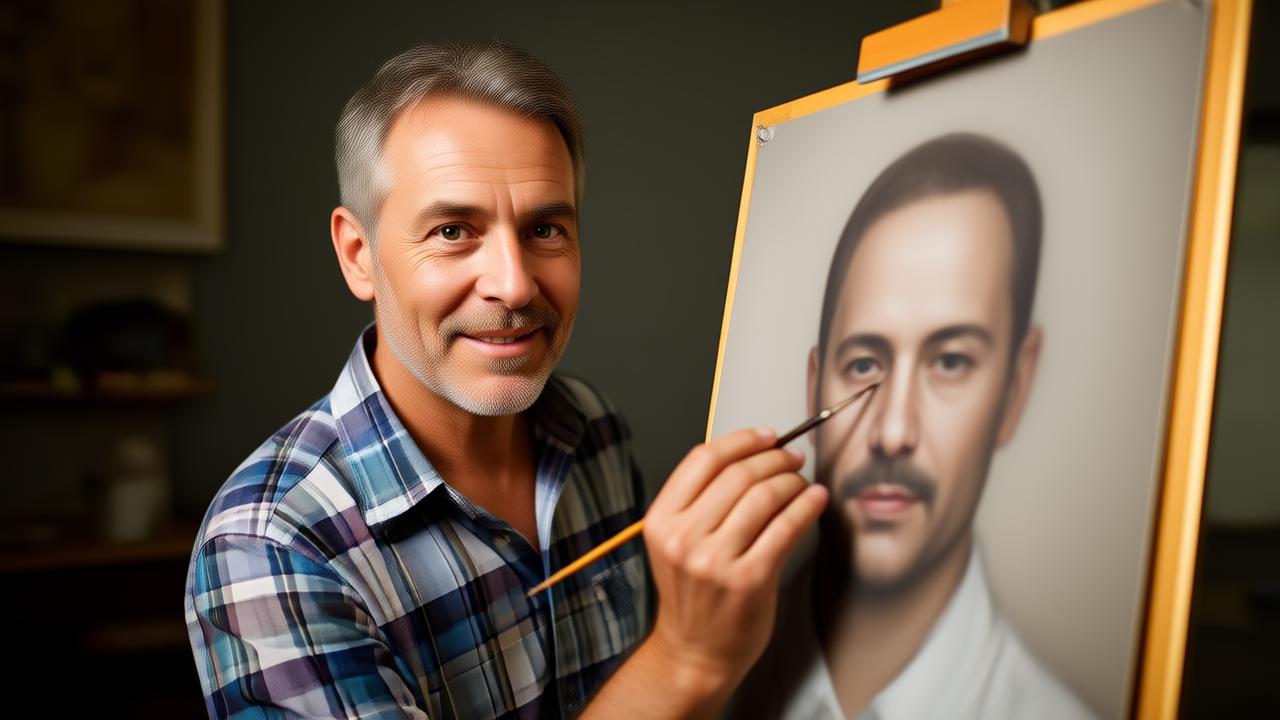}
  \end{subfigure}

  \begin{subfigure}{0.24\linewidth}
    \includegraphics[width=\linewidth]{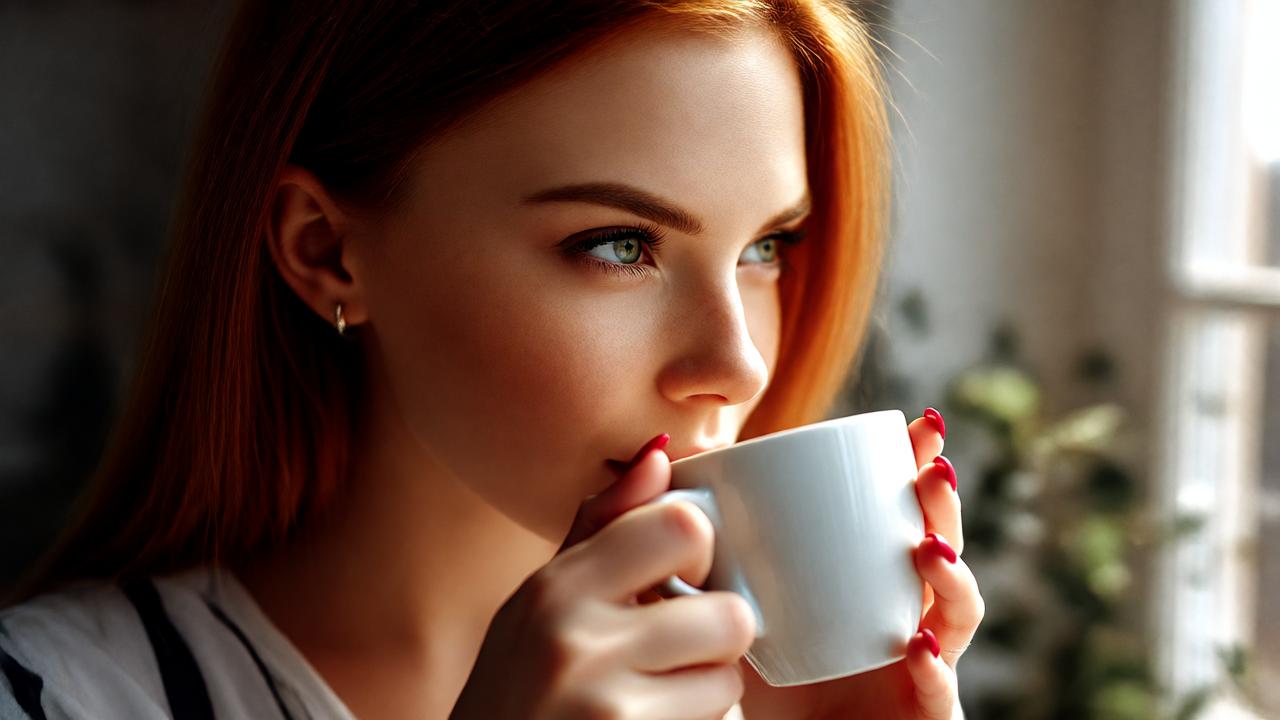}
  \end{subfigure}
  \begin{subfigure}{0.24\linewidth}
    \includegraphics[width=\linewidth]{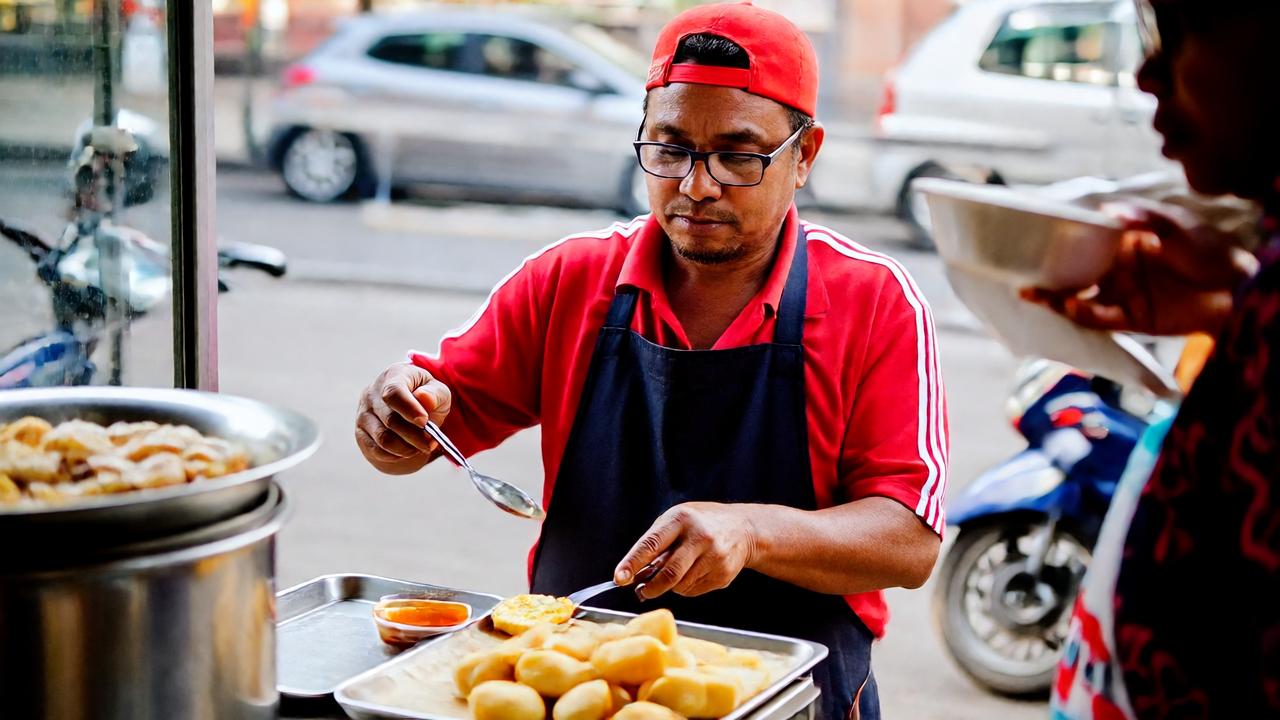}
  \end{subfigure}
  \begin{subfigure}{0.24\linewidth}
    \includegraphics[width=\linewidth]{figures/images/250921_t2i_ablation/wan21_phase_dmd/t2vA14B_0018_seed1_A_mother_braiding_her_daughters_hair_sunlight_wa_20250921_193257.jpg}
  \end{subfigure}
  \begin{subfigure}{0.24\linewidth}
    \includegraphics[width=\linewidth]{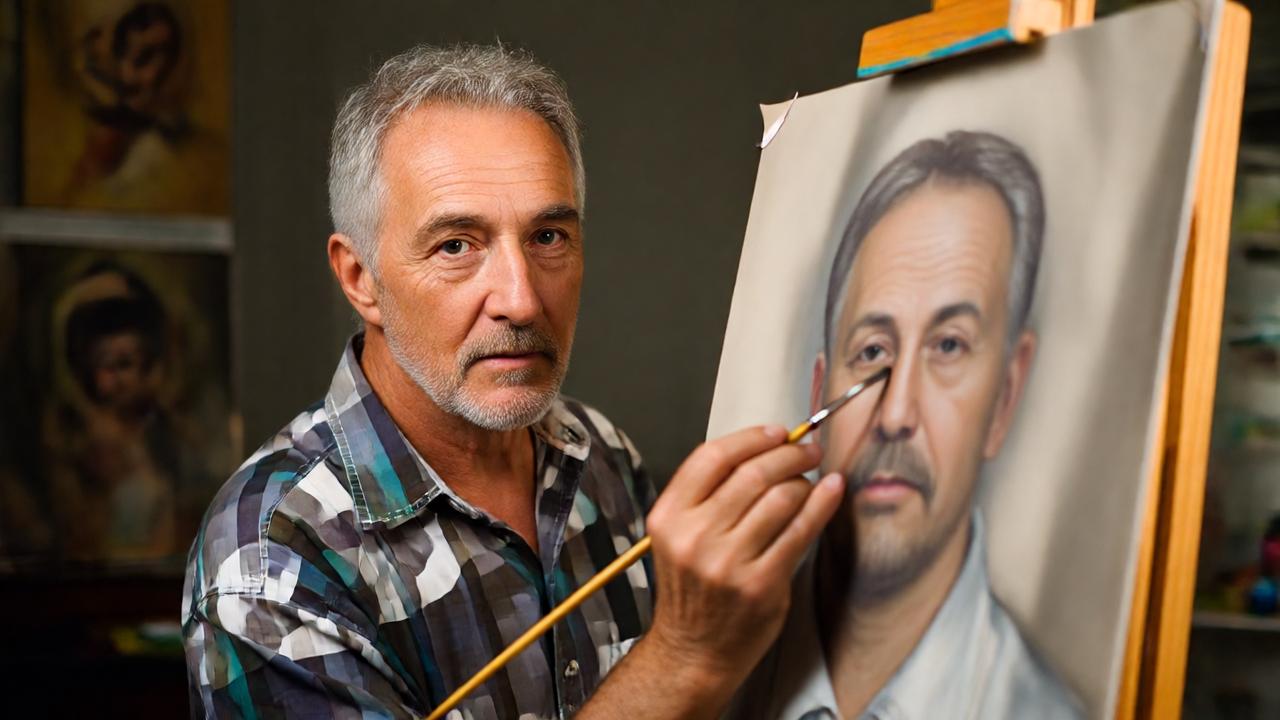}
  \end{subfigure}
  \caption{
    Samples generated with high-SNR experts from different training stages (top: 100 iterations; bottom: 400 iterations) and a shared low-SNR expert. Each column uses identical prompts and seeds.
  }
  \label{fig:image_structure_unchanged}
\end{figure}

Our empirical observations indicate that during distillation, DMD first captures structural information before acquiring finer textural details.
Prior to fully learning these textural nuances, generated images and videos often exhibit overly smooth characteristics, such as blurred hair or plastic-like skin textures.
% Prior to the complete acquisition of textural details, the generated images and videos tend to exhibit overly smooth features, such as blurry hair and plastic-like skin textures.
Meanwhile, the mode-seeking behavior of the reverse KL divergence causes a decline in generative diversity and motion intensity as training continues.
% On the other hand, the mode-seeking nature of reverse KL divergence leads to a decline in generative diversity and motion dynamic degree as training iterations increase. 
% 
\ourmodel mitigates this trade-off by partitioning the distillation process into distinct training phases.
In the low-SNR phases, the compositional structure of images and videos is effectively established.
In subsequent high-SNR phases, the low-SNR expert is frozen, enabling  prolonged training to refine generative quality without compromising the established structural layout. 
Combined with its inherent avoidance of one-step degeneration, \ourmodel leverages the MoE architecture to enhance fine-grained details while preserving structural diversity and motion intensity.
As shown in Fig.~\ref{fig:image_structure_unchanged}, extending training of the high-SNR expert primarily influences lighting and textural details, while leaving the overall structural composition of the images intact. 
% To validate the analysis, we provide visualization 
% 

\section{Related Work}
Our work is situated within the framework of Variational Score Distillation (VSD) \citep{wang2023prolificdreamer_vsd}.
VSD involves three components: a trainable generator, a fake score estimator, and a pretrained teacher score estimator. The generator is optimized to produce a distribution that approximates the real data distribution. Concurrently, the fake score estimator learns to estimate the score of the generator's output distribution. The update direction for the generator is then determined by the discrepancy between the teacher's score (for the real distribution) and the fake score estimator's score.

Similar to GANs, the VSD framework is adversarial. The fake score estimator must be precisely optimized to learn the score of the current generated distribution. This accurate estimation is crucial, as it combines with the fixed teacher model (which provides the score for the real data) to produce a correct guidance signal for the generator.
This principle explains why DMD2 \citep{yin2024DMD2} operates successfully without external real data, in contrast to its predecessor DMD \citep{yin2024_DMD1}.

A key advantage of VSD over GANs for distilling pre-trained diffusion models is initialization. The pre-trained model serves a dual role: it is a powerful multi-step generator and an accurate estimator of the real data distribution's score. This allows it to effectively initialize all three components in the VSD framework, leading to significantly enhanced training stability.

Several methods are built upon the VSD framework, including Diff-Instruct \citep{luo2023diff_instruct}, DMD \citep{yin2024DMD2}, SID \citep{zhou2024sid}, and FGM \citep{huang2024fgm}. The fundamental distinction between these approaches lies in the specific divergence they minimize. DMD, for instance, optimizes the reverse KL divergence between the real and generated distributions. A key advantage of this choice is its computational efficiency compared to alternatives like the Fisher divergence used in SID \citep{zhou2024sid}.
Specifically, during generator optimization, DMD does not require gradients to be backpropagated through the fake and teacher score estimators, whereas SID does. This does not imply the two estimators are trainable in this stage for SID, but rather reflects a difference in the computational graph.
This property makes DMD more amenable to engineering implementation and scalable to large base models.

Similar to our work, TDM \citep{luo2025TDM} also aimed to extend DMD to few-step distillation. However, our approach differs from TDM in three key aspects:
(a) The lack of proper theoretical grounding in TDM renders its fake flow training formulation suboptimal, undermining the foundations of DMD.
(b) Our framework inherently produces MoE models for few-step generation.
(c) While TDM uses disjoint SNR intervals, our method employs reverse nested intervals, where each interval is a subset of the subsequent one.

\section{Conclusion and Discussion}
Phased DMD primarily enhances structural aspects of generation, such as composition diversity, motion dynamics, and camera control.
However, for base models like Qwen-Image, whose outputs are inherently less diverse, the improvement is less pronounced.
While this work demonstrates phased distillation within the DMD framework, the approach is generalizable to other objectives like Fisher divergence in SiD \citep{zhou2024sid}, which we leave for future exploration.
It is conceivable that other methods for enhancing diversity and dynamics, such as incorporating trajectory data pre-generated by the base model, could be integrated. However, this would compromise the data-free advantage central to DMD. While we may explore such directions in the future, this work prioritizes the data-free paradigm.

\clearpage
{
\bibliography{main}
\bibliographystyle{main}
}

\clearpage

%%%%%%%%%% Appendix %%%%%%%%%%

\beginappendix
\section{Detailed Derivation}
\label{supp_sec:detailed_derivation}
We show the detailed derivation of Eq.~\ref{eq:FlowMatchTarget_2} as follows:
\begin{align}
 & J_{flowmatch}  \\
& =  E_{x_0 \sim p(x_0),  t \sim \mathcal{T},
\boldsymbol{\epsilon} \sim \mathcal{N}, x_t = \alpha_t x_0 + \sigma_t \boldsymbol{\epsilon}} [\|\boldsymbol{\psi}(x_t, t) - (\boldsymbol{\epsilon} - x_0) \|^2]  \\
& =   E_{x_0 \sim p(x_0),  t \sim \mathcal{T}, 
\boldsymbol{\epsilon} \sim \mathcal{N}, x_t = \alpha_t x_0 + \sigma_t \boldsymbol{\epsilon}}  [\|\boldsymbol{\psi}(x_t, t) - (\boldsymbol{\epsilon} - \frac{x_t - \sigma_t \boldsymbol{\epsilon}}{\alpha_t}) \|^2]  \\
& =   E_{x_0 \sim p(x_0),   t \sim \mathcal{T}, \boldsymbol{\epsilon} \sim \mathcal{N}, x_t = \alpha_t x_0 + \sigma_t \boldsymbol{\epsilon}}[\|\boldsymbol{\psi}(x_t, t) + \frac{1}{\alpha_t}x_t  - (1 + \frac{\sigma_t}{\alpha_t})\boldsymbol{\epsilon}\|^2]  \\
& =  E_{x_0 \sim p(x_0), t \sim \mathcal{T}, x_t \sim p(x_t | x_0)} [\|\boldsymbol{\psi}(x_t, t) + \frac{1}{\alpha_{t}}x_t  + (\sigma_t + \frac{\sigma_t^2}{\alpha_t})  \nabla_{x_t}\log(p(x_t | x_0)) \|^2]  \\
& =  E_{t \sim \mathcal{T}, x_t \sim p(x_t)} [\|\boldsymbol{\psi}(x_t, t) + \frac{1}{\alpha_{t}}x_t   + (\sigma_t + \frac{\sigma_t^2}{\alpha_t})\nabla_{x_t}\log(p(x_t)) \|^2] 
\end{align}
In the derivation, we use the score of $p(x_t | x_0)$, i.e.,
$\nabla_{x_t}\log(p(x_t | x_0)) = - \frac{1}{\sigma_t} \boldsymbol{\epsilon}$, and the equivalence between DSM and ESM \citep{vincent2011_connection_DSM_ESM}.

We show the detailed derivation of Eq.~\ref{eq:FlowMatchTarget_conditional} as follows:
\begin{align}
& J_{subinterval-flowmatch} \\
& = E_{t \sim \mathcal{T}(t; s, 1), x_t \sim p(x_t)} [\|\boldsymbol{\psi}(x_t, t) + \frac{1}{\alpha_{t}}x_t + (\sigma_t + \frac{\sigma_t^2}{\alpha_t})\nabla_{x_t}\log(p(x_t)) \|^2]  \\
& =  E_{x_s \sim p(x_s), t \sim \mathcal{T}(t; s, 1), x_t \sim p(x_t | x_s)} [\|\boldsymbol{\psi}(x_t, t) + \frac{1}{\alpha_{t} } x_t  + (\sigma_t + \frac{\sigma_t^2}{\alpha_t}) \nabla_{x_t}\log(p(x_t | x_s)) \|^2]  \\
& =  E_{x_s \sim p(x_s), t \sim \mathcal{T}(t; s, 1), \boldsymbol{\epsilon} \sim \mathcal{N}, x_t = \alpha_{t|s} x_s + \sigma_{t|s} \boldsymbol{\epsilon}} [\|\boldsymbol{\psi}(x_t, t) + \frac{1}{ \alpha_t } x_t - \frac{\sigma_t + \frac{\sigma_t^2}{\alpha_t}}{\sigma_{t|s}} \boldsymbol{\epsilon}  \|^2] \\
& =  E_{x_s \sim p(x_s), t \sim \mathcal{T}(t; s, 1), \boldsymbol{\epsilon} \sim \mathcal{N}, x_t = \alpha_{t|s} x_s + \sigma_{t|s} \boldsymbol{\epsilon}} [\|\boldsymbol{\psi}(x_t, t) + \frac{\alpha_{t|s} x_s + \sigma_{t|s} \boldsymbol{\epsilon}}{ \alpha_t}  - \frac{\sigma_t + \frac{\sigma_t^2}{\alpha_t }}{\sigma_{t|s}} \boldsymbol{\epsilon}  \|^2] \\
&  = E_{x_s \sim p(x_s), t \sim \mathcal{T}(t; s, 1), \boldsymbol{\epsilon} \sim \mathcal{N}, x_t = \alpha_{t|s} x_s + \sigma_{t|s} \boldsymbol{\epsilon}} [\|\boldsymbol{\psi}(x_t, t) - (\frac{\alpha_s^2 \sigma_t + \alpha_t \sigma_s^2}{\alpha_s ^ 2 \sigma_{t|s}} \boldsymbol{\epsilon} - \frac{1}{ \alpha_s} x_s ) \|^2]
\end{align}

The relationship between sample prediction (x-prediction) and score matching is derived as follows:
\begin{align}
& J_{sample} \\
&  = E_{x_0 \sim p(x_0),
 t \sim \mathcal{T}, \boldsymbol{\epsilon} \sim \mathcal{N}, x_t = \alpha_t x_0 + \sigma_t \boldsymbol{\epsilon}}[\|\boldsymbol{\mu}(x_t, t) - x_0 \|^2]  \\
&  = E_{x_0 \sim p(x_0), t \sim \mathcal{T}, \boldsymbol{\epsilon} \sim \mathcal{N}, x_t = \alpha_t x_0 + \sigma_t \boldsymbol{\epsilon}} [\|\boldsymbol{\mu}(x_t, t) - \frac{x_t - \sigma_t \boldsymbol{\epsilon}}{\alpha_t} \|^2] \\
&  = E_{x_0 \sim p(x_0),
 t \sim \mathcal{T}, \boldsymbol{\epsilon} \sim \mathcal{N}, x_t = \alpha_t x_0 + \sigma_t \boldsymbol{\epsilon}} [\|\boldsymbol{\mu}(x_t, t) - \frac{1}{\alpha_t}x_t + \frac{\sigma_t}{\alpha_t}\boldsymbol{\epsilon} \|^2]  \\
& =  E_{x_0 \sim p(x_0), t \sim \mathcal{T}, x_t \sim p(x_t | x_0)} [\|\boldsymbol{\mu}(x_t, t) - \frac{1}{\alpha_t}x_t - \frac{\sigma_t^2}{\alpha_t}\nabla_{x_t}\log(p(x_t | x_0)) \|^2]  \\
& =  E_{t \sim \mathcal{T}, x_t \sim p(x_t)} [\|\boldsymbol{\mu}(x_t, t) - \frac{1}{\alpha_t} x_t  - \frac{\sigma_t^2}{\alpha_t}\nabla_{x_t}\log(p(x_t)) \|^2] 
\end{align}

The training objective for x-prediction diffusion models within a subinterval is as follows:
\begin{align}
& J_{subinterval-sample} \\
& =  E_{t \sim \mathcal{T}, x_t \sim p(x_t)} [\|\boldsymbol{\mu}(x_t, t) - \frac{1}{\alpha_t} x_t  - \frac{\sigma_t^2}{\alpha_t}\nabla_{x_t}\log(p(x_t)) \|^2]  \\
& =  E_{x_s \sim p(x_s), t \sim \mathcal{T}(t; s, 1), x_t \sim p(x_t | x_s)} [\|\boldsymbol{\mu}(x_t, t) - \frac{1}{\alpha_t} x_t - \frac{\sigma_t^2}{\alpha_t}\nabla_{x_t}\log(p(x_t | x_s)) \|^2] \\
& =  E_{x_s \sim p(x_s), t \sim \mathcal{T}(t; s, 1), \boldsymbol{\epsilon} \sim \mathcal{N}, x_t = \alpha_{t|s} x_s + \sigma_{t|s} \boldsymbol{\epsilon}} [\|\boldsymbol{\mu}(x_t, t) - \frac{1}{\alpha_t} x_t   +  \frac{\sigma_t^2}{\alpha_t \sigma_{t|s}}\epsilon \|^2] \\
& =  E_{x_s \sim p(x_s), t \sim \mathcal{T}(t; s, 1), \boldsymbol{\epsilon} \sim \mathcal{N}, x_t = \alpha_{t|s} x_s + \sigma_{t|s} \boldsymbol{\epsilon}} [\|\boldsymbol{\mu}(x_t, t) - \frac{\alpha_{t|s} x_s + \sigma_{t|s} \boldsymbol{\epsilon}}{\alpha_t} + \frac{\sigma_t^2}{\alpha_t\sigma_{t|s}} \boldsymbol{\epsilon} \|^2]  \\
& =  E_{x_s \sim p(x_s), t \sim \mathcal{T}(t; s, 1), \boldsymbol{\epsilon} \sim \mathcal{N},  x_t = \alpha_{t|s} x_s + \sigma_{t|s} \boldsymbol{\epsilon}} [\|\boldsymbol{\mu}(x_t, t) - ( \frac{1}{\alpha_s}x_s - \frac{\alpha_t\sigma_s^2}{\alpha_s ^ 2 \sigma_{t|s}} \boldsymbol{\epsilon}) \|^2]
\label{eq:x_prediction_subinterval}
\end{align}
% Optimizing within the subinterval according to Eq.~\ref{eq:x_prediction_subinterval} gives an unbiased estimation of x-prediction.  In contrast, the objective $[\|\boldsymbol{\mu}_{\boldsymbol{\theta}}(x_t) - x_s \|^2]$ yields a biased estimation.

\section{Experimental Details}
\label{sec:appendix_exps}
For the Wan2.x base models, distillation on the text-to-image task is conducted at a fixed data resolution of one frame with width 1280 and height 720, \ie, $\text{frame}=1, \text{width}=1280, \text{height}=720$.

For the Wan2.2-x2V-A14B model, distillation on both the text-to-video and image-to-video tasks employs a mixture of data resolutions: $(81, 720, 1280)$, $(81, 1280, 720)$, $(81, 480, 832)$, $(81, 832, 480)$,  sampled with probabilities $0.1, 0.1, 0.4, 0.4$.

For the Qwen-Image model, distillation on the text-to-image task employs a uniform sampling from a set of resolutions: $(1, 1382, 1382)$, $(1, 1664, 928)$, $(1, 928, 1664)$, $(1, 1472, 1104)$,  $(1, 1104, 1472 )$, $(1, 1584, 1056)$, $(1, 1056, 1584)$.

\begin{figure}[hb]
  \centering
    \includegraphics[width=0.65\linewidth]{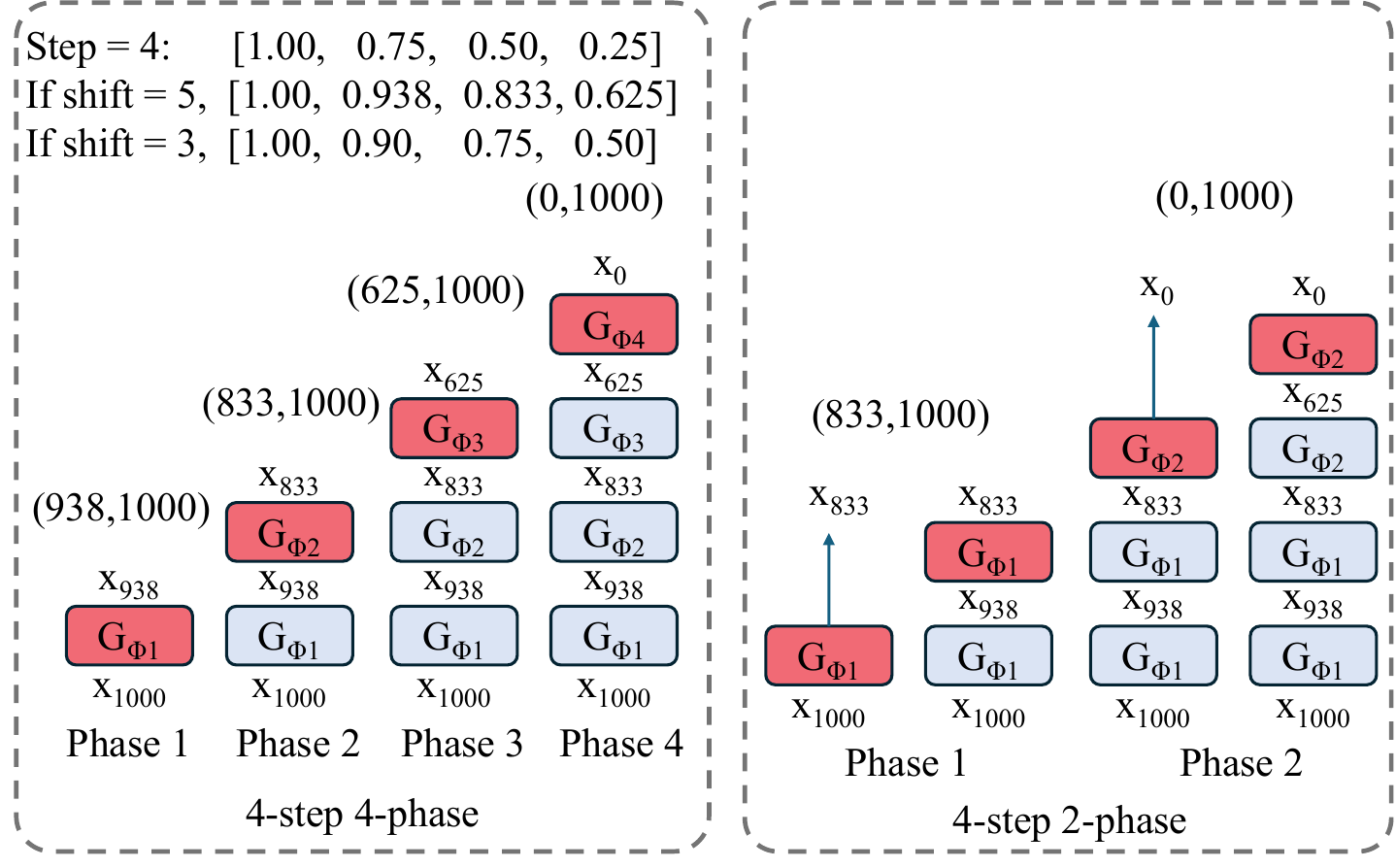}
  \caption{
    The timestep intervals are determined by the steps and the timestep shift value. For Wan-based models, shift = 5 is used, following the Self-Forcing approach. For Qwen-Image-based models, shift = 3 is adopted, based on ComfyUI's Qwen-Image workflow. The re-noising t range for each phase is represented as a tuple (min step, max step) and t is uniformly sampled before applying the shift. For instance, in a 4-step, 2-phase setting, t is sampled within (0.5, 1) and then apply a shift of 5 during Phase 1.}
  \label{fig:supp_Schematic_diagram}
\end{figure}

A timestep shift of 5 is applied for Wan2.x base models, following the self-forcing approach~\cite{huang2025_self_forcing} while a shift of 3 is used for the Qwen-Image base model, based on ComfyUI's Qwen-Image workflow. The resulting timesteps are provided in Tab.~\ref{tab:experiment_summary}. Fig.~\ref{fig:supp_Schematic_diagram} illustrates how intervals are determined by the steps and the timestep shift value.

For Wan2.2 base models, the first training phase exclusively employs the high-noise model. During this phase, the re-noising timestep $t$ is restricted (by torch.clamp) to the range (0.875, 1) for the T2V task and (0.9, 1) for the I2V task,  in accordance with the boundary timestep configuration of the high-noise model.
In the second phase, both the high-noise and low-noise models are employed and three components are trainable: the low-noise generator, the high-noise fake model and the low-noise fake model. The choice of high-noise or low-noise teacher and fake model is determined by the values of the re-noising timestep.
This training paradigm aligns with our strategy for sampling noise injection timesteps, as detailed in Sec.~\ref{sec:appendix subinterval}.

\section{More Results}
\subsection{Additional Quantitative Evaluation on Video Generation}
% 
% \begin{table*}[tbh]
% \caption{
% More quantitative comparison on text-to-video generation. The base model is Wan2.2-T2V-A14B.}
% \label{tab:more_vbench_t2v}
% \centering
% \begin{tabular}{c | c c c c c}
% \toprule
% \multirow{2}{*}{\bf{Method}} &  \bf{aesthetic}  & \bf{background} & \bf{motion}   & \bf{subject}  & \bf{temporal}   \\
% &  \bf{quality}  & \bf{consistency} & \bf{smoothness}   & \bf{consistency}  & \bf{flickering}   \\
% \midrule
% Base model & 63.62 \% & 94.03 \% & 97.67 \% & 90.06 \% & 95.70 \%  \\
% \midrule
% DMD with SGTS & 67.02 \%  & 95.29 \% & 98.57 \% & 92.93 \% & 97.08 \%  \\
% Phased DMD(Ours) & 65.73 \% & 94.40 \% & 97.74 \% & 91.15 \% & 95.26 \%  \\
% \bottomrule
% \end{tabular}
% \end{table*}

\begin{table*}[tbh]
\small
\caption{
More quantitative comparison on text-to-video generation. The base model is Wan2.2-T2V-A14B.}
\label{tab:more_vbench_t2v}
\centering
\begin{tabular}{c | c c c c c}
\toprule
\multirow{2}{*}{\bf{Method}} &  \bf{aesthetic}  & \bf{background} & \bf{motion}   & \bf{subject}  & \bf{temporal}   \\
&  \bf{quality}  & \bf{consistency} & \bf{smoothness}   & \bf{consistency}  & \bf{flickering}   \\
\midrule
Base model & 63.62 \% & 94.03 \% & 97.67 \% & 90.06 \% & 95.70 \%  \\
\midrule
DMD2 & 67.02 \% (\textcolor{ForestGreen}{+3.40 \%}) & 95.29 \% (\textcolor{ForestGreen}{+1.26 \%}) & 98.57 \% (\textcolor{ForestGreen}{+0.90 \%}) & 92.93 \% (\textcolor{ForestGreen}{+2.87 \%}) & 97.08 \% (\textcolor{ForestGreen}{+1.38 \%})  \\
Phased DMD(Ours) & 65.73 \% (\textcolor{ForestGreen}{+2.11 \%}) & 94.40 \% (\textcolor{ForestGreen}{+0.37 \%}) & 97.74 \% (\textcolor{ForestGreen}{+0.07 \%}) & 91.15 \% (\textcolor{ForestGreen}{+1.09 \%}) & 95.26 \% (\textcolor{RubineRed}{-0.44 \%})  \\
\bottomrule
\end{tabular}
\end{table*}
% \begin{table*}[tbh]
% \caption{
% More quantitative comparison on image-to-video generation. The base model is Wan2.2-I2V-A14B.}
% \label{tab:more_vbench_i2v}
% \centering
% \begin{tabular}{c | c c c c c}
% \toprule
% \multirow{2}{*}{\bf{Method}} &  \bf{aesthetic}  & \bf{background} & \bf{motion}   & \bf{subject}  & \bf{temporal}   \\
% &  \bf{quality}  & \bf{consistency} & \bf{smoothness}   & \bf{consistency}  & \bf{flickering}   \\
% \midrule
% Base model & 62.71 \% & 93.75 \% & 97.74 \% & 90.39 \% & 95.52 \%  \\
% \midrule
% DMD with SGTS & 64.14 \% & 94.44 \% & 97.85 \% & 91.84 \% & 95.73 \%  \\
% Phased DMD(Ours) & 63.91 \% & 94.16 \% & 97.66 \% & 91.40 \% & 95.17 \%  \\
% \bottomrule
% \end{tabular}
% \end{table*}

\begin{table*}[tbh]
\small
\caption{
More quantitative comparison on image-to-video generation. The base model is Wan2.2-I2V-A14B.}
\label{tab:more_vbench_i2v}
\centering
\begin{tabular}{c | c c c c c}
\toprule
\multirow{2}{*}{\bf{Method}} &  \bf{aesthetic}  & \bf{background} & \bf{motion}   & \bf{subject}  & \bf{temporal}   \\
&  \bf{quality}  & \bf{consistency} & \bf{smoothness}   & \bf{consistency}  & \bf{flickering}   \\
\midrule
Base model & 62.71 \% & 93.75 \% & 97.74 \% & 90.39 \% & 95.52 \%  \\
\midrule
DMD2 & 64.14 \% (\textcolor{ForestGreen}{+1.43 \%}) & 94.44 \% (\textcolor{ForestGreen}{+0.69 \%}) & 97.85 \% (\textcolor{ForestGreen}{+0.11 \%}) & 91.84 \% (\textcolor{ForestGreen}{+1.45 \%}) & 95.73 \% (\textcolor{ForestGreen}{+0.21 \%})  \\
Phased DMD(Ours) & 63.91 \% (\textcolor{ForestGreen}{+1.20 \%}) & 94.16 \% (\textcolor{ForestGreen}{+0.41 \%}) & 97.66 \% (\textcolor{RubineRed}{-0.08 \%}) & 91.40 \% (\textcolor{ForestGreen}{+1.01 \%}) & 95.17 \% (\textcolor{RubineRed}{-0.35 \%})  \\
\bottomrule
\end{tabular}
\end{table*}
We present additional quantitative evaluation results in Tab.~\ref{tab:more_vbench_t2v} and Tab.~\ref{tab:more_vbench_i2v}.
Interestingly, across most evaluation dimensions, the base model using 40 inference steps (80 function evaluations)  exhibits the poorest performance. 
For instance, in text-to-video generation, the base model achieves an aesthetic quality score of only 63.62 \%, whereas both distilled variants using 4 steps (4 function evaluations) obtain higher scores.
Although the Vbench~\cite{huang2024vbench} quantitative metrics suggest that DMD2 achieves the best overall performance while the base model performs worst, human preference ratings show the opposite trend. 
In the attached video, we compile all 220 generated comparison videos for the T2V task. Note that the video has been heavily compressed to reduce file size. As the video clearly demonstrates, the base model exhibits the highest overall quality in terms of aesthetics and motion dynamics, while \ourmodel preserves the base model’s performance, substantially better than DMD2.
We argue that the rankings derived from the quantitative evaluations in Tab.~\ref{tab:more_vbench_t2v} and Tab.~\ref{tab:more_vbench_i2v} are not entirely reliable.
Nevertheless, the performance gaps between the distilled models and the base model reveal that \ourmodel produces values more closely aligned with those of the base model, indicating that \ourmodel better preserves the generative distribution of the original base model.

The results presented in Tab.~\ref{tab:motion_speed_quant} are based on the 4-step 2-phase configuration, as illustrated in Fig.~\ref{fig:Schematic_diagram}d. Tab.~\ref{tab:phased_dmd_without_sgts} demonstrates that the 4-step 4-phase configuration (Fig.~\ref{fig:Schematic_diagram}c) delivers better performance in both motion dynamics and visual quality, albeit at the cost of increased system complexity. This improvement can be attributed to the avoidance of SGTS and the inclusion of more trainable parameters.
\begin{table}[tbh]
\small
\caption{
Quantitative comparison of video generation performance. 
``OF'' refers to optical flow. ``DD'' refers to dynamic degree.
Completely removing SGTS from \ourmodel leads to improved performance.}
\label{tab:phased_dmd_without_sgts}
\centering
\begin{tabular}{c | c c c c}
\toprule
{\bf{Method}} &  \bf{OF} $ \uparrow $ & \bf{DD} $ \uparrow $ & \bf{FID} $ \downarrow $  & \bf{FVD} $ \downarrow $  \\
\midrule
Base model & \bf{10.26} & 79.55 \% & \bf{0.0} & \bf{0.0}  \\
\midrule
DMD2 & 3.23 & 65.45 \%  & 55.70 & 763.1   \\
Ours (4-step 2-phase) &  9.30 & \underline{82.27 \%}  & 47.24 & 700.9   \\
Ours (4-step 4-phase) &  $\underline{9.43}$ & \bf{83.18 \%}  & \underline{45.40} & \underline{578.2}   \\
\bottomrule
\end{tabular}
\end{table}

\subsection{Ablation on Diffusion Timestep Subintervals}
\label{sec:appendix subinterval}

\begin{figure}[tb]
\centering
\begin{subfigure}{0.3\linewidth}
\includegraphics[width=\linewidth]{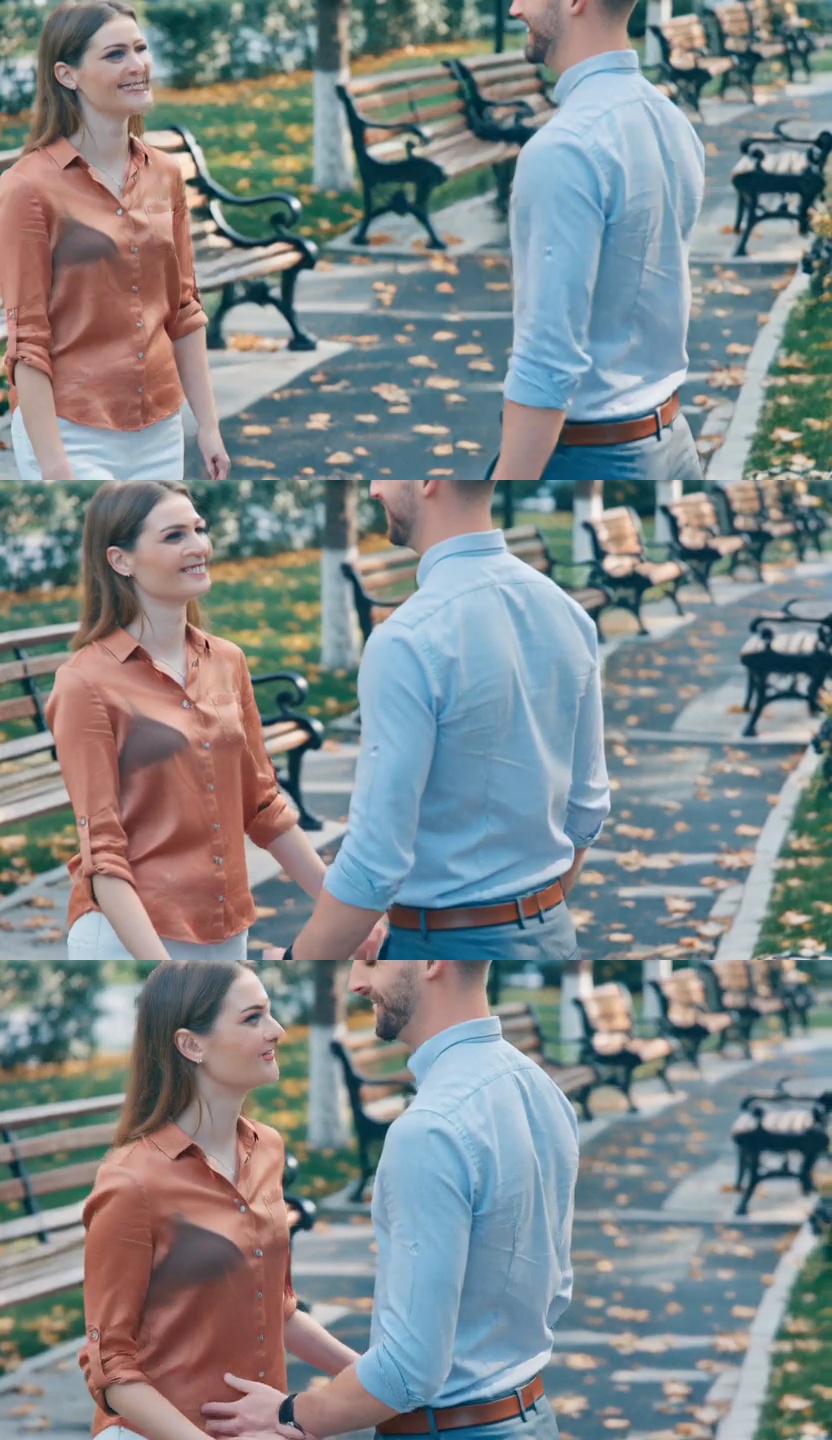}
\subcaption{Disjoint Intervals}
\label{subfig:disjoint_intervals}
\end{subfigure}
\begin{subfigure}{0.3\linewidth}
\includegraphics[width=\linewidth]{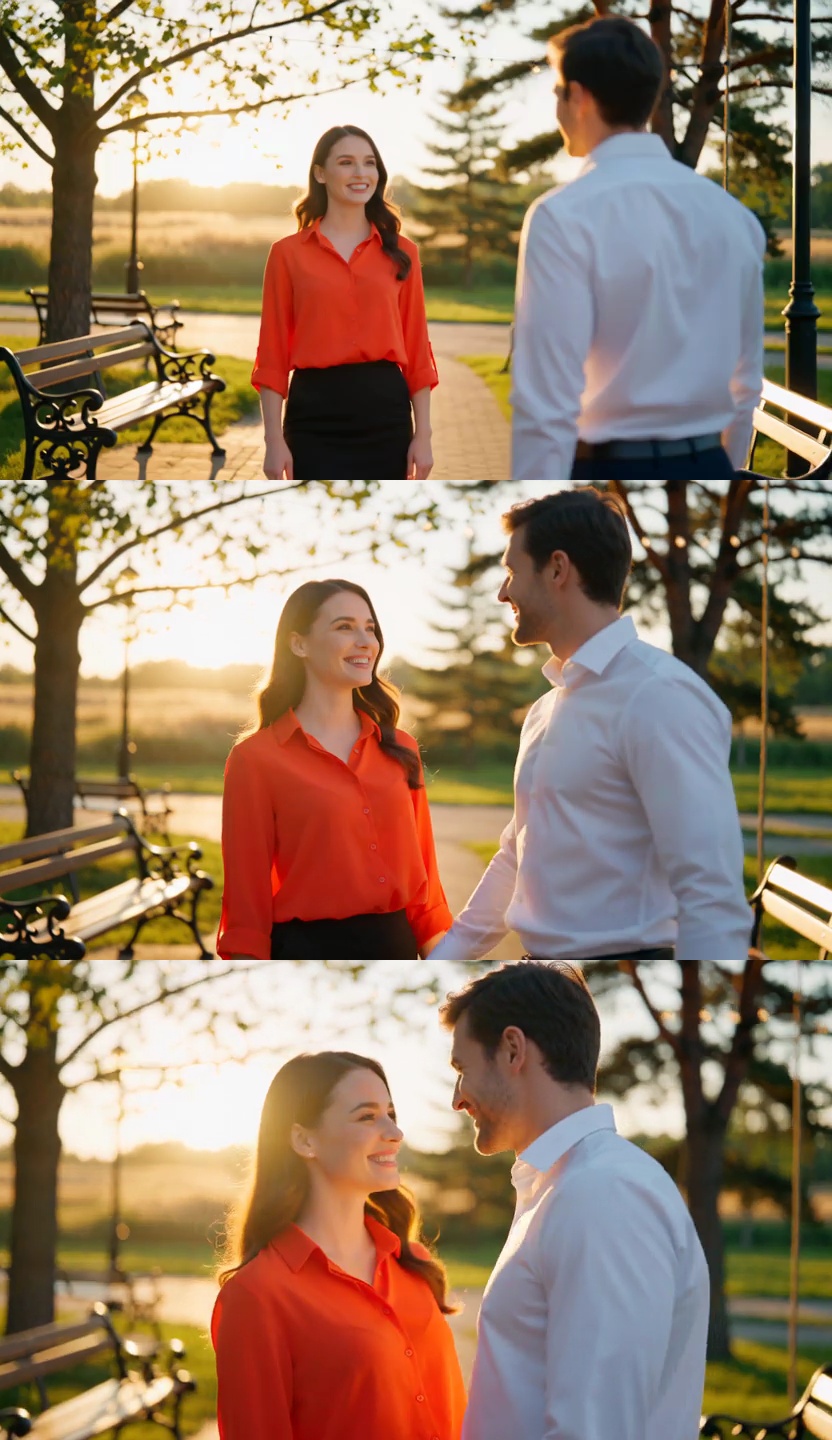}
\subcaption{Reverse Nested Intervals}
\label{subfig:reverse_nested_intervals}
\end{subfigure}
\caption{
The effect of noise injection intervals. \citet{luo2025TDM} employs disjoint noise injection timestep intervals for different generation steps, where the intervals do not overlap. In contrast, we adopt reverse nested intervals, where the diffusion timestep interval in each phase terminates at $1.0$.
Integrating disjoint intervals into Phased DMD  leads to unnatural colors and deteriorated facial structures, as illustrated on the left. Conversely, adopting reverse nested intervals yields correct results.
  }
  \label{fig:subinterval strategy}
\end{figure}

\begin{figure}[tb]
\centering
\begin{subfigure}{0.30\linewidth}
\includegraphics[width=\linewidth]{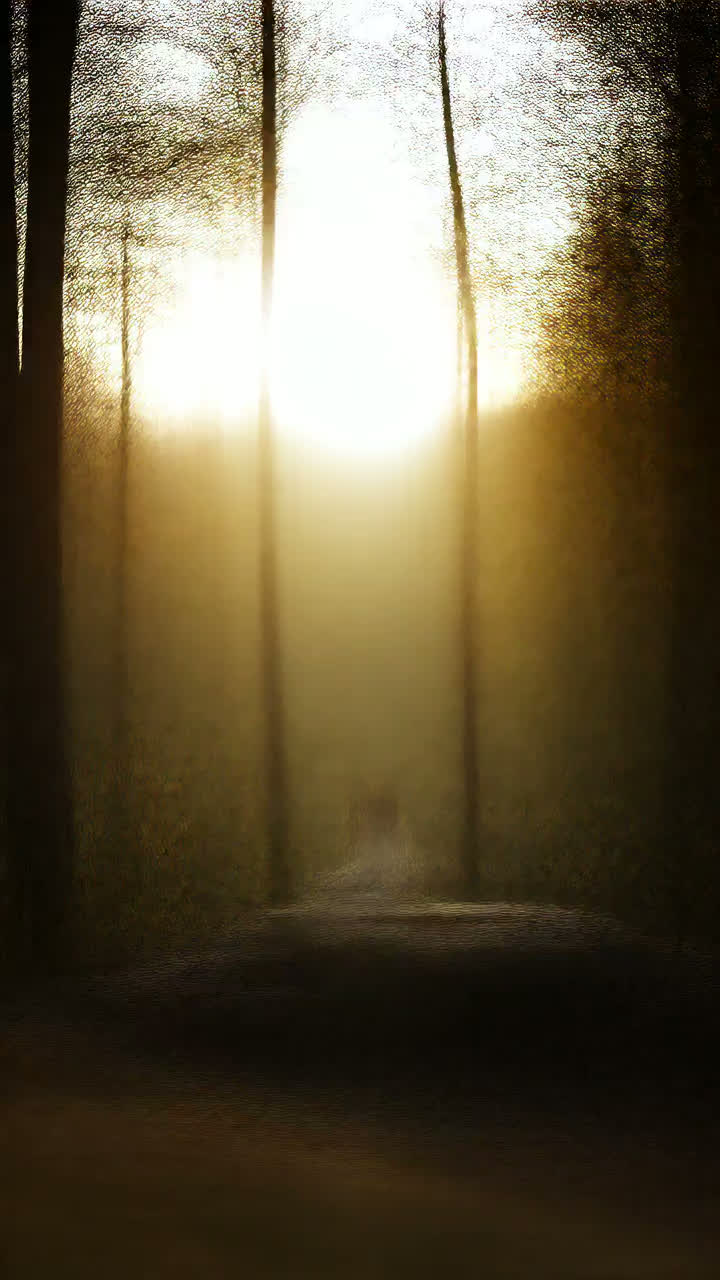}
\subcaption{t = 0.357}
\label{subfig:t_0.357}
\end{subfigure}
\begin{subfigure}{0.30\linewidth}
\includegraphics[width=\linewidth]{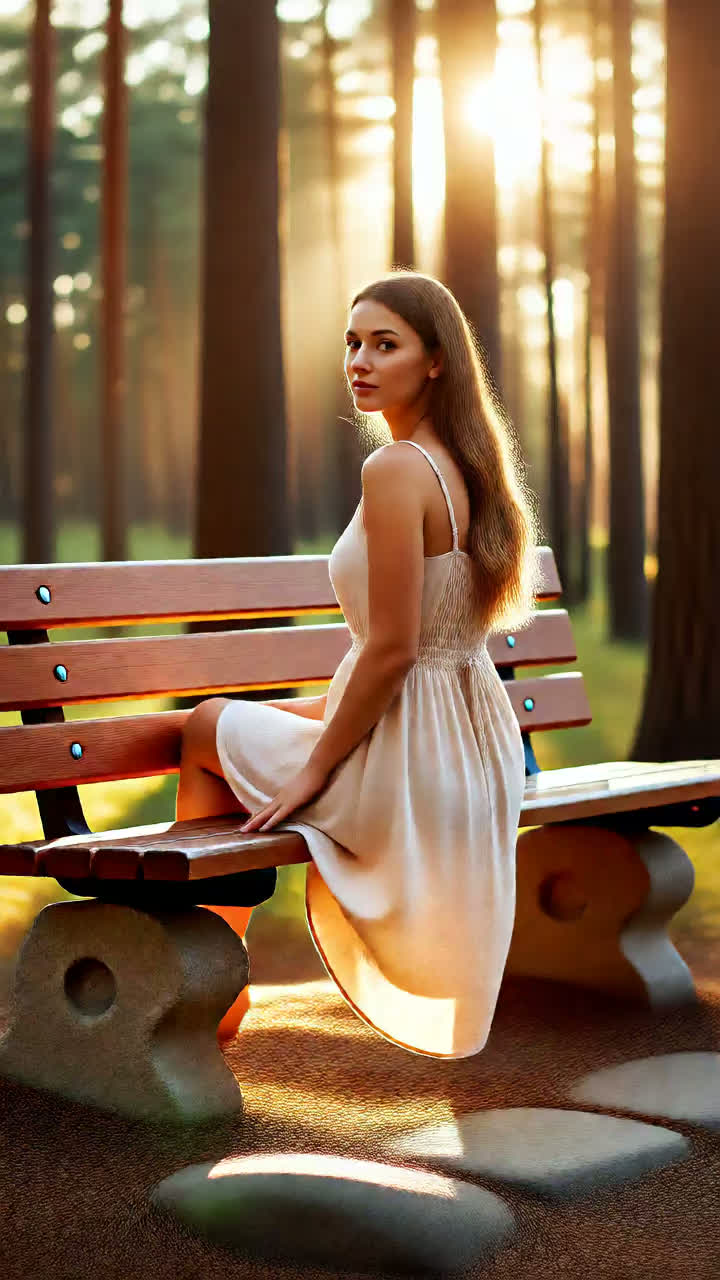}
\subcaption{t = 0.882}
\label{subfig:t_0.882}
\end{subfigure}
\caption{
The effect of noise injection timestep in DMD training. 
In DMD training, noise is injected into the generated samples at a low noise level (left) and a high noise level (right). The training fails to converge correctly when noise is injected exclusively at a low noise level.
}
\label{fig:fixed diffusion timestep}
\end{figure}

Empirically, we observe that sampling noise injection timesteps using \textbf{Reverse Nested Intervals} $t \sim \mathcal{T}(t; t_k, 1)$ outperforms \textbf{Disjoint Intervals} $t \sim \mathcal{T}(t; t_k, t_{k-1})$ in terms of generation quality. Fig.~\ref{fig:subinterval strategy} illustrates the results of these two methods in the Wan2.2 T2V distillation task. Specifically, sampling $t \sim \mathcal{T}(t; t_k, 1)$ yields normal color tones and accurate structures, whereas sampling $t \sim \mathcal{T}(t; t_k, t_{k-1})$ results in low-contrast tones and degraded facial structures.

At the beginning of each phase in Phased DMD, there is a substantial gap between the distribution of samples generated by the few-step generator and the distribution of real samples. 
The generated samples fall outside the domain of the teacher model, leading to inaccurate score estimations. This discrepancy is particularly pronounced in the high-SNR (low-noise level) range, where samples are less corrupted by noise. In contrast, in the low-SNR (high-noise level) range, the diffused generated distribution overlaps more significantly with the diffused real distribution, enabling the teacher model to provide more accurate score estimations. Consequently, noise injection at high-noise levels plays a crucial role in DMD training.

To validate this analysis, we perform ablation studies on vanilla DMD for the Wan2.1 T2I task. Specifically, the diffusion timestep $t$ is fixed at 0.357 for one experiment and at 0.882 for another. \citet{wang2023prolificdreamer_vsd} has proven that
$D_{KL}(p_{fake}(x_t) \| p_{real}(x_t)) = 0 \Leftrightarrow D_{KL}(p_{fake}(x_0) \| p_{real}(x_0)) = 0$  for any $0 < t < 1$.
Thus, both experiments are theoretically valid. However, the experiment with a diffusion timestep $t = 0.357$ fails to converge, as illustrated in Fig.~\ref{fig:fixed diffusion timestep}, while the experiment with $t = 0.882$ demonstrates correct results. This controlled experiment highlights that incorporating high-noise levels is essential for effective DMD training.
This observation, to some extent, explains our rationale for adopting \textbf{Reverse Nested Intervals}, wherein the training interval at each stage includes the high-noise range.

\subsection{Additional Discussion on MoE}

\begin{figure}[tb]
\centering
\begin{subfigure}{0.30\linewidth}
\includegraphics[width=\linewidth]{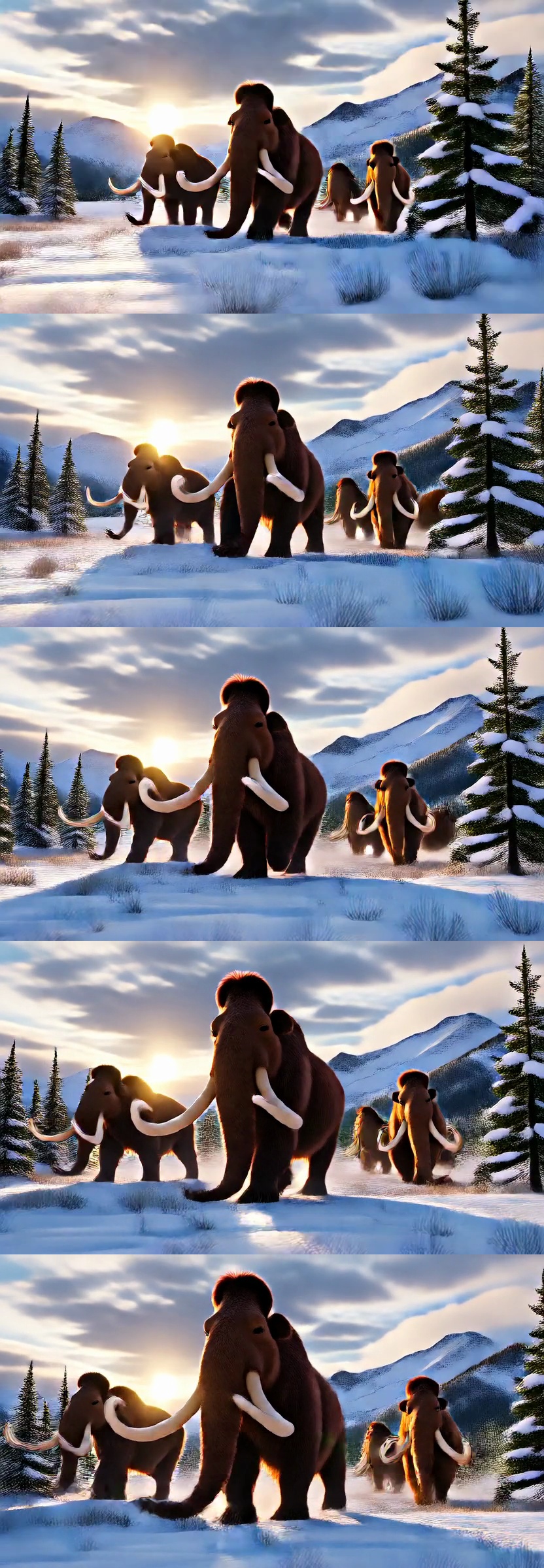}
\subcaption{low-SNR only}
\label{subfig:high_noise}
\end{subfigure}
\begin{subfigure}{0.30\linewidth}
\includegraphics[width=\linewidth]{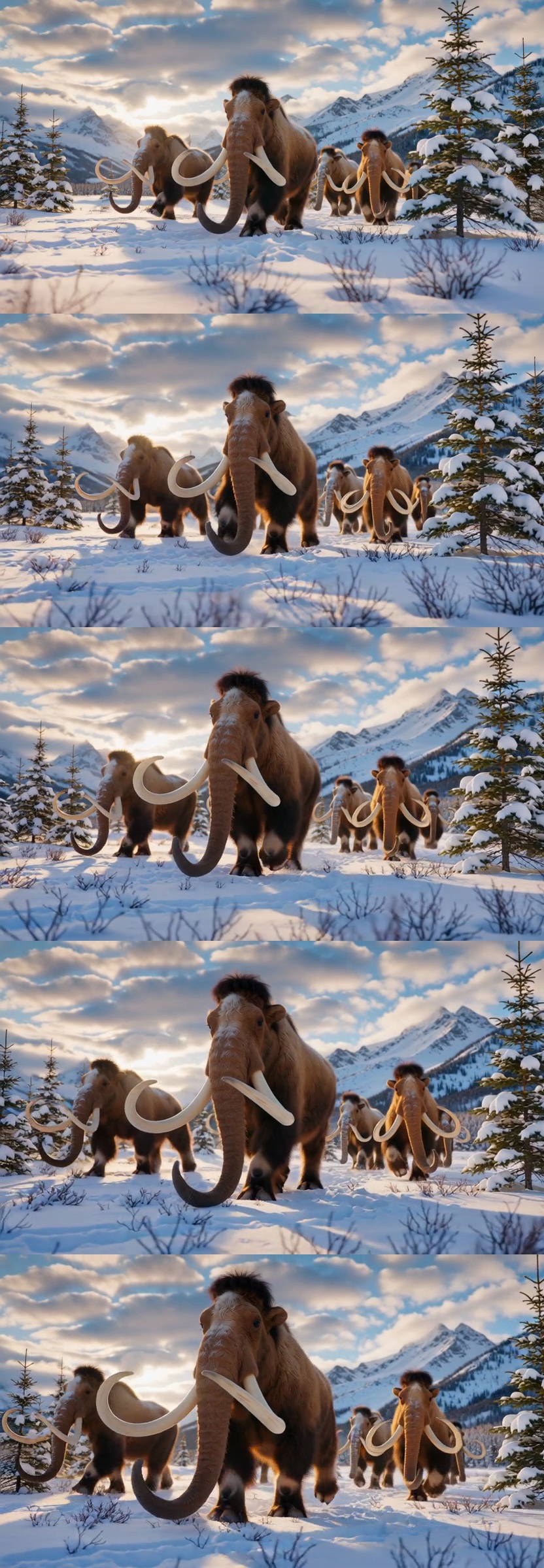}
\subcaption{low-SNR + high-SNR}
\label{subfig:low_noise}
\end{subfigure}
\caption{
Visualization of the functional roles of the low-SNR (high-noise) and high-SNR (low-noise) experts. (a) Video sequences generated by the distilled high-noise expert of Wan2.2-T2V-A14B, evaluated over only the first two denoising steps. (b) Video sequences generated by the combined pipeline of the distilled high-noise and low-noise experts, evaluated over all four denoising steps. A comparison of (a) and (b) demonstrates that the low-SNR expert is responsible for modeling global structure and dynamics, whereas the high-SNR expert refines local details.
}
\label{fig:role of high noise}
\end{figure}

Mixture-of-Experts (MoE) architectures are widely employed in large language models~\cite{yang2025qwen3technicalreport,deepseekai2025deepseekr1incentivizingreasoningcapability}, where they are typically adapted within the feed-forward network (FFN) layers.
In diffusion models, however, MoE is implemented differently.
Here, each expert typically features a dense architecture and is assigned to a specific denoising range, allowing it to be optimized for a distinct subset of the generative process.
This functional division, which aligns with the different requirements across the denoising trajectory, is illustrated in Fig.~\ref{fig:role of high noise}: the low-SNR (high-noise) stage is critical for modeling global structures and dynamics, whereas the high-SNR (low-noise) stage focuses on refining fine-grained details.
During inference, these experts are applied sequentially as the SNR increases throughout the sampling process.

The scaling of diffusion models has recently increased interest in MoE architectures. By dedicating experts to different SNR levels, MoE enhances model capacity and generative quality without a proportional increase in inference cost. This performance gain is particularly pronounced in video generation \cite{wan2025}, where a dedicated high-noise expert excels at capturing coherent temporal dynamics.

Our training framework, \ourmodel, is naturally compatible with such MoE-based models. For a base model with $N$ experts, the optimal practice is to employ an $N$-phase training scheme. In the $k$-th phase, the setup comprises one trainable generator and $k$ trainable fake models.
Empirical observations indicate that 4-step sampling represents a performance-efficient balance. Consequently, we posit that a base model architecture with four denoising experts is an effective choice.
Given the inherent compatibility between diffusion models and MoE architectures, as well as the demonstrated benefits of specialized experts, we argue that MoE will become an increasingly prevalent choice for image and video generation. Correspondingly, \ourmodel will play a more significant role in the distillation of diffusion models, as it is inherently compatible with MoE-based foundations.

\end{document}